\documentclass[12pt]{report}
\usepackage[latin2]{inputenc}
\usepackage[T1]{fontenc}
\usepackage{textcomp}

\usepackage{amsmath}
\usepackage[pdftex]{graphicx}
\usepackage{amsfonts}
\usepackage{amssymb}
\usepackage{gensymb}
\usepackage[left=1.5in]{geometry}
\usepackage{geometry}
\usepackage{wrapfig}
\usepackage{setspace}
\usepackage{epsfig}
\usepackage{color}
\usepackage{tikz}
\usepackage{UVMThesisStyle-March2016}
\usepackage{cancel}
\usepackage{xr}
\usepackage{float}
\usepackage{floatflt}
\usepackage{url}
\usepackage{hyperref}
\usepackage{booktabs}
\usepackage{multicol}
\usepackage{multirow}
\usepackage[export]{adjustbox}
\usepackage{soul,xcolor}
\usepackage{placeins}

\usepackage[backend=biber,maxbibnames=99,defernumbers=true,style=ieee,giveninits=true]{biblatex}
\DeclareMathOperator*{\alr}{\leftrightarrow}
\DeclareMathOperator*{\aud}{\updownarrow}

\title{Exploring the effects of robotic design on learning and neural control}
\author{Joshua Paul Powers}
\defensedate{July 6, 2021}
\dissertation                     %% or \thesis
\doctorphilosophy               %% or \doctorphilosophy
\cs                          %% this is the only speciality defined.
\auggrad                           %% \maygrad or \octgrad or \marchgrad
\advisor{Josh Bongard, Ph.D.}
\chair{Mathew Weston, Ph.D.}
\readerone{Safwan Wshah, Ph.D.}
\readertwo{Joe Near, Ph.D.}
\dean{Cynthia J. Forehand, Ph.D.}

\begin{document}

\maketitle
%\makeacceptance
\pagenumbering{roman}

\begin{abstract}
\vspace{10mm}
The ongoing deep learning revolution has allowed computers to outclass humans in various games and perceive features imperceptible to humans during classification tasks. Current machine learning techniques have clearly distinguished themselves in specialized tasks. However, we have yet to see robots capable of performing multiple tasks at an expert level. Most work in this field is focused on the development of more sophisticated learning algorithms for a robot's controller given a largely static and presupposed robotic design. By focusing on the development of robotic bodies, rather than neural controllers, I have discovered that robots can be designed such that they overcome many of the current pitfalls encountered by neural controllers in multitask settings. Through this discovery, I also present novel metrics to explicitly measure the learning ability of a robotic design and its resistance to common problems such as catastrophic interference.

Traditionally, the physical robot design requires human engineers to plan every aspect of the system,  which is expensive and often relies on human intuition. In contrast, within the field of evolutionary robotics, evolutionary algorithms are used to automatically create optimized designs, however, such designs are often still limited in their ability to perform in a multitask setting. The metrics created and presented here give a novel path to automated design that allow evolved robots to synergize with their controller to improve the computational efficiency of their learning while overcoming catastrophic interference.

Overall, this dissertation intimates the ability to automatically design robots that are more general purpose than current robots and that can perform various tasks while requiring less computation.
\end{abstract}

\section*{Citations}
{
Material from this dissertation has been published in the following form:\newline

\noindent
Powers. J., Kriegman. S., and Bongard. J.. ``The effects of morphology and fitness on catastrophic interference,'' in Artificial Life Conference Proceedings. MIT Press, 2018, pp. 606-613.\newline

\noindent
Powers. J., Grindle. R., Kriegman. S., Frati. L., Cheney. N., and Bongard. J.. ``Morphology dictates learnability in neural controllers,'' in Artificial Life Conference Proceedings. MIT Press One Rogers Street, Cambridge, MA 02142-1209 USA journals-info mit, 2020, pp. 52-59.\newline

\noindent
Powers. J., Grindle. R., Frati. L., and Bongard. J.. ``A good body is all you need: avoiding catastrophic interference via agent architecture search,'' in Online digital archive for electronic preprints of scientific papers (ArXiv) 2021.\newline

\noindent
Shah. D.*, Powers. J.*, Tilton. L., Kriegman. S., Bongard. J., and Kramer-Bottiglio. R.. ``A soft robot that adapts to environments through shape change,'' Nature Machine Intelligence, vol. 3, no. 1, pp. 51-59, 2021.
\textit{*These authors contributed equally}
}

\begin{acknowledgements}
Thanks to my advisor Josh Bongard, my exemplary mentor and teacher, and to the other member of my committee Mathew Weston, Safwan Wshah, and Joe Near for their support. 
Thanks to my family who make all of my work worth it, and more especially my wife Nina, for her constant love and unwavering support.
\end{acknowledgements}

\tableofcontents
\newpage

\listoffigures
\newpage

\listoftables
\newpage

\doublespacing
\pagenumbering{arabic}

\chapter{Introduction}
Current machine learning algorithms should not be underestimated; they have drastically changed the research landscape in various fields such as vision~\cite{kolesnikov2019big, cohen2018spherical, zhang2018adversarial, brock2018large}, natural language processing~\cite{dai2019transformer, zhang2019bridging, devlin2018bert}, and control~\cite{vinyals2019grandmaster, akkaya2019solving, schrittwieser2020mastering}. While neural networks and deep learning (the techniques used in these advancements) are behind this progress, most of these techniques were developed in the past and have only now come to fruition. The perception, which is a simple binary model of a biological neuron was developed as early as 1957~\cite{rosenblatt1958perceptron}, from there research preceded to sigmoid neurons (a continuous value, non-linear model of biological neurons) and back propagation~\cite{kelley1960gradient, dreyfus1962numerical, werbos1988generalization, rumelhart1986learning}. Eventually, work was also done to improve neural architectures by creating neural networks with many layers and introducing new types of neural network layers such as convolutions layers (principally used in vision research today)~\cite{fukushima1988neocognitron, le1989handwritten} and recurrent network layers (principally used in natural language processing). Only recently have these neural techniques supplanted traditional state of the art methods, largely due to advancements in computer hardware and data availability~\cite{goodfellow2016deep}. This brief history of deep learning and neural networks also highlights that the key focus of research in the past and present is neurocentric: focused on improving neuron models, neural architectures, and training algorithms.

Today, neural-network-based machine learning continues to set new benchmarks and achieve superhuman performance in many tasks, and the spread of domains subject to such advancements continues to broaden. However, one area where machine learning adoption has been slow, and in some cases non-existent, are domains which require multitask learning and/or multiple behaviors~\cite{goodfellow2013empirical}; this is especially true in robotics.

\begin{figure*}[t]
    \centering
    (A) Catastrophic Forgetting \hfill (B) Catastrophic Interference\\
    \vspace{17pt}
    \includegraphics[width=0.46\linewidth]{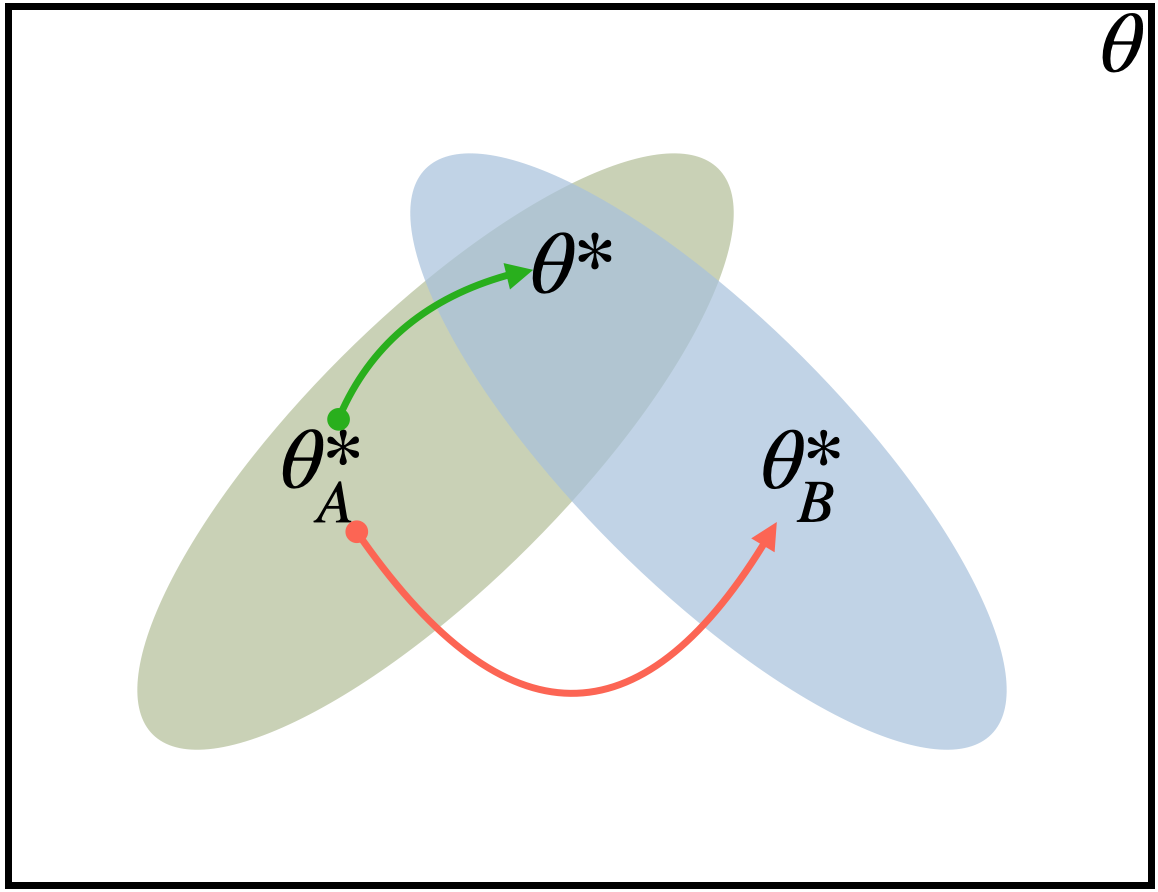}\hfill
    \includegraphics[width=0.46\linewidth]{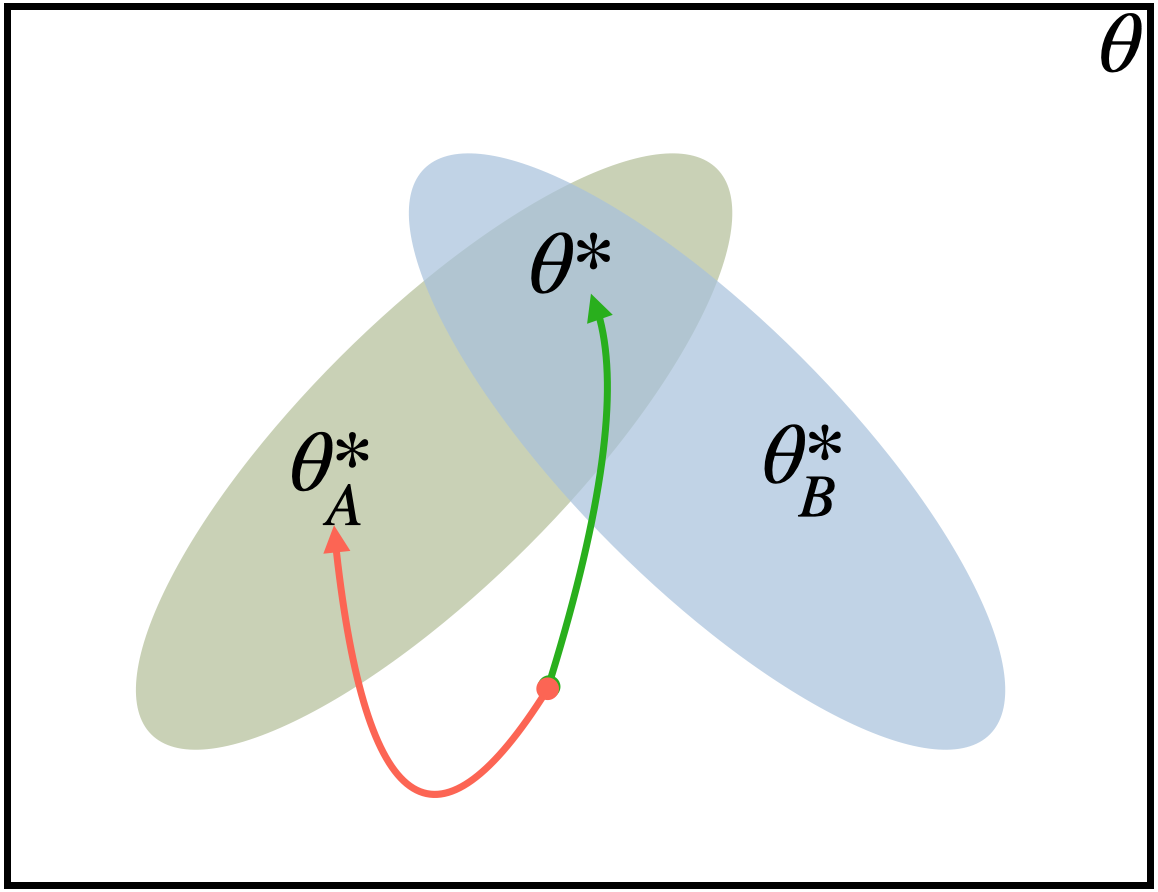}
    \vspace{15pt}
    \caption{\textbf{(A)} a representation of the occurrence of catastrophic forgetting during training: task A is learned first and then when task B is learned the network ends in a local optima that no longer solves task A. \textbf{(B)} a representation of the occurrence of catastrophic interference during training: during simultaneous training on both tasks the robot falls into a local optima that solves only one task. \textbf{(Both)} $\theta$: A two dimensional representation of the complete parameter space of a neural networks weights. $\theta^*_A$: The network weights that produce optimal performance on a given task A. $\theta^*_B$: The network weights that produce optimal performance on a given task B. $\theta^*$: The set of network weights that allow a single neural network to perform optimally on all tasks. The red trajectory shows the actual trajectory that occurs when a network experiences a catastrophic failure during training, while the green arrows show the desired training trajectory. In both cases the failure occurs because the network find a local optima that is not in $\theta^*$. This figure has been adapted from~\cite{kirkpatrick2017overcoming}.}
    \label{fig:catastrophic}
\end{figure*}

\begin{figure*}[t]
    \centering
    \includegraphics[width=0.9\linewidth]{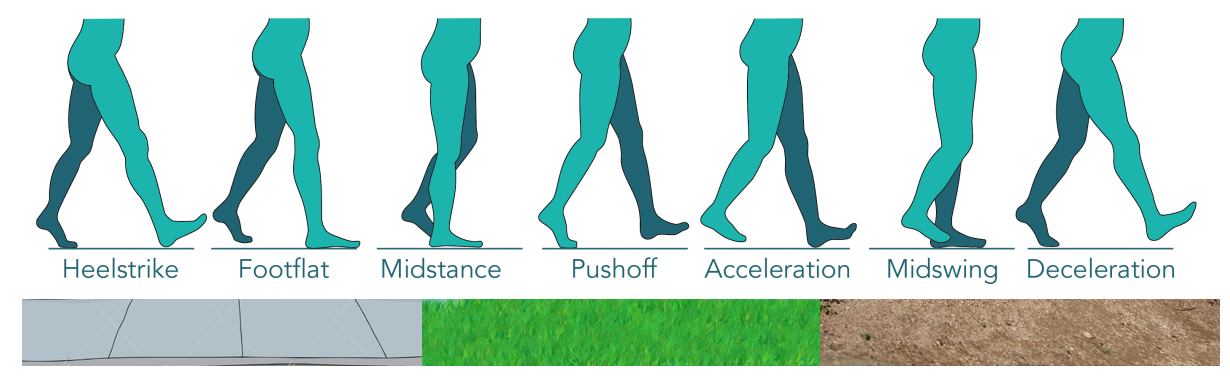}
    \vspace{15pt}
    \caption{Due in part to the design of the muscular and skeletal systems, a human performs a similar gait pattern across various terrains. However, current bi-pedal rigid robots must adapt their gait for even the smallest of terrain changes. In essence the human form allows for the training of a single walking strategy for a variety of circumstances and thus reduces the required computational effort (compared to a robot) to discover the function for walking in multiple domains.}
    \label{fig:good_design}
\end{figure*}

\section{The Problem}
Given the current computational capabilities, robots in simulation can often be trained to perform a single task very well, but when that same robot is trained to perform a new task, the robot performs increasingly poor on the former task, or more often, forgets its prior training completely~\cite{mccloskey1989catastrophic}. A robot may also be trained on multiple desired tasks simultaneously in which case training steps that results in improvement in one task often cause greater reduction in the others resulting in a robot that learns to specialize in only one of the tasks. These crippling phenomenon are known as catastrophic forgetting and catastrophic interference(see Fig.~\ref{fig:catastrophic}). Thus, currently we see industrial and consumer robots performing only specialized tasks. While there have been many proposed methods for attempting to overcome these problems, few have been successful beyond a few tasks~\cite{goodfellow2013empirical, kirkpatrick2017overcoming} and often perform much worse than specialist robots on the same tasks. Furthermore, training such neural networks requires impressive amounts of computational costs of up to \$12,976,128~\cite{ken2020, vinyals2019grandmaster} for the training of a single agent on a single task. Thus, current strategies of machine learning also lack the ability to scale down to the computational restrictions on common robot platforms.

As in other fields where neural networks have been successful, most research on methods to combat catastrophic interference are neurocentric: focused on the development of better machine learning algorithms and improvements to neural architectures. One line of research attempts to link individual specialist networks in an effort to produce a generalist whole~\cite{ans1997avoiding, coop2013ensemble}. Such methodologies often suffer from requiring extremely complex networks, massive amounts of computing power, and excessively long training periods, while being very sensitive to small changes in algorithmic parameters~\cite{coop2013ensemble}. Other methods attempt to hide the problem by producing robots that learn new or forgotten tasks rapidly enough that they appear to know how to perform all tasks simultaneously~\cite{finn2017model}; and one of the most famous methods freezes sections of the neural network when training on a new task to prevent the overwriting of previously learned information~\cite{kirkpatrick2017overcoming}. However, these methods similarly suffer from requiring a robot to learn a large amount of prior information, besides being even more complex and computing intensive.

Much of the motivation for these neurocentric based resolutions stems from the \emph{Universal Approximation Theorem}~\cite{cybenko1989approximation}, that shows that a neural network is capable of approximating any function, assuming that the function exists. For example, if there exists a function that can turn written English words into spoken English sounds (which we assume exists since humans do this every day), then a neural network is capable of approximating that function if we can just find the correct parameters. In robotics the functions we approximate generally involve taking sensor data (sight, touch, sounds) and converting them to robotic motor commands that achieve a task. For instance, we may design a bipedal robot that needs to learn a function to identify a specific object and then output motor commands to walk toward it. We first assume that such a function exists (which is likely since humans can perform this task) and then train a network to solve the problem. However, if we expand the problem such that the robot must perform this task over multiple terrains, this could prove quite difficult and is even a challenge for the most sophisticated bipedal robots today even using the best neural network techniques. One way humans may overcome this problem is possibly due to the design of the muscular and skeletal systems (see Fig.~\ref{fig:good_design}). The human foot, unlike a rigid robot foot, is complaint against an uneven surface and adapts such that walking on rough terrain is essentially the same as walking on smooth terrain ~\cite{kent2019changes}. Thus, humans use a similar gait pattern across various terrains, where current bipedal rigid robots must adapt their gait for even the smallest of terrain changes. The human form allows for the training of a single walking strategy for a variety of circumstances and thus reduces the required computational effort (compared to a robot) to discover the function for walking in multiple domains. Thus in robotics, the robot design in which a neural network is embodied, is likely to have a measurable impact on the ability to approximate the function required to complete the desired task. Research that is focused on the synergy between AI and robotic design often falls in the category of evolutionary robotics.

\section{Evolutionary Robotics}
Evolutionary robotics is an embodied approach to AI in which both a robots' design and controller are often optimized simultaneously using an evolutionary algorithm~\cite{bongard2013evolutionary}. This is done by optimizing the parameters of a robot against a behavioral goal (e.g. move in a direction). The parameters of the robot usually include the weights of a neural network as well as parameters that define aspects of the robot's design such as physical structure, sensor number and placement, motor strength, etc. Thus, research in this field has often incorporated methods beyond the architecture and training of neural networks to overcome the hurtles faced by traditional machine learning approaches~\cite{bongard2013evolutionary}. 

The first artificial evolution of a robots design and controller was done in 1994 by Karl Sims~\cite{sims1994evolving}, creating simulated robots that displayed lifelike forms and behaviors. This has since inspired others to utilize the principles of embodiment to build upon his work such as \cite{lipson2000automatic} which optimized both the body and control of robot that could then be 3D printed, and \cite{cheney2018scalable} who used these principles to design simulated soft robots. However, all of these experiments focused solely on a simple locomotion task. They also focused mainly on developing new methods in the evolution of body design rather than looking at the effects of those designs on problems typically faced in neural control. An example of recent work that does apply these principles to the domain of machine learning is \cite{cappelle2017reducing} which evolved robot designs that were modular to varying environments with a net effect of reducing the overall training time required by the neural network.

The principles of evolutionary robotics are considered most beneficial when we encounter initially unintuitive problems. In this case, using evolutionary algorithms can give new insight to a problem as they tend to ignore human bias, thus generating unique solutions that otherwise may typically be disregarded\cite{hornby2006automated, cariani1993evolve}. While, this evolutionary process almost always takes place in simulation, this knowledge can then be used to increase our capacity to build real robots with more desirable characteristics. As mentioned, identifying scalable methods for creating more general, multi-task robots is currently one such unintuitive space and is thus the task to which evolutionary robotics is applied in this dissertation.
\newpage

\begin{figure}[ht!]
    \centering
    \includegraphics[width=0.5\linewidth]{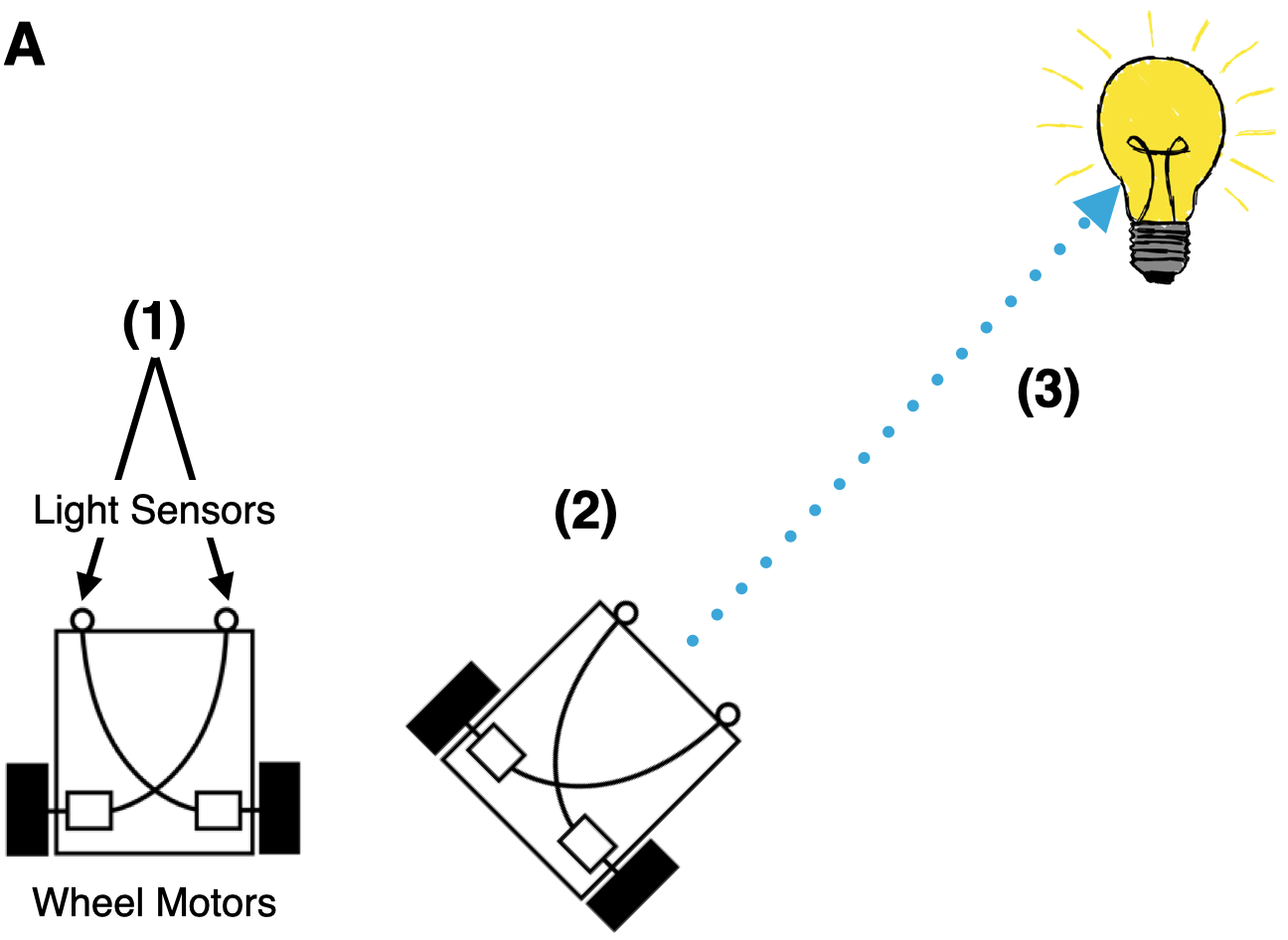}\\
    \vspace{50pt}
    \includegraphics[width=0.9\linewidth]{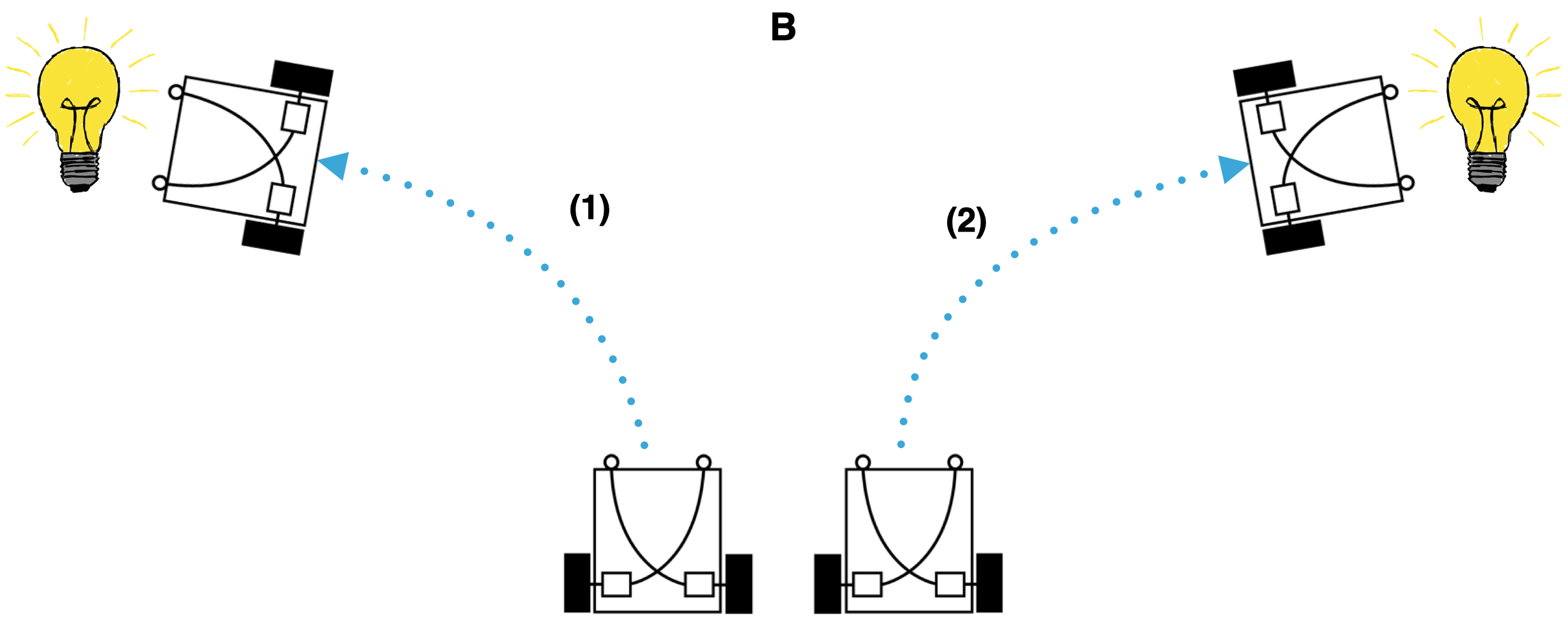}
    \vspace{15pt}
    \caption{Ways in which the single task of phototaxis (locomotion toward a light source) may be broken down into multiple tasks. \textbf{(A)} We can break phototaxis down into it's parts such as (1) identifying the direction of the light, (2) orienting towards the light (3) locomotion toward the light. \textbf{(B)} We can break the phototaxis task down by environment. (1) locomote toward the light when it is on the left (2) locomote toward the light when it is on the right. In this dissertation I use the latter method. In the context of catastrophic interference, this means that a robot that is simultaneously trained to walk toward a light source at at multiple different locations but fails to learn to walk toward a light source in at least one of the locations would be subject to this catastrophic interference}
    \label{fig:task_description}
\end{figure}
\newpage

\section{Description of Tasks and Environments}
Often in robotics we describe a specific task (e.g., locomotion) or environment (e.g., flat ground, incline ground) in which we would like our robot to be successful, and the description is mostly accepted. However, when we move to multiple tasks or environments there are often disputations as to what constitutes a single task or environment. For example in this dissertation, much of the work is explored under phototaxis, that is, getting a robot to locomote toward a light source. This could be considered as a single task in and of itself, or could be broken down as multiple tasks in different ways as shown in Fig.~\ref{fig:task_description}. One way is to break the task down into its parts, such as, identifying the direction of light and locomotion in a given direction. Another way would be to consider each different location of the light source in relationship to the robot to be a separate task (e.g., move toward the light when it is on the left, move toward the light when it is on the right). In both cases, the combination of these sub tasks would then constitute a single larger task of general phototaxis. Thus, it is not uncommon for researchers in the field to disagree on what is a multi-task robot depending on how one defines a task.

In this dissertation I consider each placement of the light to be a different task. In the context of catastrophic interference, this means that a robot that is simultaneously trained to walk toward a light source at at multiple different locations but fails to learn to walk toward a light source in at least one of the locations would be subject to this catastrophic interference. While there may be neural centric methods to make robots overcome this particular problem instance and perform general phototaxis, in this dissertation, I reduce the learning method and network size of the robots such that they do experience catastrophic interference. The focus of this dissertation is on alternate embodied methods to overcome this phenomenon that can synergize rather than compete with traditional methods on this problem. Thus, the description of phototaxis as a multitask problem is justified in this sense.

\begin{figure*}[ht!]
    \centering
    \includegraphics[width=0.6\linewidth]{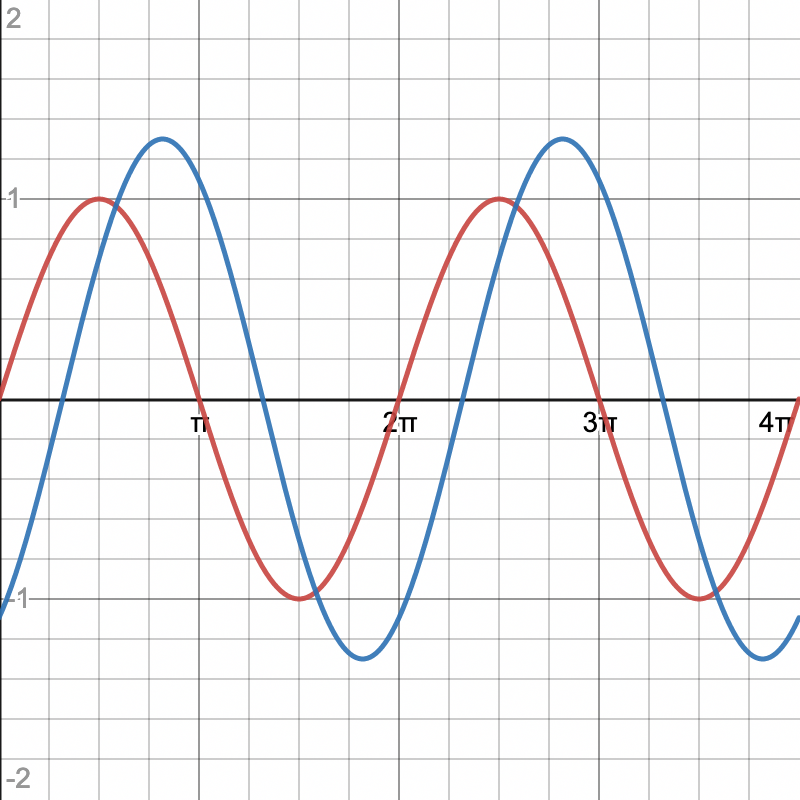}
    \vspace{15pt}
    \caption{Two sin waves with the same frequency but different amplitude and phase; if the waves both represent the values of the sensor of a robot over time, the robot would be considered homeostatic in relation to this sensor. The chief concern for homeostasis in this dissertation is thus overall signal pattern.}
    \label{fig:homeostatis}
\end{figure*}

\section{Homeostasis}
Throughout this dissertation, I will create metrics to explain how certain robot designs affect the loss landscape of neural control policies associated with that design. I also develop a theory to explain why those designs have certain effects on that landscape. The main theory brought forth is that robots that are resistant to catastrophic interference exhibit designs that allow for sensor homeostasis.

Homeostasis as a principle component of intelligence has a rich history in the literature and is built principally on the ideas presented English Cyberneticist and Psychiatrist Ashby~\cite{vargas2009homeostasis}. For Ashby, homeostasis is a base principle in intelligence and the ability of a living or artificial system to adapt to a continuously changing environment~\cite{ashby2013design}, specifically he designed a famous system called the homeostat to demonstrate these principles. Later, Dyke and Harvey would conclude that understanding the mechanisms of homeostasis is necessary for understanding real or artificial life~\cite{dyke2005hysteresis, dyke2006pushing}. In all this work, researchers found that intelligent systems are created with, among other goals, a desire to maintain a consistent internal state. These ideas have been used to motivate the creation of neural networks capable of dynamically changing their connections by means of plasticity rules~\cite{di2000homeostatic,hoinville2004comparative}. However, this work did not extend these ideas to the body design of robots used in these studies.

In the context of phototaxis, homeostasis means that as a robot locomotes toward light sources placed at different locations, the time series data of its light sensor values rapidly converges to the same values across these trials. That is, the robot ``sees'' these seemingly different environments as the same. This description of homeostasis can be seen in Fig.~\ref{fig:homeostatis}. Internal homeostasis has been postulated as a possible biological evolutionary goal~\cite{vargas2009homeostasis,torday2015homeostasis} and has seen considerable focus as a reward method for improving the behaviors in reinforcement learning ~\cite{keramati2014collecting, morville2018homeostatic} as well. In a previous example, we described how humans maintain a similar gait across a variety of terrains. In this sense, each terrain is internally experienced in much the same way. Thus, according to the previous work cited~\cite{ashby2013design,kent2019changes,torday2015homeostasis}, some evolutionary pressure may have favored functional anatomy in the human foot that mechanically ``erased'' slight differences in terrain, thereby obviating the need for neural adaptation of gait over such terrains. We can see many such homeostatic patterns throughout biology as various levels: many animals build structures (beaver damn, human home, etc) to provide consistent environmental conditions, and single cell organisms clump together in colonies such that internal cells experience minimal variation in their environment~\cite{levin2019planarian}. However, work to date has not used homeostasis as a tool for optimization and analysis in the context of catastrophic interference. Thus, In this dissertation we will focus on measuring sensor homeostasis as an analysis tool to describe robot behavior and designs. Doing so will allow for the future use of homeostasis as a fitness objective which may guide the automatic design of robot structures that are both performant and resistant to catastrophic interference. 

\section{Contribution Outline}
In my work, I have taken a radically different morphological approach to the multitask learning problems facing neural networks in robotics. Rather than work on improving neural training algorithms or architectures, my research focuses on the how different aspects of a robot's design can facilitate faster training and the likely hood of achieving multitask behavior. Overall this dissertation introduces novel metrics that quantify why certain designs improve neural performance and how these principles can be applied to the design of other robots.

Chapter 2 demonstrates the learning differences between three equally capable robot morphologies in phototaxis. Additionally, I introduce novel metrics to quantify catastrophic interference and definitively show that morphology is the principle factor in determining the amount of interference a robot experiences during training. This chapter, however, only provides speculation as to why but lays the foundation for embodiment as an important aspect of the catastrophic interference problem. Consequently, Chapter 3 develops a theoretical foundation for measuring a robot design's sample efficiency ($M_{L}$) and resistance to catastrophic interference ($M_{CF}$) that also serve as an explanation for why embodiment is such an important component of neural training. Chapter 4 tests the theoretical work from Chapter 3 across various robot designs and training algorithms. I show that $M_{L}$ and $M_{CF}$ are predicative of a design's real world performance. Furthermore, I demonstrate that optimizing for the design of variations in a robot's body along with a robot's controller leads to the creation of more general purpose robots that have the added benefit of requiring reduced training times. In this chapter I also show that optimizations naturally follow a homeostatic design gradient, that is, the robotic designs are more homeostatic in later generations than in earlier ones. This last finding provides an explanation for the improved performance of certain robot designs. By focusing on advancements in robot design rather than neural controllers, my work is synergistic rather than competitive with traditional machine learning research in this area. Thus, this work intimates that algorithms which automatically detect and exploit complementary features of the brain and the body will enhance current state-of-the-art techniques in the field.

While currently most of my work is conducted in simulation, I have also begun to validate some of this work in physical systems. Thus in Chapter 5, I showcase a simulated soft robot that can change its shape (or design) to operate in different environments. Using an evolutionary algorithm, the shapes and controllers for the different environments are automatically determined to allow for successful locomotion.

\chapter{The effects of morphology and fitness on catastrophic interference}
\section{Abstract}
Catastrophic interference occurs when an agent improves in one training instance but becomes worse in other instances. Many methods intended to combat interference have been reported in the literature that modulate properties of a neural controller, such as synaptic plasticity or modularity. Here, we demonstrate that adjustments to the body of the agent, or the way its performance is measured, can also reduce catastrophic interference without requiring changes to the controller. Additionally, we introduce new metrics to quantify catastrophic interference. We do not show that our approach outperforms others on benchmark tests. Instead, by more precisely measuring interactions between morphology, fitness, and interference, we demonstrate that embodiment is an important aspect of this problem. Furthermore, considerations into morphology and fitness can combine with, rather than compete with, existing methods for combating catastrophic interference.

%%%%%%%%%%%%%%%%%%%%%%%%%%%%%%%%%%%%%%%%%%%%%%%%%%%%%%%%%%%%%%%%%%%%%%%%%%%%%%

\section{Introduction}
Currently, a popular method for realizing intelligent machines is to optimize the parameters of fixed-architecture deep neural networks \cite{lecun2015deep}. However, increasing interest is coming to bear on optimizing the cognitive architecture of such networks as well \cite{miikkulainen2017evolving}. Indeed, investigations into the evolution of cognitive architectures has long been a target of study \cite{gruau1993adding, bongard2001repeated, stanley2002evolving} in the evolutionary computation community.

It follows from this, if dealing with robots, that optimizing body plan influences sensory repercussions of action, which in turn will influence which cognitive architecture will facilitate learning for a given task. To begin investigations into this last observation, here we investigate how the choice of robot morphology and fitness affect one specific aspect of neural networks: their ability to resist catastrophic interference. We employ an evolutionary robotics approach to investigate this question.

\subsection{Evolutionary robotics}
Since its beginnings, many investigators in the field of evolutionary robotics \cite{floreano1994automatic, harvey1997evolutionary, Bongard13} have used evolutionary algorithms to optimize both the body plan and neural controllers of robots \cite{Sims94, Lipson00, cheney2013unshackling}. Here we show that indeed the choice of body plan can influence the efficacy of training neural controllers: some body plans enable greater resistance to catastrophic interference.

\begin{figure*}[t]
(a)\includegraphics[height=0.14\textheight]{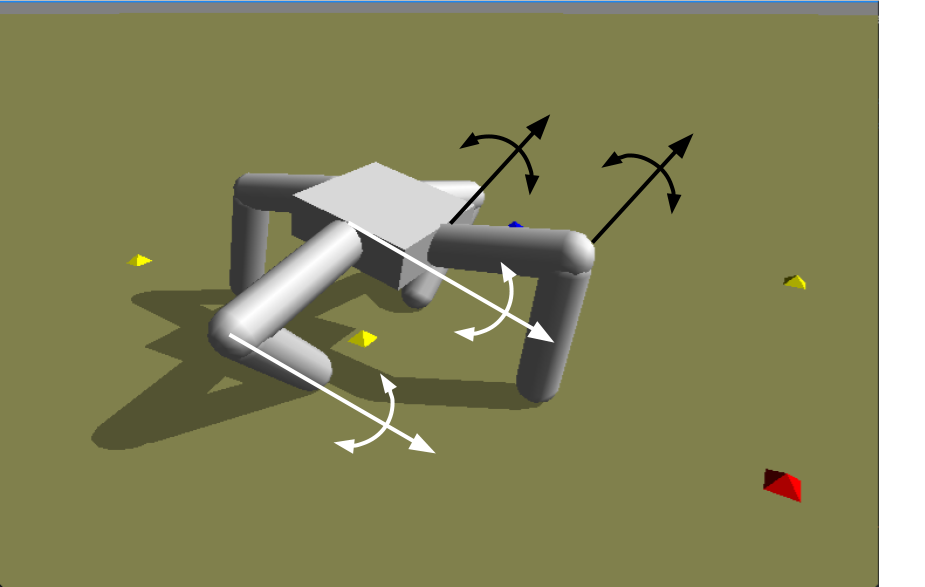}
(b)\includegraphics[height=0.14\textheight]{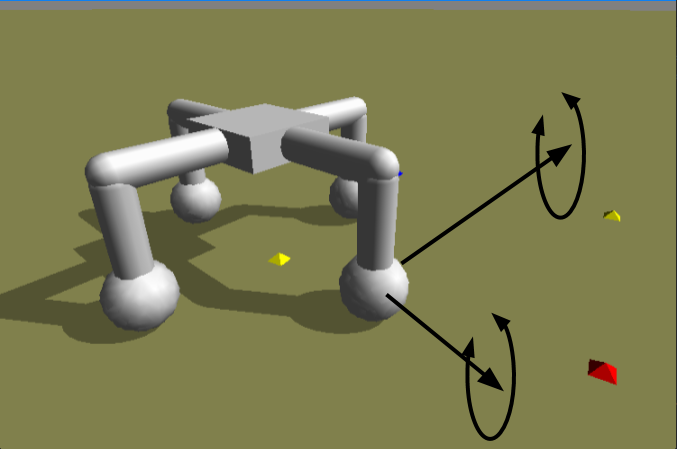}
(c)\includegraphics[height=0.14\textheight]{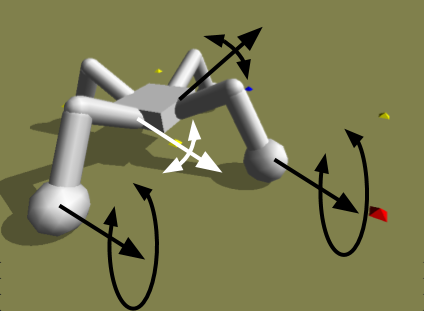}
\caption{
Three classes of phototaxic robots---(a) legged, (b) wheeled, and (c) whegged---and their environments were simulated using Pyrosim (\href{https://ccappelle.github.io/pyrosim/}{ccappelle.github.io/pyrosim}). Each robot has eight degrees of freedom, as depicted by the black and white arrows which indicate the axis (straight) and direction (curved) of rotation for a particular hinge-joint (a, c) or wheel (b, c).
}
\label{fig:robots}
\end{figure*}

\subsection{Catastrophic interference}
It has been acknowledged since the early days of neural network research that catastrophic interference \cite{mccloskey1989catastrophic}, also known as catastrophic forgetting \cite{french1999catastrophic, goodfellow2013empirical}, is a major challenge to training them effectively. Even in the most common forms of network training such as the backpropagation of error, there is no guarantee that reducing the network's error on the current training sample does not increase error on the other training samples.

For these reasons, much effort has been expended to combat this challenge. One family of solutions involves constructing modular networks \cite{lipson2002origin, ellefsen2015neural, clune2013evolutionary, espinosa2010specialization, kashtan2005spontaneous, sabour2017dynamic} in which different modules deal with different subsets of the training set. In such networks, changes to one module may result in improved performance for the training subset associated with that module without disrupting performance on other subsets. Such modularity has indeed been demonstrated to minimize catastrophic interference \cite{ellefsen2014neural, rusu2016progressive, lee2016dual, fernando2017pathnet}.

Related to this concept of modularity are networks in which some subsets of the network that have a large impact on the current training set are made less resistant to change during subsequent training \cite{kirkpatrick2017overcoming, velez2017diffusion}. The remaining parts of the network remain adaptive, and thus able to deal with new training instances without disrupting behavior on previous instances.

The drawback of these approaches however is that network size tends to increase with the amount of training data, because new modules must be implicitly or explicitly added for new training data.

Another guard against catastrophic interference is to reduce the magnitude of behavioral impact after some change is made during training. The intuition here is that small changes to network behavior may increase the likelihood of local improvements for new training instances while minimizing or nullifying performance decreases on the previous training set.

In evolutionary methods, one way of reducing behavioral impacts is to dynamically tune mutation rates \cite{dang2016self} and/or crossover events \cite{teo2016fixed}. A recent approach demonstrated for neuroevolution is to dynamically tune individual synaptic weights proportionally to their impact on the network's behavior \cite{lehman2017safe}. In the genetic programming community, semantic variation operators have been reported \cite{Vanneschi_2014_A-Survey, szubert2016semantic}. These operators take into account the semantics of subtrees or individual tree nodes, and attempt to replace them with new genetic material that exhibits similar semantics.

In this work we show that the body plan itself as well as the manner in which fitness is defined can buffer the behavioral impact of mutations such that the embodied agent's behavior can improve in one environment without adversely impacting its performance in another environment in which it is already proficient without using any methods for specifically preventing interference in the controller.

\subsection{Embodied impacts on neural properties}
Besides catastrophic interference, it has been shown elsewhere that embodiment can influence the positive or negative aspects of neural networks. For instance, work in morphological computation has shown that a good choice of morphology can allow for simplified neural networks (e.g.~\cite{hauser2011towards}).

Morphology may also render a robot more robust to external environmental perturbation \cite{Bongard11} or internal changes to the neural controller \cite{kriegman2017morphological}. Some body plans also facilitate or obstruct the discovery of modular neural networks \cite{bongard2015evolving, bernatskiychoice}. In this work we introduce a heretofore unexplored aspect of the interaction between body plan and neural control of embodied agents: how the choice of body plan may render the neural controller more or less resistant to catastrophic interference.

The next section describes our methodology for demonstrating this phenomenon; the following sections provide results from evolving neural controllers in different body plans; the final sections provide some discussion as to how and why this phenomenon arises and concludes with avenues for future study.

%%%%%%%%%%%%%%%%%%%%%%%%%%%%%%%%%%%%%%%%%%%%%%%%%%%%%%%%%%%%%%%%%%%%%%%%%%%

\section{Methods}
\subsection{The robots}
Three types of robots were used in this study (Fig. \ref{fig:robots}). All are variations on a standard radially symmetric quadrupedal form used in other evolutionary robotics studies \cite{bongard2006resilient, lohmann2012aracna, belter2015affordable}. The robots differ in their differential use of legs and wheels.\footnote{\href{https://github.com/jpp46/ALIFE2018}{github.com/jpp46/ALIFE2018} contains the source code necessary for reproducing the results reported in this here.} Combining wheels and legs in different ways is itself an active area of study in robotics \cite{schroer2004comparing, jehanno2014design, kim2014wheel}.

\subsubsection{The legged robot}
This robot consists of a body and four legs attached to the body by a joint. Each joint rotates $180 \degree$ through the plane defined by the two cylinders comprising that leg. Each leg consists of two limbs bent at $90\degree$ attached by a joint which also rotates $180\degree$ through the plane defined by the upper and lower legs (Fig. \ref{fig:robots}a). Each of the resulting eight joints are actuated using position control: a value arriving from the motor neuron attached to the joint is treated as a desired angle. The length of each upper leg, lower leg, and the two sides of the main body are 0.3 units of length long (the physics engine is agnostic to the length scale).

\subsubsection{The wheeled robot}
This robot freezes the previously mentioned joints at their initial angles and cannot change these positions. In addition, each leg has a wheel attached on the end that rotates through the plane defined by its upper and lower leg components. These wheels are attached to the legs by axles which rotate through the sagittal plane (Fig. ~\ref{fig:robots}b). This allows the wheels to change two directions of their rotation, acting as caster wheels. Values received at each of the two motors controlling the wheel treat the incoming values as desired angular velocity. In the spirit of keeping the robots as similar as possible, the motor neurons innervate the wheels output as desired angular velocities in $[-90\degree/s,+90\degree/s]$ to match the legged robot where motor neurons dictate desired angles in $[-90\degree,+90\degree]$.

\subsubsection{The whegged robot}
This robot combines features of the previous two robots. In this robot, only the joints connecting the two leg limbs are frozen and can not change from their initial $90\degree$ angle. The joints joining the legs to the body are the same as the first robot: position controlled, actuated one-degree-of-freedom rotational joints. The robot also has wheels on the legs, but they do not act as caster wheels (Fig. ~\ref{fig:robots}c). This robot combines position control at the four shoulder joints and velocity control at the four wheels.

\subsubsection{The sensors} Each robot contains a light sensor $\ell$ that responds to light as a float value according to the inverse square law for light propagation:
$\ell = 1/d^2$, where $d$ is the distance from a light source. Occlusion is not simulated in the light sensor: if an object is between the sensor and the light, there is no change in sensor value. Each robot contains a single binary touch sensor in each leg. These four sensors read $+1$ when in contact with the ground, and $-1$ otherwise.

\subsection{The controllers}
The controller for each robot is a neural net with $5$ input neurons, one for each sensor, and $8$ output neurons, one for each motor. The neural net has no hidden layers and is fully connected. The connection weights are captured by a $5 \times 8$ matrix, which also represents the genome of a robot. The weights of the matrix are constrained to values in the interval $[-1, 1]$. The update function for a neuron during simulation follows this function:
\begin{equation}\label{neuron_update}
m_i^{(t)} = \tanh \left[ m_i^{(t-1)} + \tau_i \sum_{s=1}^{5} w_{s_i} m_s^{(t)} \right]
\end{equation}
where $m_i^{(t)}$ denotes the value of the $i^{th}$ motor neuron at the current
time step, $m_i^{(t-1)}$ is a momentum term that guards against `jitter' (high-speed and continuous reversals in the angular velocity of a joint), $\tau_i$ is a time constant that can strengthen or weaken the influence of sensation on the $i^{th}$ motor neuron relative to its momentum, and $w_{s_i}$ is the weight of the synapse connecting the $s^{th}$ sensor neuron to the $i^{th}$ motor neuron. In order to ensure that random controllers produce diverse yet not overly-energetic motion, all $\tau_i$ were set to 0.3 via empirical investigation.

\subsection{The task environment}
Robots are evolved to perform phototaxis: minimizing the distance between themselves and a light source in their environment. Each robot is exposed to all training environments. Each environment consists of a light source that is 30 body lengths away from the robot (equivalent to nine units of distance). The robot starts at coordinate space $(0, 0)$ and the light sources are placed at coordinate space $(0, 9)$, $(0, -9)$, $(9, 0)$, $(-9, 0)$ for environments $1$, $2$, $3$, and $4$ respectively. The robots are evaluated for $1000$ time steps in each environment using a fixed time step of 0.05 seconds.

\subsection{The fitness functions}
The fitness function applied to a single environment is the value of the robot's light sensor at the end of an evaluation period ($t=1000$). We combine fitness values drawn from multiple environments in three different ways:
\vspace{-0.5em}
\begin{equation}
\label{fitnesses}
F_{\Sigma} \doteq  \sum_{i = 1}^{n} f_i , \qquad
F_{\Pi} \doteq  \prod_{i = 1}^{n} f_i , \qquad
F_{\min} \doteq  \min_{i = 1}^{n} f_i .
\end{equation}
where $f_i$ is the individual's fitness in environment $i\in(1,n)$.

\subsection{The evolutionary algorithm}
We employed a simple parallel hill climber to nine different experimental treatments as shown:
\begin{center}
\begin{tabular}{l c c c}
\toprule
{}               & \textbf{Legged} & \textbf{Wheeled} & \textbf{Whegged} \\
\midrule
{$\sum$}  & Treatment 1     & Treatment 2      & Treatment 3 \\
{$\prod$} & Treatment 4     & Treatment 5      & Treatment 6 \\
{$\min$}  & Treatment 7     & Treatment 8      & Treatment 9 \\
\bottomrule
\end{tabular}
\end{center}
For each treatment, 30 independent evolutionary runs were conducted each with a population size of $S = 100$. Each individual $p$ was encoded as a $5 \times 8$ matrix of synaptic weights. Thus population $P$ is represented by a $100 \times 5 \times 8$ tensor. We used a mutation strength of $m = 0.05$, and conducted each run for $G = 3000$ generations. At each generation $g$ we took the current population $P_{g}$ and generated a new population $P_{(g+1)}$ by mutation such that:
\begin{equation}
\label{mutation}
P_{(g + 1)} = \mathcal{N}\left(P_{g}, m\right)
\end{equation}
Where $\mathcal{N}\left(\mu, \sigma \right)$ is the standard normal distribution.
We then updated the new population using:
\begin{equation}\label{survival}
p_{g+1} =
\begin{cases}
p_{g+1} , \;  F \left( p_{g+1} \right) > F \left( p_{g} \right)\\
p_{g} , \; \text{otherwise}
\end{cases}
\end{equation}
where $p$ denotes an individual in the population at generation $g$ and the fitness function $F$ is determined by the treatment as 
defined in Eq. \ref{fitnesses}. As shown, each child in the evolutionary algorithm only competes with its direct parent, creating $S$ individual climbers.

\subsection{Measuring catastrophic interference}
\begin{figure}[t]
\vspace{-1.4em}
\centering
\includegraphics[width=0.4\textwidth]{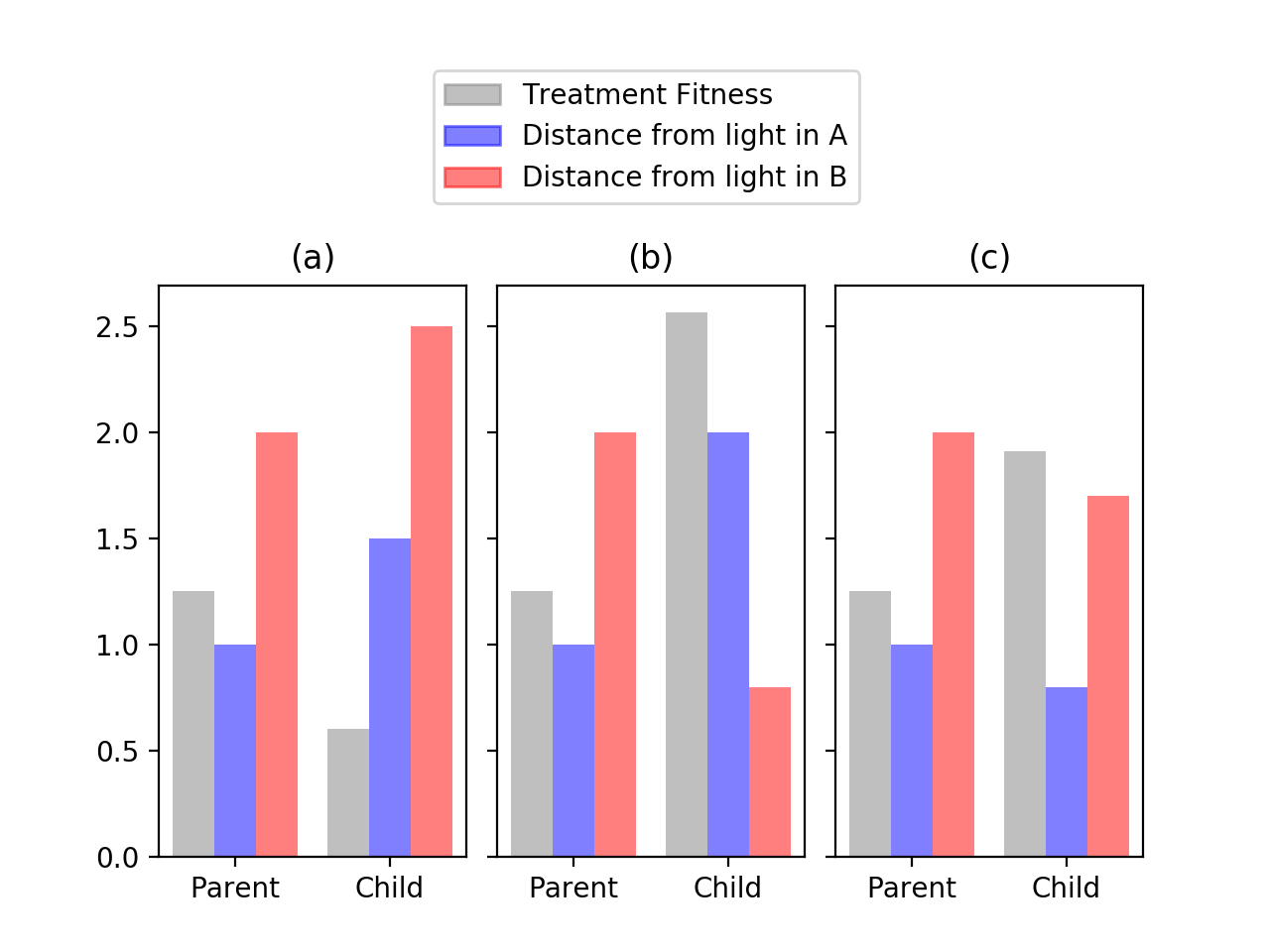}
\vspace{-1em}
\caption{
Example mutations that
(a) are deleterious, (b) result in catastrophic forgetting, and (c) avoid catastrophic forgetting. 
The smaller the distance from the light source (blue, red), the higher the fitness (inverse square law).
}
\label{fig:metrics}
\end{figure}

Catastrophic interference, in its simplest formulation, occurs when an improvement in one environment incurs reduced performance in one or more other environments. In an evolutionary setting, catastrophic interference can be measured at the highest temporal resolution by considering mutations: the change in performance between a parent and child for each environment experienced by both agents (Fig. \ref{fig:metrics}).

For the purposes of analysis, we only track performance changes between mutations where overall fitness increased (we do not record deleterious or stagnant mutations). Also we only record and perform analysis on the change in Euclidean distance of the robot from the light source (blue and red bars in Fig. \ref{fig:metrics}). This provides a more intuitive understanding of changes in performance and allows comparison between any 2 treatments regardless of the fitness function used in that treatment.

We define a function $D$ on an individual such that it returns a vector $[x_1, x_2,...,x_n]$, where $x_i$ is the distance from the light source at the end of simulation in environment $i$. As shown in Fig. ~\ref{fig:metrics}, fitness and distance are inversely correlated therefore we record change in distance after every successful mutation as:
\begin{equation}
\label{dist_record}
\Delta D = -\left[D\left( p_{(g+1)} \right) - D\left( p_g\right)\right].
\end{equation}
We negate this difference, so that a positive increase in fitness in an environment causes an increase in the corresponding component of $\Delta D$. We apply four metrics $(M_1, M_2, M_3, M_4)$ to this data ($\Delta D$) to measure the overall effect of catastrophic interference in each each of the treatments.

\subsubsection{$M_1$: Average Worst Absolute Distance}
For this metric, we take the distances from the light sources for the champion from each run, for each treatment. We here define a run champion such that:
\begin{eqnarray}
\label{runchampion}
C = \min_{P_g} \left[ \max D(p_{g}) \right],\; \text{where } \; g=3000.
\end{eqnarray}
Recall that distance is measured from the light source so higher is worse. Therefore a champion $C$ is the one with the lowest distance in it's worst environment. 
We compute $C$ for each run and take the average across all runs, for each treatment:
\begin{equation}
\label{m1}
M_1 = \frac{1}{N} \sum_{i=1}^{N = 30} C_i
\end{equation}
If this number is particularly high for a given treatment then it either failed to evolve robots that behaved well in at least one environment, a sign of catastrophic interference. If this number is sufficiently low for a given treatment, it was able to evolve agents that solved all environments, and at worst suffered mild amounts of interference.

\subsubsection{$M_2$: Average $\theta$ from $V$}
If a treatment is able to avoid evolving specialists by producing children that enjoy increases in fitness in all environments (Fig. \ref{fig:metrics}c), the most beneficial mutation possible would be represented by the vector $V = [x_1, x_2,...,x_n]$ where all elements are equal, $x_1 = x_2 = ... = x_n$. To compute $M_2$ we take all of the distance vectors from the mutations of a champion $C$ and record the cosine angle difference between $\Delta D$ and $V$. If the angle is less than or equal to $45 \degree$, then that vector, although perhaps biased slightly toward one or the other environment, nevertheless represents a mutation that avoided catastrophic interference. For this metric, we again record all such beneficial mutations within the lineage of a run that produced that run's champion. This can be represented as:
\begin{eqnarray}
\label{m2}
\theta' & = & \frac{1}{A_i} \sum_{a=1}^{A_i} \arccos \frac{\Delta D_a \cdot (V)}{|\Delta D_a| |V|} \\
\theta_i & = &
  \begin{cases} 
    \theta' : \theta'_i \leq 180 \degree \\
    |\theta' - 360 \degree| : \text{otherwise} \\
  \end{cases}\\
M_2 & = & \frac{1}{N} \sum_{i=1}^{N=30} \theta_i
\end{eqnarray}
where $A_i$ is the number of recorded $\Delta D$'s, $\theta'$ denotes the angle between the $\Delta D$ and $V$, and $\theta_i$ represents the average amount of catastrophic interference that individuals in the lineage of the champion from run $i$ experienced (lower $\theta_i$ represents less catastrophic interference). $M_2$ thus represents the average amount of catastrophic interference experienced by the run champion lineages in a treatment. If $M_2$ is low for a given treatment, then that treatment can be considered to be resistant to interference.

\subsubsection{$M_3$: Average $\Delta D$ Length}
Again we take the vector $\Delta D$ of a beneficial mutation that occurred within the ancestral lines of run champions. We then perform a similar method as in Eq. \ref{m2} on the length $|\Delta D|$ as shown:
\begin{eqnarray}
\label{m3}
L_i & = & \frac{1}{A_i} \sum_{a=1}^{A_i} |\Delta D_a| \\
M_3 & = & \frac{1}{N} \sum_{i=1}^{N=30} \Delta L_i
\end{eqnarray}
$M_3$ represents the average magnitude of improvement made during mutations by the run champion lineages in a treatment. This metric only makes sense in conjunction with $M_2$. Consider a treatment with small $M_2$ yet also small $M_3$. The treatment will suffer less from catastrophic interference, but is insignificant since the mutations yield insignificant improvements. 
However, if $M_2$ is small and $M_3$ is large for a given treatment, that treatment not only yielded mutations that avoided catastrophic interference but also exhibited high evolvability.

\subsubsection{$M_4$: Average Percentage of Points in Quadrant $I$}
Even though we attempt to measure catastrophic interference among just beneficial mutations, it is possible for a beneficial mutation to make a sufficiently large improvement in fitness in one environment even though there is some degradation in performance in the other environment (e.g.~Fig. \ref{fig:metrics}b). This equates to a $\Delta D$ where one of its elements is negative. When plotted as a point for the case of 2 environments the point falls in the upper left or lower right quadrant of a scatter plot. Thus, to buttress our measurement of catastrophic interference we devised a fourth metric, which is simply the fraction of beneficial mutations within run champions' ancestral lines where all elements of $\Delta D$ are positive. In the case of 2 environments they fall within the upper right quadrant (Quadrant $I$). This can be seen visually in Fig. \ref{fig:best}.

%%%%%%%%%%%%%%%%%%%%%%%%%%%%%%%%%%%%%%%%%%%%%%%%%%%%%%%%%%%%%%%%%%%%%%%%%%%%%%%%

\section{Results}
We analyze the performance and relative amounts of catastrophic interference for all nine treatments across $M_{1 - 4}$. 
We use the Mann-Whitney U test (with Bonferroni correction for eighteen comparisons) to indicate statistical significance at the $p = 0.05$ level.
The metrics generally show that there is an interaction between fitness function and morphology. If there was no interaction, an entire row (morphology does not matter) or an entire column (fitness function does not matter) would not be significantly different.

\begin{table}[h]
\centering
\caption{\label{table:M1} Mean of $M_1$ across treatments. 
Arrows indicate statistical significance between adjacent cells.
}
\resizebox{0.5\textwidth}{!}{\begin{tabular}{l c c c c c c}
\toprule
{$D$} & \textbf{Legged} && \textbf{Wheeled} && \textbf{Whegged} & \\
\midrule
{$\sum$}   & 8.181           && 8.386            &$\alr$& 5.915      &$\alr$ \\
{}         & $\aud$          && $\aud$           && $\aud$           & \\
{$\prod$}  & 6.530           &$\alr$& 3.593      &$\alr$& 1.234      &$\alr$ \\
{}         &                 &&                  &&                  & \\
{$\min$}   & 5.827           &$\alr$& 3.575      &$\alr$& 1.296      &$\alr$ \\
\\
\bottomrule
\end{tabular}}
\end{table}

\subsection{\texorpdfstring{$M_1$}{M1}: Average Worst Absolute Distance}
As can be seen in Table \ref{table:M1}, evolutionary performance generally improves moving from top to bottom row-wise and left to right column-wise. This metric shows that the whegged robot significantly outperformed the other two robots and that the min function is generally a better fitness function for this task. Interestingly, however, the treatments under the product and min fitness functions aren't significantly different. This suggests morphology might have a greater impact on this metric.

\begin{table}[h]
\centering
\caption{\label{table:M2} Mean of $M_2$ across treatments. 
Arrows indicate statistical significance between adjacent cells.
}
\resizebox{0.5\textwidth}{!}{\begin{tabular}{l c c c c c c}
\toprule
{$\theta$} & \textbf{Legged} && \textbf{Wheeled} && \textbf{Whegged} & \\
\midrule
{$\sum$}   & 71.035          && 73.362           && 72.640           & \\
{}         &                 &&                  &&                  &  \\
{$\prod$}  & 66.292          && 71.090           && 66.895           & \\
{}         & $\aud$          && $\aud$           && $\aud$           &  \\
{$\min$}   & 57.843          && 58.226           &$\alr$& 48.844     &$\alr$  \\
{}         & $\aud$          && $\aud$           && $\aud$           &  \\
\bottomrule
\end{tabular}}
\end{table}

\subsection{\texorpdfstring{$M_2$}{M2}: Average \texorpdfstring{$\theta$}{Theta} from \texorpdfstring{$V$}{V}}
We now wish to investigate whether the greater evolvability seen for the whegged robot under the product and min fitness functions is a result of those treatments being better able to resist catastrophic interference. Table \ref{table:M2} shows that the whegged robot with the min fitness function achieves beneficial mutations that yield improvements in both environments, or least only slight decreases in fitness in one of them evidenced by their higher relative proximity to $V$.

\begin{table}[h]
\centering
\caption{\label{table:M3} Mean of $M_3$ across treatments. 
Arrows indicate statistical significance between adjacent cells.
}
\resizebox{0.5\textwidth}{!}{\begin{tabular}{l c c c c c c}
\toprule
{$|\Delta D|$} & \textbf{Legged} && \textbf{Wheeled} && \textbf{Whegged} & \\
\midrule
{$\sum$}       & 1.734           &$\alr$& 3.209      && 4.971            &$\alr$ \\
{}             &                 &&                  &&                  & \\
{$\prod$}      & 1.889           &$\alr$& 3.372      &$\alr$& 5.341      &$\alr$ \\
{}             & $\aud$          &&                  && $\aud$           & \\
{$\min$}       & 1.164           &$\alr$& 2.411      && 2.001            &$\alr$ \\
{}             & $\aud$          &&                  && $\aud$           & \\
\bottomrule
\end{tabular}}
\end{table}

\subsection{\texorpdfstring{$M_3$}{M3}: Average \texorpdfstring{$\Delta D$}{Delta D} Length}
Even though the whegged robot with the min fitness function may yield evolutionary improvements in both environments after a single mutation, those increases in fitness may be very small and thus not contribute to the observed evolvability in that treatment. If so, one would expect $M_3$ to be very low for this treatment. However, as Table \ref{table:M3} reports, this is not the case: the legged robot with the min fitness function has the lowest $M_3$ value, and the value for the whegged robot with the min fitness function is significantly higher.

\begin{table}[h]
\centering
\caption{\label{table:M4} Mean of $M_4$ across treatments. 
Arrows indicate statistical significance between adjacent cells.
}
\resizebox{0.5\textwidth}{!}{\begin{tabular}{l c c c c c c}
\toprule
{\% in $I$} & \textbf{Legged} && \textbf{Wheeled} && \textbf{Whegged} & \\
\midrule
{$\sum$}    & 25.655          && 22.903           && 30.824           & \\
{}          &                 &&                  &&                  & \\
{$\prod$}   & 22.973          && 22.167           && 26.981           & \\
{}          & $\aud$          && $\aud$           && $\aud$           & \\
{$\min$}    & 49.519          && 48.674           &$\alr$& 54.332     &$\alr$ \\
{}          & $\aud$          && $\aud$           && $\aud$           &  \\
\bottomrule
\end{tabular}}
\end{table}

\subsection{\texorpdfstring{$M_4$}{M4}: Average Percentage of Points in Quadrant \texorpdfstring{$I$}{I}}
Using this metric we can conclude that the min function, regardless of morphology, seems to better force mutations to results in less catastrophic interference, at least according to this particular metric (Table \ref{table:M4}). We can conclude this because each $M_4$ value in the min row is significantly higher than the two $M_4$ values in the two entries above it.

\subsection{Performance in four environments.}
The same general pattern held when we scaled our approach from two to four environments. 
Namely, the pairwise comparisons that were significant (at the 0.05 level) in two environments remained so in four environments.
However, while the whegged robot with the product and min fitness functions similarly
outperformed the other treatments, they could not solve all four environments.

\begin{figure}[H]
\centering
\includegraphics[width=0.9\linewidth]{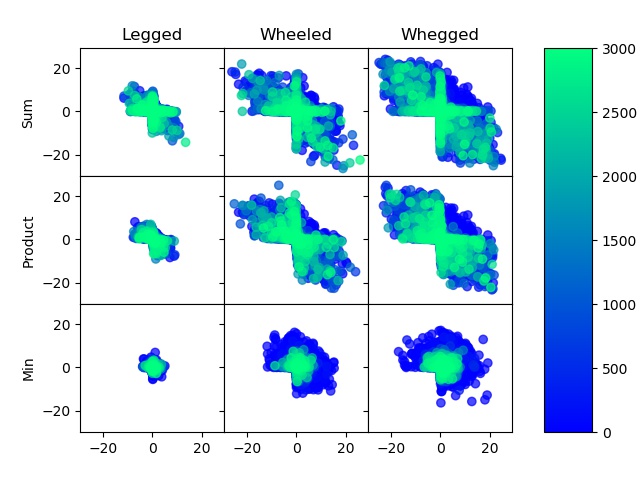}
\caption{\label{fig:raw}
Change in fitness ($\Delta D$, as defined in Eq.\ref{dist_record}), in two environments, for the three robots (Fig. \ref{fig:robots}) and three fitness metrics (Eq. \ref{fitnesses}), colored by the generation of the mutation. Dots in the upper-right quadrants of each robot-fitness cell represent beneficial changes in both environments; these mutations avoided catastrophic interference. Dots in the upper-left and lower-right quadrants of each cell are mutations that were beneficial in one environment but deleterious in the other; these changes caused catastrophic interference. We did not record mutations that were deleterious in both environments (lower-left quadrants).
}
\end{figure}
\begin{figure}[H]
\centering
\includegraphics[width=0.9\linewidth]{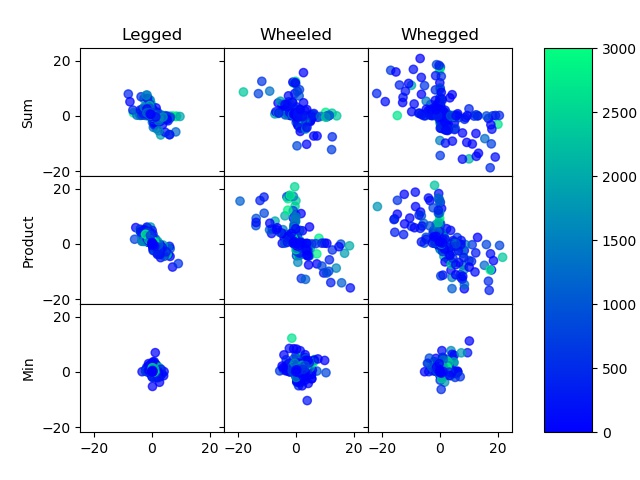}
\caption{\label{fig:best}
The same as Fig \ref{fig:raw} , but only for the run champs.
}
\end{figure}
\begin{figure}[H]
\centering
\includegraphics[width=0.9\linewidth]{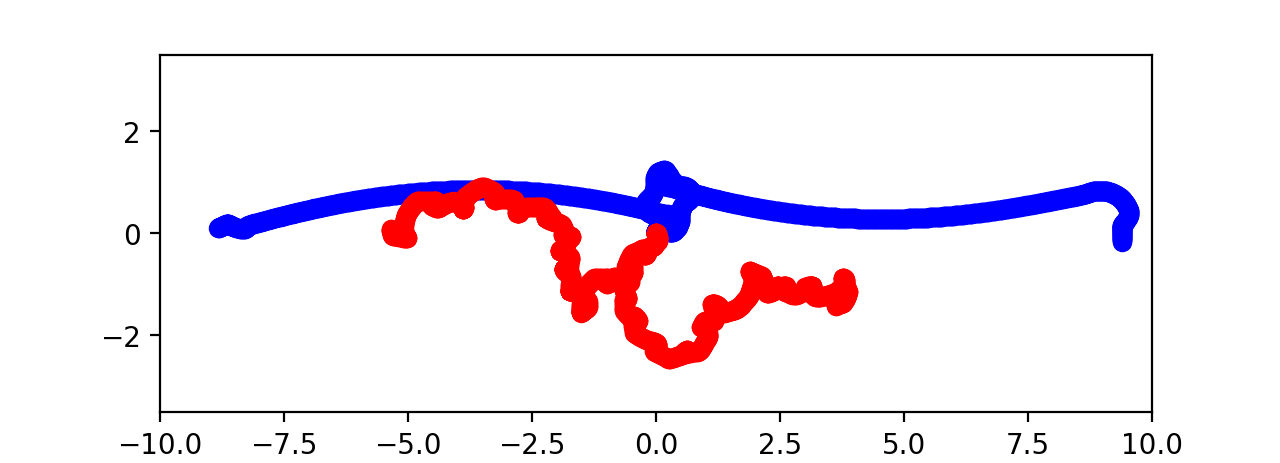}
\caption{\label{fig:trace}
A tracing of a typical whegged robot (blue) and legged robot (red) trained in two environments, under the min fitness function.
The light source is first placed at (9, 0), and then at (-9, 0). 
Video is available at \href{https://youtu.be/uWy33A5HZGM}{youtu.be/uWy33A5HZGM} .
}
\end{figure}

%%%%%%%%%%%%%%%%%%%%%%%%%%%%%%%%%%%%%%%%%%%%%%%%%%%%%%%%%%%%%%%%%%%%%%%%%%%%%

\section{Discussion}
As shown by Figures ~\ref{fig:best} and ~\ref{fig:raw} as we change the morphology from legged to whegged the robots demonstrate increased evolvability.
Thus the fitness landscape allows for larger jumps towards the optima.
This includes those jumps that avoid catastrophic interference altogether: mutations visualized by points in Figs. ~\ref{fig:best} and ~\ref{fig:raw} that lie in the upper right quadrant.

In conjunction, as we change the fitness function from sum to min, we see the spread of points in Figures ~\ref{fig:best} and ~\ref{fig:raw} condense toward the origin.
When combined with the whegged robot, we see a significant improvement in the metrics we used to measure catastrophic interference.
It appears that it is the combination of correct fitness function (min) with the correct morphology (whegged) that resists catastrophic interference: changes in morphology and fitness alone are not sufficient.
We hypothesize that this greater resistance to catastrophic interference is what enables the whegged robot, under the min fitness function, to achieve higher fitness within environments and consistent fitness across environments.

One objection to this hypothesis could attribute the performance of the whegged robot to the increased speed allowed for by wheels.
We do not feel that this is valid for two reasons: the wheeled robot also has wheels and does not achieve the same level of performance, and the evaluation time of a simulation was set such that all morphologies are able to reach the light source before the end of simulation.
Indeed we observed that all morphologies reached and waited at the light source when trained against a single environment.

In observing the behavior of the robots we noticed a pattern among whegged robots that could account for their resistance to interference.
Whegged robots move very rapidly in a circular pattern during the initial time steps of a simulation which may allow them to `sidestep' catastrophic interference by rapidly turning unfamiliar environments into familiar ones.
An example is shown in Fig. \ref{fig:trace}: the rotationally symmetric trajectories of the blue whegged robot indicates it has recognized two versions of the same environment. The red legged robot does not: its two trajectories are different, and take longer to diverge.
The wheeled and legged robot both seem to have much more difficulty in turning.

%%%%%%%%%%%%%%%%%%%%%%%%%%%%%%%%%%%%%%%%%%%%%%%%%%%%%%%%%%%%%%%%%%%%%%%%%%%%%%

\section{Conclusions}
This work suggests not only that the particular choice of morphology and fitness function for embodied agents can affect their resistance to catastrophic interference, but the very fact that the agent has a body can help.
In effect, an agent can use its body to move in such a way that a seemingly different training instance converges sensorially to a familiar instance.
The implication of this is that the very phenomenon of catastrophic interference itself may be to some degree a false problem arising from investigations using non-embodied systems: Since such systems do not have control over their input, they cannot align objects of interest in different training instances and thus reduce catastrophic interference. 

A simple example may suffice here: a human face that appears in two different locations in an image may be difficult for a non-embodied learner to recognize, unless there is a large amount of training data that contains diversity along that feature (face position).
In contrast, an embodied agent equipped with a camera that experiences the same two stimuli may learn to move such that the face is centered in its field of view. Furthermore, it may be that different types of embodied agents may more easily discover and perform this centering.
Finally, such appropriately embodied agents may thus be able to generalize about faces regardless of position using less training instances than the non-embodied agent because of this ability.
However, whether this latter system is indeed more scalable in this way compared to an equivalent non-embodied system remains as future work.

\chapter{Morphology dictates learnability in neural controllers}
\section{Abstract}
Catastrophic forgetting continues to severely restrict the learnability of controllers suitable for multiple task environments. Efforts to combat catastrophic forgetting reported in the literature to date have focused on how control systems can be updated more rapidly, hastening their adjustment from good initial settings to new environments, or more circumspectly, suppressing their ability to overfit to any one environment. When using robots, the environment includes the robot's own body, its shape and material properties, and how its actuators and sensors are distributed along its mechanical structure. Here we demonstrate for the first time how one such design decision (sensor placement) can alter the landscape of the loss function itself, either expanding or shrinking the weight manifolds containing suitable controllers for each individual task, thus increasing or decreasing their probability of overlap across tasks, and thus reducing or inducing the potential for catastrophic forgetting.

%%%%%%%%%%%%%%%%%%%%%%%%%%%%%%%%%%%%%%%%%%%%%%%%%%%%%%%%%%%%%%%%%%%%%%%%%%%%%%

\section{Introduction}
It has been shown in various single-task settings how an appropriate robot design can simplify the control problem \cite{lichtensteiger1999evolving,vaughan2004evolution,brown2010universal,bongard2011morphological,kriegman2019automated,PERVAN2019197}, but because these robots were restricted to a single training environment, they did not suffer catastrophic forgetting.

Catastrophic forgetting is a major and unsolved challenge in the machine learning literature \cite{french1999catastrophic,goodfellow2013empirical,kirkpatrick2017overcoming,masse2018alleviating}. Regardless of learning algorithm or task domain, a neural network trained to perform task A and then challenged with learning task B as well usually forgets A at the same rate as it learns B. Such interference can also occur when an agent attempts to learn tasks A and B simultaneously if gradients of improvement in A lead away from those of B.

In a multitask setting, \cite{powers2018effects} recently demonstrated that certain body plans suffer catastrophic forgetting, while others do not. It was hypothesized that a robot with the right morphology could in some cases alias separate tasks: certain designs are able to move in such a way that a seemingly different training instance converges sensorially to a familiar instance. However, this conjecture was not isolated and tested. Likewise, the relationship between the body and the loss landscape was not investigated.

In this paper, we provide a more thorough investigation on the role of embodiment in catastrophic forgetting based on the assumption that in order to avoid catastrophic forgetting, there must exist a set of control parameters that are adequately performant across multiple task environments simultaneously.
Since a robot's mechanical design can influence the set of controller parameters suitable for each individual task environment, we here test the hypothesis that a specific physical property of the robot's design|namely, the location of its sensors along its body|can help or hinder continual learning
by allowing for more or less overlap in suitable parameter settings across multiple task environments.

Using a simple yet embodied agent as our model, we analytically and empirically investigate how sensor location affects the weight manifolds of the neural controller over multiple tasks.
We show how morphological optimization often results in asymmetrical and unintuitive sensor arrangements with much more potential to allow learning algorithms to avoid catastrophic forgetting than more intuitive, symmetrical designs.
Thus, human designer bias, while often useful, can sometimes inadvertently increase the likelihood of catastrophic forgetting during learning.
This suggests that we should scrutinize our prior assumptions about the body plan of robots challenged with continual learning, and where possible replace them with end-to-end data-driven design automation.

%%%%%%%%%%%%%%%%%%%%%%%%%%%%%%%%%%%%%%%%%%%%%%%%%%%%%%%%%%%%%%%%%%%%%%%%%%%%%%

\section{Methods}

\begin{figure}[h]
    {\sf \textbf{A} \hspace{16em} \textbf{B} \hspace{18em} \textbf{C}} \\
    {\centering
    \includegraphics[height=1.3in]{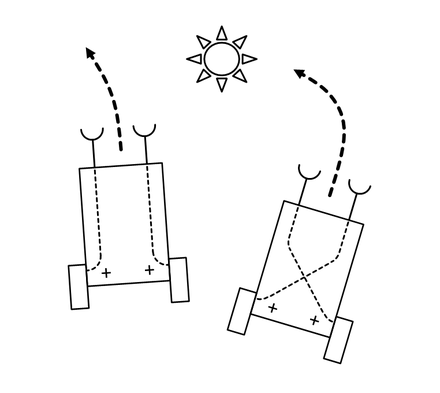}\hfill
    \includegraphics[height=1.3in]{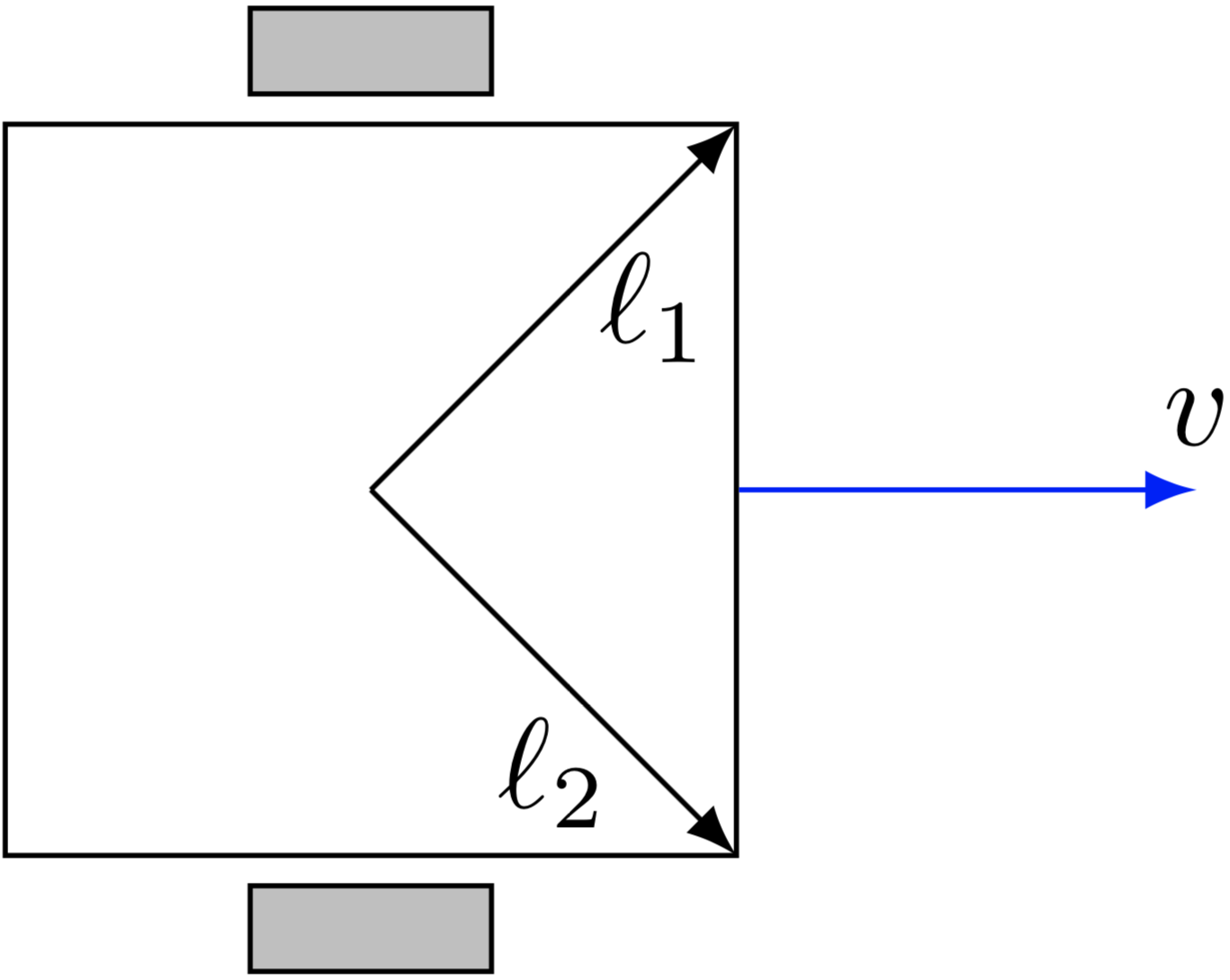}\hfill
    \includegraphics[height=1.3in]{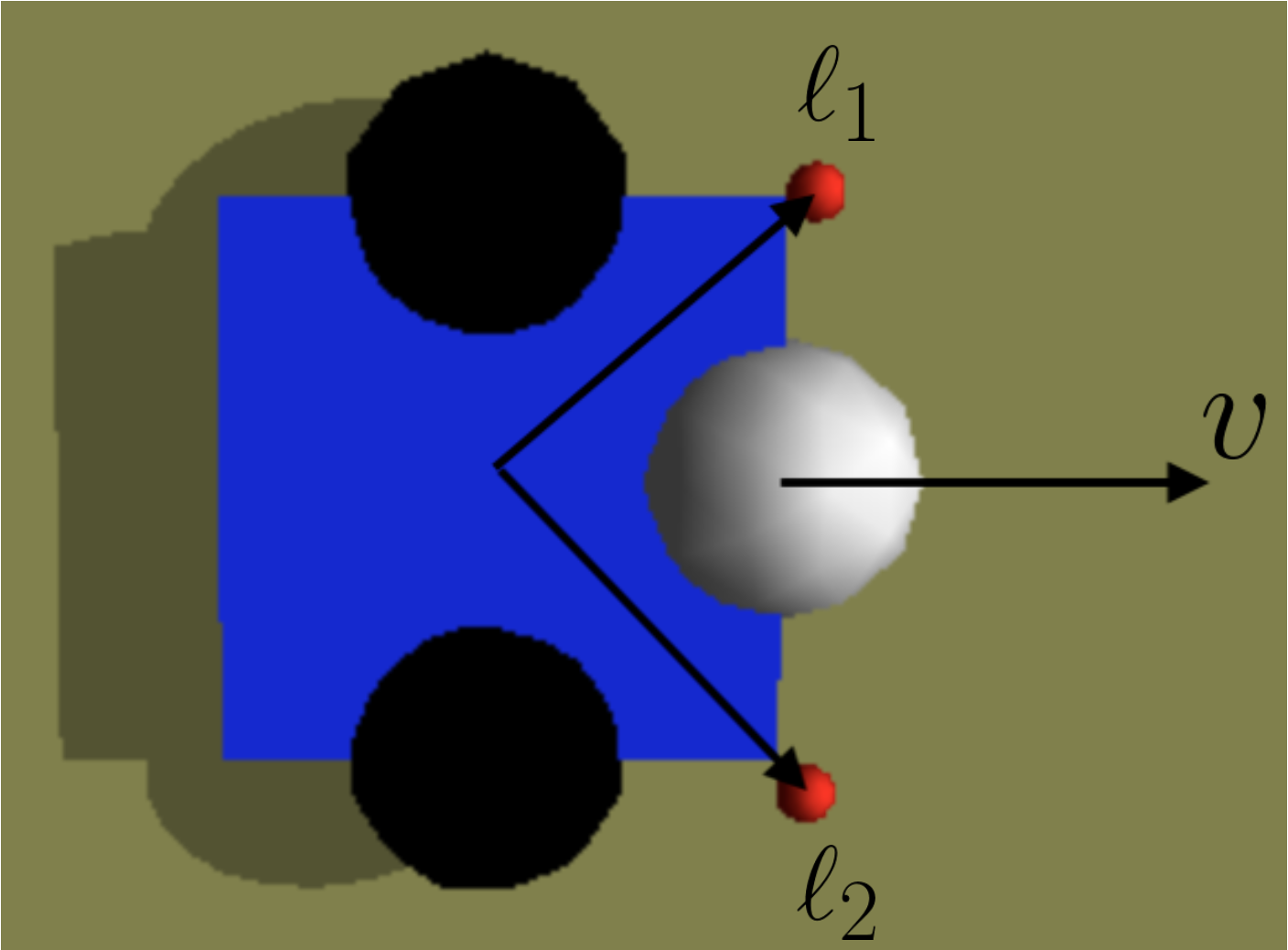}
    }
    \caption{\textbf{Modeling the robot.}
    \textbf{A:} The effect of lateral and contralateral synaptic connections (adopted from \cite{braitenberg1986vehicles}).
    \textbf{B:} The theoretical model with sensor positions determined by $\ell_1$ and $\ell_2$.
    \textbf{C:} The simulated robot with two light sensors (red), two motorized wheels (black), and a passive, anterior castor wheel for balance (gray).
    The robot is drawn (A-C) with symmetrical, anteriormost sensor placement, which we refer to in this paper as the ``canonical design''.
    }
    \label{fig:vehicles}
    
\end{figure}

\subsection{The robot}
\label{sec:vehicles}
The robot has a square frame, two separately-driven wheels, and two infrared sensors (Fig.~\ref{fig:vehicles}).
The sensors detect light according to the inverse square law, i.e., $1/d^2$, where $d$ is the distance from the light source;
occlusion was not modeled.
The motors driving the wheels are contralaterally connected to the sensors by weighted synapses yielding two trainable parameters $w_1, w_2 \in [-1.0, 1.0]$.

We here consider change to a single, isolated morphological attribute: the physical location of the two sensors, which can be placed anywhere on the dorsal surface of the robot's square body.
The location of the $i$-th sensor $\ell_i$ can be described by its Cartesian coordinates $\ell_i = (x, y)$, where $x, y \in [-0.5, 0.5]$, and (0, 0) denotes the center of the body (Fig.~\ref{fig:vehicles}B).

The effect of sensor location $\ell_i$ can be measured with respect to the space, denoted $\theta$, of possible synapse weight pairs ($w_1$, $w_2$).
Since we cannot perform an exhaustive sweep over the infinitude of possible sensor positions,
we discretized each dimension of $\ell_i$ into nine uniformly-spaced bins.
Because sensors are varied in two dimensions ($x$ and $y$) there are $9^2=81$ possible locations for each sensor;
and because there are two such sensors, the space $\theta$ is discretized into a 81-by-81 uniformly-spaced grid, thus yielding a searchable space of 6561 possible robot designs.

For each of the 6561 designs, we conducted another sweep over the synapse weights $(w_1, w_2)$, likewise discretizing each weight into 121 evenly-space values, yielding $121^2 = 14641$ different weight configurations.
Finally, for each of the $6561\times14641=96059601$ evaluated combinations of sensor locations and weight values, we analyzed the robot analytically using differential equations and empirically using a physics engine. These discretizations were chosen to be as small as possible within the limit of our computational resources and time.

%%%%%%%%%%%%%%%%%%%%%%%%%%%%%%%%%%%%%%%%%%%%%%%%%%%%%%%%%%%%%%%%%%%%%%%%%%%%%%%

\subsection{The task environments}
\label{sec:environments}
The task is phototaxis in four environments, which differ in their position of the light source in relation to the robot. 
The light source is placed at polar coordinates $(r, \varphi)$ where $\varphi \in \{45^{\circ}, 135^{\circ}, 225^{\circ}, 315^{\circ}\}$ and $r$ is a fixed distance. 
A controller was considered successful for a given environment if the robot comes within 0.2~cm of the light source at any time during its evaluation period.

While there is of course a general strategy that solves the task for all environments (follow the light), the easiest gradients to follow in the loss landscape are initially those which produce forward locomotion in a single direction and cause the robot to ignore the light. 
This is because, from the robot's perspective, due to the inverse square law of light decay, improving its ability to move in the one environment with least loss earns quadratically
more reward than improvements to locomotion in any of the other three environments in which the robot is less proficient. This causes the catastrophic forgetting experienced by neural learning algorithms.

%%%%%%%%%%%%%%%%%%%%%%%%%%%%%%%%%%%%%%%%%%%%%%%%%%%%%%%%%%%%%%%%%%%%%%%%%%%%%%%

\subsection{The metrics}
\label{sec:metrics}
We here define two metrics: $M_L$ and $M_{CF}$, that are measured over $k=4$ environments. These metrics measure how a robot design impacts the weight space of the controller and consequently measure how amenable to learning a robot would have been if the controller were to be learned with a standard learning algorithm rather than found by grid search.
$M_L$ measures controller learnability: how easy it would be to learn a generalist controller. 
$M_{CF}$ measures resistance to catastrophic forgetting: the probability that a environment-specific controller will generalize to other environment.

\begin{floatingfigure}[lrp]{0.5\textwidth}
    \centering
    \includegraphics[width=0.4\textwidth]{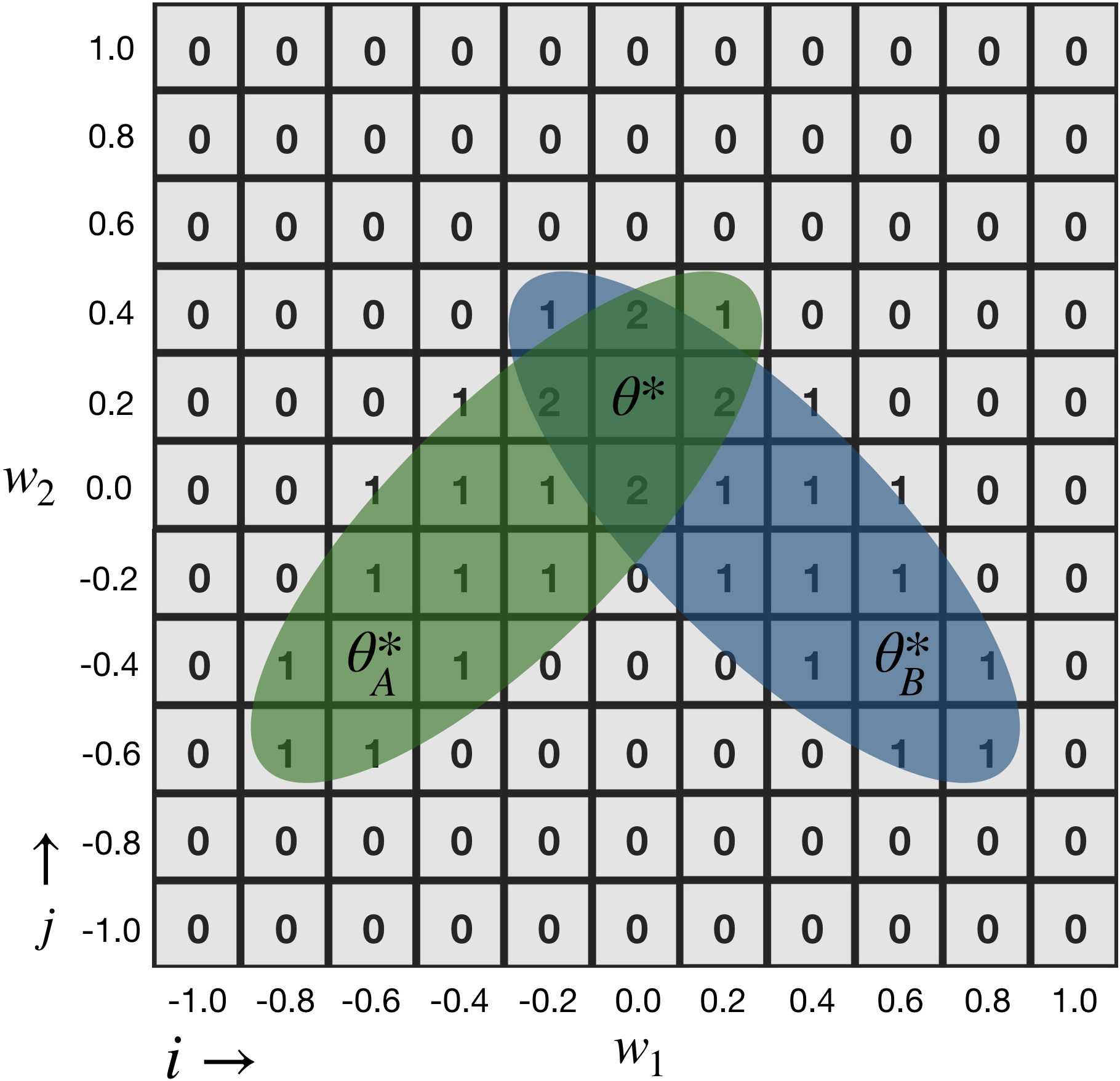}
    \caption{A general example of overlapped binary success matrices for some tasks A and B.
    Each element represents a different controller. Generalist controllers sit inside the intersection $\theta^*$ of successful environment-specific controllers $\theta_k^*$.}
    \label{fig:metrics2}
\end{floatingfigure}

For each mechanical design $(\ell_1, \ell_2)$, we expect some optimal manifolds $\theta_k^*$ in the space of control parameters $(w_1, w_2)$ to succeed for a specific environment $k$. For a controller to be successful in multiple environments, it must reside within the intersection of environment-specific manifolds, $theta^*$, on the loss surface. Thus, the likelihood of finding a generalist controller|its learnability|will be proportional to the size of the intersection ($M_L$). Likewise, a controller's potential to resist catastrophic forgetting ($M_{CF}$) will be proportional to the ratio of generalist controllers (those successful in all four environments) to specialists (those successful in at least one environment).

Given a design $(\ell_1, \ell_2)$ and environment $k$, a binary success matrix $S^{k}(\ell_1, \ell_2)$ is constructed such that each element $S^{k}_{i, j}(\ell_1, \ell_2)$ is either 1 (success) or 0 (failure). By overlapping the success matrices for a fixed design across the four environments, we can visualize the manifolds $\theta_k^*$ where $k\in\{1, 2, 3, 4\}$ for the robot (Fig.~\ref{fig:metrics2}).

We define the overlap $\mathcal{O}$ as a element-wise sum of the success matrices over each environment $k$:
\begin{equation}
    \label{eq:overlap}
    \mathcal{O} = \sum_{k=1}^4 S^k(\ell_1, \ell_2) .
\end{equation}

The learnability metric is simply the proportion of 4s (where a 4 represents success in all 4 environments) in the overlapped success matrices to the entire matrix space:
\begin{equation}
    \label{eq:metric1}
    M_L = \frac{g_4(\mathcal{O})}{n^2},
\end{equation}
where $g_k$ is a function that counts the total elements of a matrix with value equal to $k$ and $n$ is the square dimension of the matrix defined by the discrete parameter sweep.

Resistance to catastrophic forgetting is measured by:
\begin{equation}
    \label{eq:metric2}
    M_{CF} = \begin{cases}
    0 \; \text{ if } \mathcal{O} \text{ is a null matrix}, \\
    g_4(\mathcal{O}) \left[\sum_{k=1}^4 g_k(\mathcal{O})\right]^{-1}
    \; \text{otherwise}.
    \end{cases}
\end{equation}
which is the number of control parameters that solved all four environments divided by the number of control parameters that solved at least one.

%%%%%%%%%%%%%%%%%%%%%%%%%%%%%%%%%%%%%%%%%%%%%%%%%%%%%%%%%%%%%%%%%%%%%%%%%

\subsection{The theoretical model.}
\label{sec:theoretical}
The location and orientation of the robot can be defined by a system of differential equations, where the change in position and orientation is determined by the change in light captured by two sensors. 
Ignoring deviations from the idealized environment, such as sensor noise and friction,
the rate of angular and linear velocities will be proportional to a linear combination of the sensor values.

Let $\alpha(t)$ be the angle of the robot at time $t$, where $\alpha=0$ denotes the positive $x$ direction, and $\phi(t) = (x(t), y(t))$ be the position of the robot in the world, then if the robot is located at the origin and facing east ($\alpha = 0$), its two light sensors are located exactly at $\ell_1$ and $\ell_2$, and they each capture a some amount of light $s_1(t)$ and $s_2(t)$, respectively. 

Hence the absolute position of the $i$-th sensor is $\phi(t) + R_\alpha \ell_i^{\hspace{3pt}T}$, where
	\begin{equation}
		R_\alpha = \begin{bmatrix}
			\cos \alpha & -\sin \alpha \\
			\sin \alpha & \cos \alpha \\
		\end{bmatrix}
	\end{equation}
is the two-dimensional counterclockwise rotation matrix (in the amount $\alpha$). 

If we formulate the problem such that it is the robot's initial position and heading that is adjusted in each environment, instead of the position of the light source, we can assume that the source is always at the origin. Then, the distance of $\ell_i$ from the light source is given by:  \mbox{$\|\phi(t) + R_\alpha \ell_i^{\hspace{3pt} T}\|$.}
And since the intensity of light is inversely proportional to the square of the distance, the sensor values are given by: 
\begin{equation}
\label{eq:sensor_value}
s_i(t) = c \cdot \|\phi(t)^T + R_\alpha {\ell_i}^T\|^{-2},
\end{equation}
where $c$ is a constant that we set equal to one. 

Assuming the robot turns based on the difference between the sensor values (with weights applied), the velocity of the robot is the average of the two sensor values. 
Thus, the following system of equations determines the location and orientation of the robot:
	\begin{equation}
	    \label{eq:ode}
		\begin{cases}
			\dot{x} = v(t) \cos \alpha	\\
			\dot{y} = v(t) \sin \alpha \\
			\dot{\alpha} = w_1s_1(t) - w_2s_2(t), \\
		\end{cases}
	\end{equation}
where $v$ is the velocity of the robot given by $2v(t) = w_1s_1(t)+w_2s_2(t).$

%%%%%%%%%%%%%%%%%%%%%%%%%%%%%%%%%%%%%%%%%%%%%%%%%%%%%%%%%%%%%%%%%%%%%%%%%

\subsection{The empirical model.}
\label{sec:empirical}
Because our theoretical model is highly abstracted from the real world and built on a number of assumptions (no friction, motor limits, collisions, etc.) which may potentially affect the robot's behavior, we also empirically test our claims by simulating the robots inside a physics engine.

The robot is simulated using Open Dynamics Engine (Fig.~\ref{fig:vehicles}C).
Just like the theoretical model, the simulated robot contains two light sensors, which innervate two motorized, spherical wheels (each with a single axis of rotation), which are attached midway along the sides of a $1 \times 1 \times 0.13$~cm box.
Additionally, an anterior passive castor wheel was added for balance. Finally, a light source is simulated on the floor of the environment at polar coordinates $(r, \alpha)$ as a fixed sphere with radius 0.2 cm. In simulation, the behavior of a robot in a given environment is taken to be successful if it collides with the light source at anytime during an evaluation period of 2500 time steps ($dt=0.05$) or 125 seconds.

In order to replicate the baseline behavior of the canonical robot design it was necessary to pre-optimize various physical attributes of the robot's body, including the mass of each component, the radii of the wheels, and the maximum torque, speed, and target actuation rate. A multiobjective optimization algorithm \cite{hadka12b} was used to find a base morphology, with the sensors fixed in the canonical position, that was both performant and stable. The first objective was to maximize the performance of the robot (distance from the light source), summed across all the four environments. The second objective was to minimize the sum of the maximum torque, speed and target actuation rate. This second objective is used to avoid both simulator instability and behavior that is unlikely to transfer to reality.

After discovering a good base morphology, we performed the nested grid search described in \S\ref{sec:vehicles}, for sensor locations ($\ell_1$, $\ell_2$) and weights ($w_1, w_2$).

%%%%%%%%%%%%%%%%%%%%%%%%%%%%%%%%%%%%%%%%%%%%%%%%%%%%%%%%%%%%%%%%%%%%%%%%%%%%%%%

\begin{figure}[htbp]
    {\sf \textbf{A} \hspace{17em} \textbf{B} \hspace{19em} \textbf{C}}\\
    {\centering
    \includegraphics[height=1.3in]{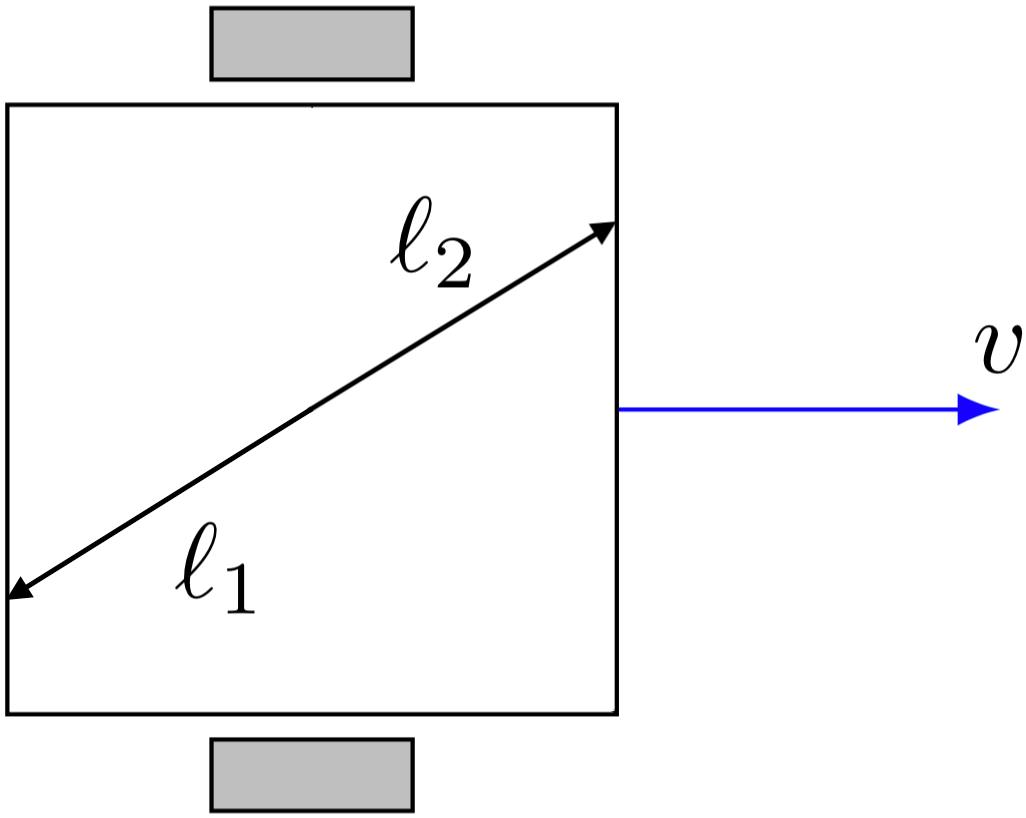}\hfill
    \includegraphics[height=1.3in]{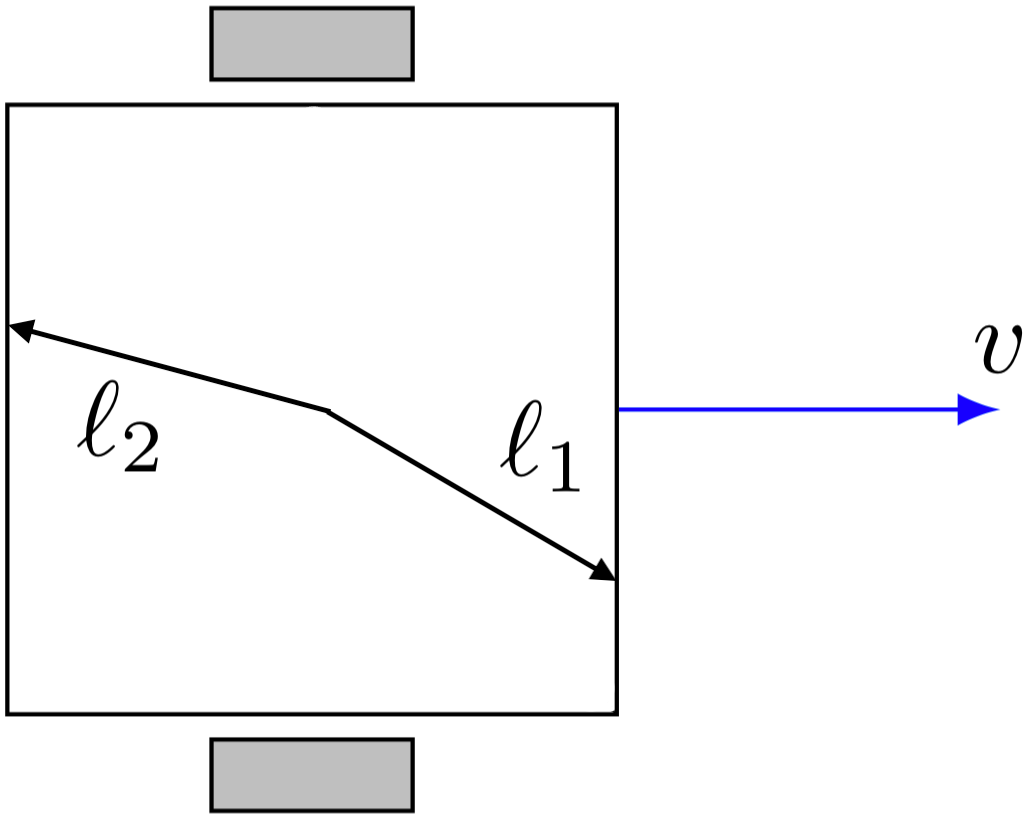}\hfill
    \includegraphics[height=1.3in]{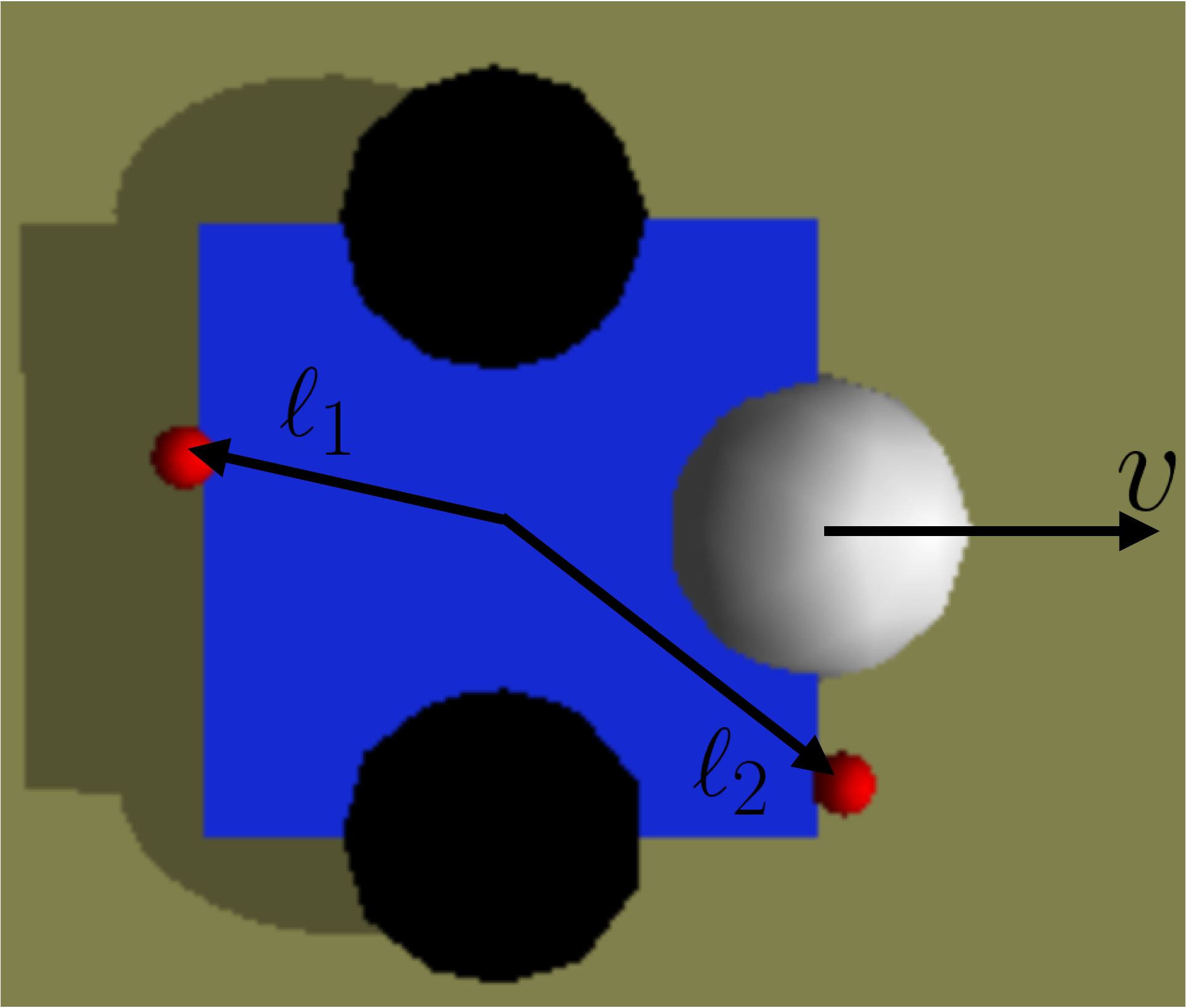}
    }
    \caption{\textbf{The best designs} under the 
    theoretical model according to controller learnability (\textbf{A}; Metric $M_L$) and resistance to catastrophic forgetting (\textbf{B}; Metric $M_{CF}$).
    Under the empirical model, the design with the highest controller learnability was also the most resistant to catastrophic forgetting (\textbf{C}).
    Although the design space we swept over contains many symmetrical sensor arrangements, and most real robots utilize symmetrical sensor distributions,
    the best designs are notably asymmetrical.
    }
    \label{fig:best_morphology}
\end{figure}

\begin{figure}[t]
    \centering
    \includegraphics[width=\textwidth]{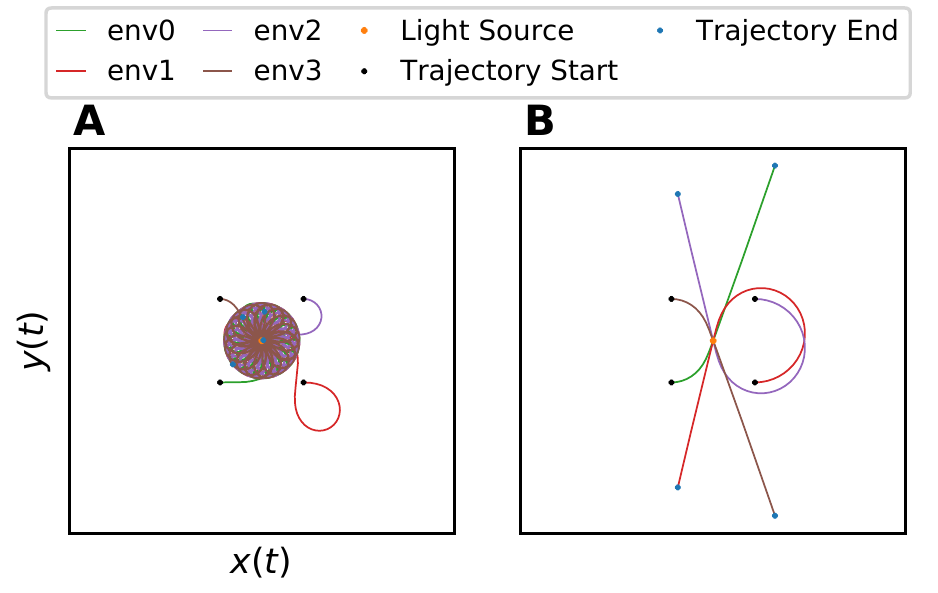}
    \caption{\textbf{Successful trajectories with canonical (symmetrical) sensor location under the theoretical model.} 
    With canonical sensor placement $ \{ \ell_1 = (0.5, 0.5)$, $\ell_2 = (0.5, -0.5) \} $ (Fig.~\ref{fig:vehicles}B),
    only 57 of the $121^2$ evaluated controllers (0.4\%) were successful all four environments.
    \textbf{A:} The trajectories generated by one of the successful controllers $(w_1, w_2) = (0.6, 0.98)$.
    This controller initially generated phototaxis, but passed through the light source and continued to move away from it.
    \textbf{B:} The trajectories generated by another successful controller $(w_1, w_2) = (0.77, 0.77)$. 
    This controller continuously spirals about the light source.
    The light source is drawn once at the origin, and the initial positions/orientations of the robot relative to the it are superimposed for the four environments.
    }
    \label{fig:canonical_trajectories}
\end{figure}

\begin{figure}[t]
    \centering
    \includegraphics[width=\textwidth]{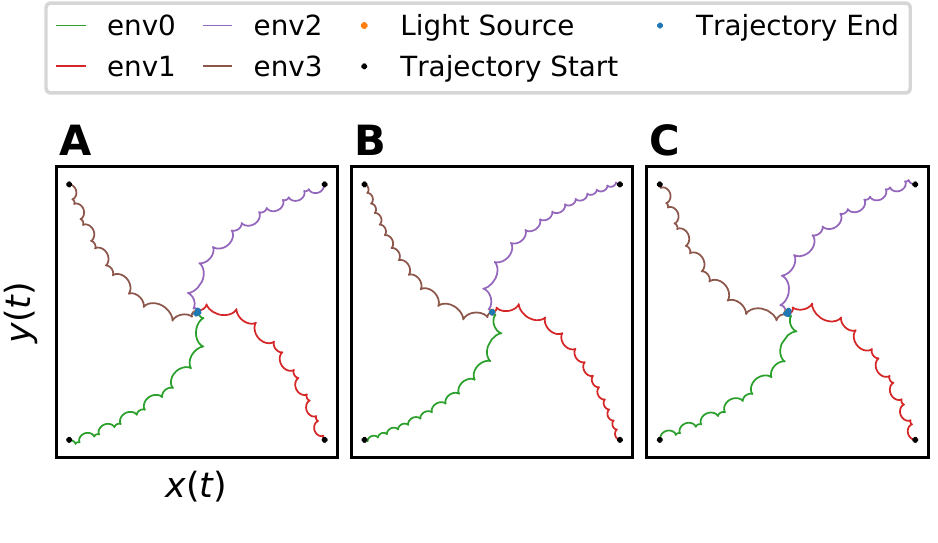}
    \caption{\textbf{Trajectories of the design with maximal controller learnability, as measured by metric $M_1$ (Eq.~\ref{eq:metric1}) under the theoretical model.} 
    With sensor locations $\ell_1 = (-0.5, -0.25)$ and $\ell_2 = (0.5, 0.25)$ (Fig.~\ref{fig:best_morphology}A), 2255 of the $121^2$ evaluated controllers (15.4\%) were successful all four environments. 
    \textbf{A:} The trajectories generated by one of the successful controllers, parameterized by weights $(w_1, w_2) = (-0.85, 0.82)$.
    \textbf{B:} The trajectories with weights $(-0.8, 0.6)$.
    \textbf{C:} The trajectories with weights $(-0.28, 0.37)$. 
    The axes are equivalent to those in Fig.~\ref{fig:canonical_trajectories}.
    }
    \label{fig:best_trajectories}
\end{figure}
\vspace{-1.0em}

\section{Results}

\subsection{Theoretical results.}
We employed SciPy (\textit{scipy.integrate.odeint}) for numerical integration of the robot's location and orientation (Eq.~\ref{eq:ode}), for $10^5$ timesteps.

For each evaluated mechanical design and controller (sensor locations and synapse weights), the robot's trajectory is computed in each of the four environments defined in \S\ref{sec:environments}. 
As in the empirical model, if robot's trajectory comes within 0.2 units of the light source, the robot is determined to have succeeded in that environment. 
Otherwise, it is determined to have failed.

The mechanical design sketched in Fig.~\ref{fig:best_morphology}A (and its mirror image when reflected about the sagittal plane) had the highest controller learnability score, with $M_L = 0.286$. However it did score the best in resistant to catastrophic forgetting: the proportion of resistant to nonresistant controllers for that design was $M_{CF}=0.636$, whereas several other found designs had full resistance $M_{CF}=1$. But those with a perfect ratio $M_{CF}=1$ had much smaller optimal weight manifold: the highest learnability score achieved by this group was $M_L=0.206$. 
In other words, while all the successful environment-specific controllers for these designs generalize across all four environments, the manifold containing them is much smaller and thus would be more difficult to find if controller parameters were to be optimized by learning.

The canonical design had a much lower controller learnability ($M_L=0.049$) and resistance to catastrophic forgetting ($M_{CF}=0.24$), than many found asymmetrical designs.

For both the canonical, symmetrical design (Fig.~\ref{fig:canonical_trajectories}) and the design with the highest controller learnability score (Fig.~\ref{fig:best_trajectories}) there are initial conditions that generate persistent phototaxis: the robot moves toward the light source and remains near it.
However, whereas 35 of the 35 found phototaxing controllers for the found design remain in the neighborhood of the source, only 2 of the 6 found controllers for the canonical design do so.
Some initial conditions of the canonical design initially produce phototaxis, but the design passes through the source and then continues to move away from it (Fig.~\ref{fig:canonical_trajectories}A). This was not observed to occur with the ``optimized'' designs.

\begin{figure}[htbp]
    \centering
    \includegraphics[width=\textwidth]{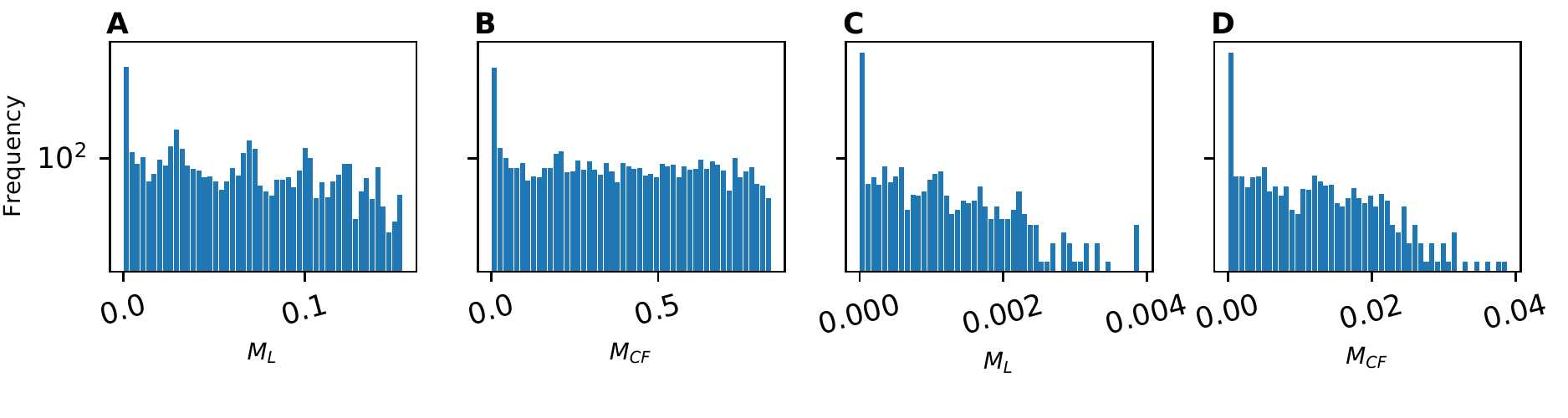}
    \vspace{-2.5em}
    \caption{\textbf{Measuring successful multitask learning.}
    The distribution of metrics $M_L$ and $M_{CF}$, for all evaluated designs, in the theoretical (A, B), empirical (C, D).
    Metric $M_L$ (Eq.~\ref{eq:metric1}) indicates controller learnability: the proportion of controllers deemed successful in all four environments, for a given design.
    Metric $M_{CF}$ (Eq.~\ref{eq:metric2}) indicates resistance to catastrophic forgetting: the ratio of the number of controllers successful in all environments, over the number successful in at least one. If no controllers are successful for a given design, $M_{CF}=0$.}
    \label{fig:histogram}
\end{figure}

\subsection{Empirical results.}
As with the theoretical model the empirical model showed that non-intuitive asymmetrical designs scored higher in learnability and in resistance to catastrophic forgetting. However unlike the theoretical model one design performed the best on both metrics.

The found asymmetrical design shown in Fig.~\ref{fig:best_morphology}C  
had both the highest generalist controller learnability ($M_L = 0.0039$) and resistance to catastrophic forgetting ($M_{CF} = 0.038$). 
Overall, there were 57 generalist phototaxing controllers found (out of 14641 evaluated; 0.389\%) for this design,
compared to only one generalist phototaxing controller found (0.0068\%) for the canonical, symmetrical design.
The controller learnability of the canonical design was thus $M_L=0.000068$; and its resistance to catastrophic forgetting was $M_{CF}=0.00052$.
Thus, the found asymmetrical design has both higher controller learnability and resistance to catastrophic forgetting.

\begin{figure}[htbp]
    \centering
    \includegraphics[width=\textwidth]{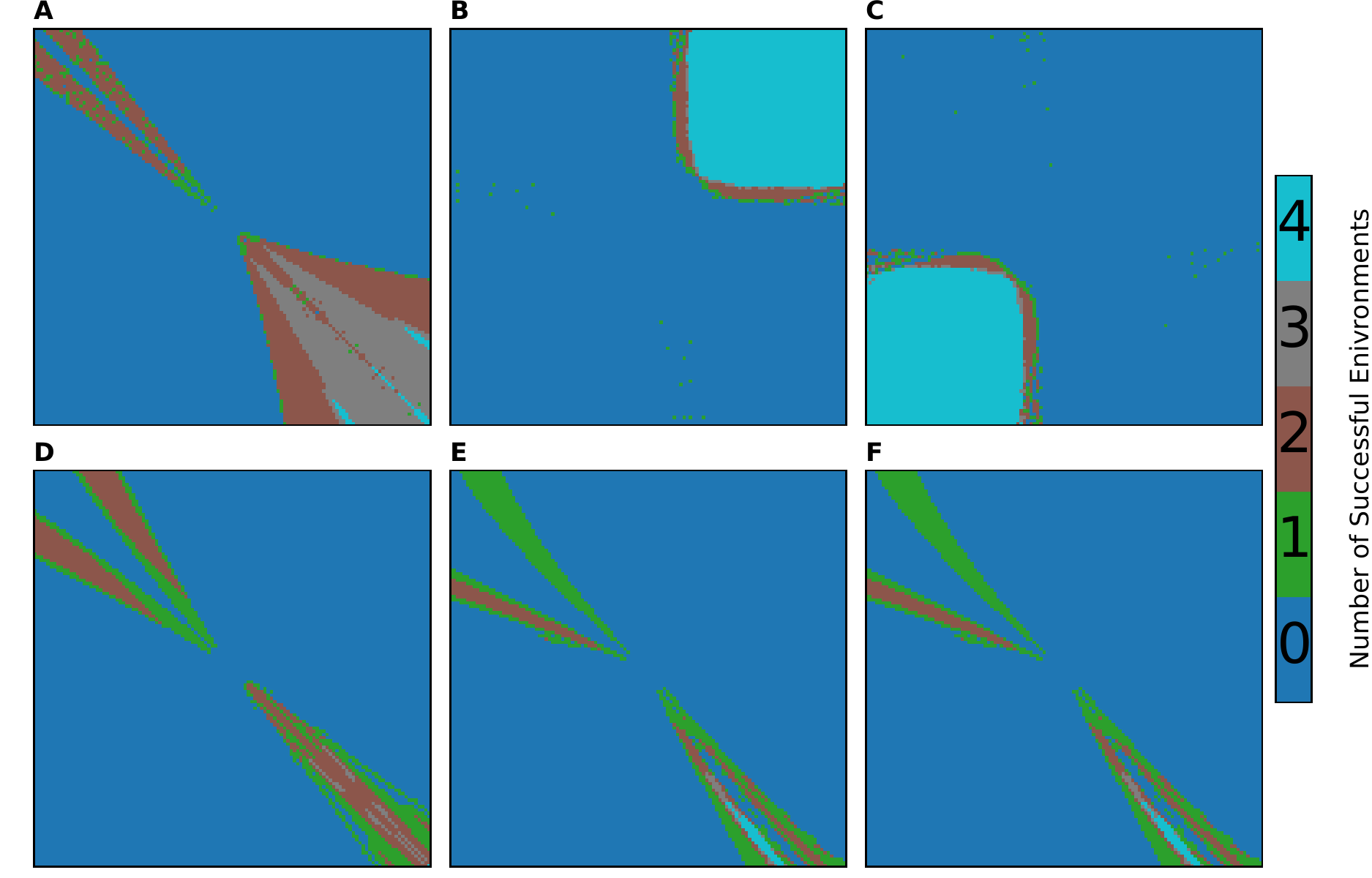}
    \caption{\textbf{Measuring learnability and forgetting.}
    For both the theoretical (\textbf{A-C}) and empirical models (\textbf{D-F}), 
    we performed a 121-by-121 grid search of controller weights (14641 unique controllers) nested within a 81-by-81 grid search for sensor locations (6561 unique designs).
    The controller space its mapped for the canonical, symmetrical design (A, D), the design with highest controller learnability ($M_L$; Eq.~\ref{eq:metric1}) (B, E), and the design most resistant to catastrophic forgetting ($M_{CF}$; Eq.~\ref{eq:metric2}) (C, F).
    Under the controller sweep on (D-F), the design with the highest controller learnability also had the greatest resistance to forgetting, so E and  are identical.
    Each pixel represents a different controller $(w_1, w_2)$ for the given design, and is colored by the number of environments that the combination successfully exhibited phototaxis 
    (i.e., the overlapped binary success matrices, defined by Eq.~\ref{eq:overlap}).
    Under both the theoretical and empirical models,
    the unintuitive asymmetrical designs (B, E) were found to have higher controller learnability and greater resistance to forgetting in their landscape than their respective canonical design (A, D) as measured by the number pixels in the heatmap that are successful in all four environments (cyan).
    Likewise, the asymmetrical designs (C, F) had higher resistance to catastrophic forgetting as measured by the number of cyan pixels to non-blue pixels.
    }
    \label{fig:landscape}
\end{figure}

\subsection{Overview.}
In Fig.~\ref{fig:landscape} the successes of weight manifolds for all of these design in both the theoretical and empirical model can be seen in detail, where cyan represents weight assignments that succeed in all for environments for a given design. These weight manifolds show clearly that in this case the weight assignments for the asymmetrical would be much easier to find by a learning algorithm while the canonical design is akin to looking for a needle in a haystack.

Fig.~\ref{fig:histogram} plots the frequency of metrics $M_L$ and $M_{CF}$ (Eqs.~\ref{eq:metric1} and \ref{eq:metric2}, respectively) within each bin of the grid search. This again shows how there are many designs (including intuitive symmetric ones) that score poorly on $M_L$ and $M_{CF}$ while there are relatively few designs that perform well. Thus a given design has a drastic effect on the theoretical learnability of a robots controller parameters.

%%%%%%%%%%%%%%%%%%%%%%%%%%%%%%%%%%%%%%%%%%%%%%%%%%%%%%%%%%%%%%%%%%%%%%%%%%%%%

\section{Discussion}
In this paper, we considered a simple robot and task in order to sample the entire loss landscape of the weight manifold at a relatively high resolution. 
While we haven't tested these robot with any specific learning algorithm, our results suggest that changes in one element of a robot's design (sensor location) can fundamentally alter the loss surface, thus influencing the controller's learnability, and resistance to catastrophic forgetting. 
More specifically, by changing sensor location, we observed changes in the number and placement along the loss surface of control parameters suitable for individual environments, as well in how these optimal yet environment-specific parameters overlapped across different environments to produce generalist controllers which resist catastrophic forgetting. However, we acknowledge that this work mainly builds a theoretical foundation and that our metrics need to be tested against existing methods for learning.

Previous efforts to avoid catastrophic forgetting have relied almost exclusively on increased control complexity. 
Most were focused on making changes to small subsets of neural network weights \cite{kirkpatrick2017overcoming,masse2018alleviating,french1991using,robins1995catastrophic,he2018overcoming,beaulieu2018combating,schwarz2018progress,titsias2019functional}. 
Others have attempted to sidestep the problem by learning good initial weights such that they can be quickly updated when switching between tasks \cite{finn2017model,gidaris2018dynamic}. 
We have shown here that, in theory, regardless of the algorithm used it is also possible to alleviate catastrophic forgetting by changing aspects of the robot's design, without increasing control complexity, but doing so can be non-intuitive.

We found that even the seemingly trivial case of phototaxis with contralateral connections described by \cite{braitenberg1986vehicles} can require morphological tuning to work as expected in a single simulated environment, and that, when challenged to perform in additional environments, other adjustments in morphology, specifically to sensor location, could either suppress or multiply the potential for catastrophic forgetting by expanding or shrinking the overlap of performant controller settings for that body plan across different task environments.

The physical location of sensors is thus a relevant property of robots that is nevertheless abstracted away in the (mostly disembodied) systems that address catastrophic forgetting reported in the literature to date.
While sensor location could in principle be dynamically controlled via a lattice of sensors \cite{kramer2011wearable} or adjustable antenna \cite{fend2003active}, 
change in (and rational control over) other morphological attributes|such as geometry \cite{kriegman2019automated}, material properties \cite{narang2018transforming}, or the number and placement of actuators \cite{lipson2000automatic}|is much more difficult in practice, and such design elements are almost always presupposed and fixed prior to training \cite{cheney2018scalable}.

However, unless experimental proof is obtained in the real world, this theory will remain speculation.
In fact it is possible that the proposed empirical model using rigid body physics was more disconnected from reality than our theoretical model.
The simulated wheels, for instance, have just a single point of contact with the ground.
A more realistic surface contact geometry might completely change the optimal sensor locations, 
but there's also reason to believe that the loss surface manifolds containing adequate controllers for a compliant body could be larger than those of a rigid body \cite{kriegman2019automated,hauser2011towards}, further increasing the probability of overlap across tasks.

In the limit, machines with the right morphology may use a single controller to accomplish a set of tasks that appear 
disparate to a robot with a different body plan.
For example, a granular jamming gripper \cite{brown2010universal} need not precisely control the placement of each joint around differently shaped objects: a single policy (vacuum air, hold, relax) works regardless of object shape.
However, this control policy is exceedingly simple.
The degree to which morphology influences learnability in more complex robots,
task environments and behaviors has yet to be investigated, but will be the focus
of future work.

In this work, two control- and two morphology parameters 
were optimized. In future work we will investigate whether 
co-optimizing the morphology and control parameters
confers greater overall learnability on the robot compared to a robot with a fixed
morphology and four control parameters. This will help determine whether a poorly
chosen mechanical design can be compensated for by increased control complexity.

\chapter{A good body is all you need: avoiding catastrophic interference via agent architecture search}
\section{Abstract}
In robotics, catastrophic interference continues to restrain policy training across environments. Efforts to combat catastrophic interference to date focus on novel neural architectures or training methods, with a recent emphasis on policies with good initial settings that facilitate training in new environments. However, none of these methods to date have taken into account how the physical architecture of the robot can obstruct or facilitate catastrophic interference, just as the choice of neural architecture can. In previous work we have shown how aspects of a robot's physical structure  (specifically, sensor placement) can facilitate policy learning by increasing the fraction of optimal policies for a given physical structure. Here we show for the first time that this proxy measure of catastrophic interference correlates with sample efficiency across several search methods, proving that favorable loss landscapes can be induced by the correct choice of physical structure. We show that such structures can be found via co-optimization---optimization of a robot's structure and control policy simultaneously---yielding catastrophic interference resistant robot structures and policies, and that this is more efficient than control policy optimization alone. Finally, we show that such structures exhibit sensor homeostasis across environments and introduce this as the mechanism by which certain robots overcome catastrophic interference.

%%%%%%%%%%%%%%%%%%%%%%%%%%%%%%%%%%%%%%%%%%%%%%%%%%%%%%%%%%%%%%%%%%%%%%%%%%%%%%%%

\section{Introduction}
Catastrophic interference is a phenomenon that occurs when training policies on multiple environments simultaneously: a training step that results in improvement in one environment causes a greater reduction in performance in other environments. This comes with various consequences, one of which are policies that specialize to only one of the environments. This problem is often related to the problem of catastrophic forgetting \cite{french1999catastrophic,goodfellow2013empirical,kirkpatrick2017overcoming,masse2018alleviating}, however in catastrophic forgetting environments are trained sequentially, where in catastrophic interference they are trained simultaneously as stated.

Many of the methods used to combat catastrophic intereference rely solely on advances to neural architecture or training methods \cite{zhang2020differentiable}. Most are focused on making changes to small subsets of neural network weights \cite{kirkpatrick2017overcoming,masse2018alleviating,french1991using,robins1995catastrophic,he2018overcoming,beaulieu2018combating,schwarz2018progress,titsias2019functional}. Others have attempted to sidestep the problem by learning good initial weights such that they can be quickly updated when switching between tasks \cite{finn2017model,gidaris2018dynamic}.

None of these methods however consider the effects that the architecture of the robot housing the policy has on this phenomenon. The robot's body is often considered a part of the environment \cite{ha2019reinforcement}. From this perspective, it is the only part of the task environment that is changeable prior to or during training. Consequently, it has been shown in various task settings that an appropriate robot design can simplify the problem of learning a sufficient control policy \cite{lichtensteiger1999evolving,vaughan2004evolution,brown2010universal,bongard2011morphological,kriegman2019automated,PERVAN2019197}. However, so far these findings have been restricted to examples with a single training environment; they did not need to overcome multi-domain problems such as catastrophic interference. In a multitask setting, \cite{powers2018effects} recently demonstrated that certain body plans suffer catastrophic interference, while others do not. It was hypothesized there that a robot with an appropriate design could in some cases alias separate tasks: certain designs are able to move so that seemingly different environments converge sensorially to a common and familiar observation. This is not unlike tilting one's head to recognize a familiar face in a rotated image. However, this conjecture was not isolated and tested. Likewise, the relationship between the body and the loss landscape was not investigated.

Here, we provide a more thorough investigation of the role of embodiment in catastrophic interference, based on the assumption that in order to avoid catastrophic interference, there must exist a set of policy parameters that yield adequate performance across several task environments. Since a robot's mechanical design can change which sets of policy parameters are appropriate for each individual task environment, we here test the hypothesis that a specific physical property of the robot's design---sensor distribution---can help or hinder learning by inducing more or less overlap in specialized policy parameter sets across multiple task environments. These regions of overlap thus correspond to sets of general purpose policy parameters.

We first define metrics to explicitly measure the distribution and number of specialized and general policy parameter sets for a given robot design. We then tested whether those metrics actually correlate with sample efficiency for different training algorithms, and found that they do for all of the tested algorithms. We show how robot designs that facilitate learning can be found by co-optimizing policy and sensor distribution parameters. We found that co-optimization results in significantly better sample efficiency than policy optimization alone. Lastly, we show that the designs found during co-optimization follow a sensor homeostasis gradient. This leads to the hypotheses that part of what allows an agent's body to facilitate policy search is its ability to facilitate sensor homeostasis, and that including homeostasis in future loss functions may further improve sample efficiency in multi domain policy and agent design training.

%%%%%%%%%%%%%%%%%%%%%%%%%%%%%%%%%%%%%%%%%%%%%%%%%%%%%%%%%%%%%%%%%%%%%%%%%%%%%%%%

\section{Methods}
\subsection{Experimental Overview}
\label{sec:neurips_overview}
A minimal agent, task, and environment were employed to test the hypothesis that co-optimizing policy and structure parameters increases sample efficiency during training: agents are trained to approach a light source, and they are placed in environments with differing light source locations.

\begin{figure}
    \centering
    \includegraphics[width=0.3\linewidth]{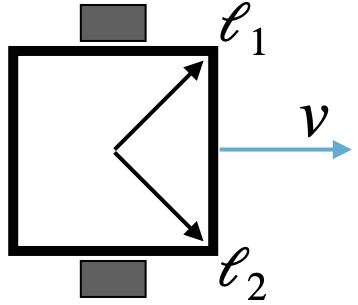}
    \caption{The robot is modeled as having a square frame with two separately-driven wheels and two infrared sensors located on the dorsal surface of the frame at locations defined by two vectors emanating from the center of the frame $\ell_1$ and $\ell_2$. The vector $v$ shows the forward direction. This design is a hand designed baseline based on a Braitenberg Vehicle \cite{braitenberg1986vehicles}.}
    \label{fig:neurips_robot}
\end{figure}

The robot is modeled as having a square frame (edge length $0.5m$) with two separately-driven wheels and two infrared sensors. A baseline, hand designed robot can be seen in Fig.~\ref{fig:neurips_robot}.
The sensors detect light according to the inverse square law, i.e., $1/d^2$, where $d$ is the distance from the light source and are modeled without occlusion. The motors driving the wheels are contralaterally connected to the sensors by weighted synapses yielding two trainable parameters $w_1, w_2 \in [-1.0, 1.0]$.

In this paper we explore change to a single design attribute: the physical location of the two sensors, which can be placed anywhere on the dorsal surface of the robot's square body.
The location of the $i$-th sensor $\ell_i$ can be described by its Cartesian coordinates $\ell_i = (x, y)$, where $x, y \in [-0.5, 0.5]$, and (0, 0) denotes the center of the body (Fig.~\ref{fig:neurips_robot}).

If we define $\theta$ as the space of all possible synapse weight pairs ($w_1$, $w_2$), we can investigate how a given sensor distribution ($\ell_1$, $\ell_2$) affects $\theta$.
Since we cannot perform an exhaustive sweep over the infinitude of possible sensor positions, we discretize each dimension of $\ell_i$ into nine uniformly-spaced bins. Because sensors are varied in two dimensions ($x$ and $y$) there are $9^2=81$ possible locations for each sensor. Since there are two sensors, and each can be in 81 possible discrete locations, the total design space consists of $81^2=6561$ possible robot designs.

For each of these 6561 designs, we conducted another sweep over the synapse weights $(w_1, w_2)$, likewise discretizing each weight into 121 evenly-spaced values within $[-1,1]$, yielding $121^2 = 14641$ different weight controllers per robot design. For each of the $6561\times14641=96059601$ robot design and policy pairs, we evaluated the simulated robot for a phototaxis task across four environments as described below in \S\ref{sec:neurips_simulation}. Doing a grid search as described creates an exhaustive investigation of the weight space ($\theta$) for each robot design, thus allowing an analysis as described in \S\ref{sec:neurips_metrics} to measure the predicted ability of a robot design to facilitate policy training. We then  applied four different training algorithms to each robot design (see \S\ref{sec:neurips_training}) to test for any correlation between the utility of a given design predicted by our metrics and actual performance from training policies on it. Finally, we compare policy optimization using the default design against simultaneous co-optimization of design and policy (see \S\ref{sec:neurips_cooptimization}), and analyze the resulting optimized designs.

All of the experiments were run in parallel on a cpu computing cluster that is powdered by $95\%$ renewable energy. In total the experiments took $564,786$ cpu hours.

\subsection{Simulating the robot.}
\label{sec:neurips_simulation}
The robot is trained on a phototaxis task across four environments: the robot should move toward a light source in each environment. In each environment a light is placed in the center of an $(x,y)$ plane and the robot is placed diagonally at a distance of $8$ body lengths away from the light source; specifically at positions $(d, d)$, $(d, -d)$, $(-d, d)$, $(-d, -d)$ where $d=\frac{4}{\sqrt{2}}$ constitutes the four differing environments. Success in an environment is determined by how close the robot comes to the light source. We assume the light source has a radius ($r=0.075m$) that determines its physical size, and maximum loss is defined to be reached if the robot touches the light source during the course of its evaluation.

Given the robot's and environment's simplicity, rather than using a simulated robot and task environment, the location and orientation of the robot in the plane can be defined by a system of differential equations, where the change in position and orientation is determined by the change in light captured by two sensors. Ignoring deviations from the idealized environment, such as sensor noise and friction, the rate of angular and linear velocities will be proportional to a linear combination of the sensor values.

Let $\alpha(t)$ denote the angle of the robot at time $t$, where $\alpha=0$ denotes the positive $x$ direction, and $\phi(t) = (x(t), y(t))$ denote the position of the robot in the world. If the robot is located at the origin and facing east ($\alpha = 0$), its two light sensors are located exactly at $\ell_1$ and $\ell_2$, and they each capture some amount of light $s_1(t)$ and $s_2(t)$, respectively. 

Hence the absolute position of the $i$-th sensor is $\phi(t) + R_\alpha \ell_i^{\hspace{3pt}T}$, where
	\begin{equation}
		R_\alpha = \begin{bmatrix}
			\cos \alpha & -\sin \alpha \\
			\sin \alpha & \cos \alpha \\
		\end{bmatrix}
	\end{equation}
is the two-dimensional counterclockwise rotation matrix (in the amount $\alpha$). 

Since we formulate the problem such that it is the robot's initial position and heading that is adjusted in each environment instead of the position of the light source, we can assume that the light source is always at the origin. Then, the distance of $\ell_i$ from the light source is given by:  \mbox{$\|\phi(t) + R_\alpha \ell_i^{\hspace{3pt} T}\|$.}
And since the intensity of light is inversely proportional to the square of the distance, the sensor values are given by: 
\begin{equation}
\label{eq:neurips_sensor_value}
s_i(t) = \|\phi(t)^T + R_\alpha {\ell_i}^T\|^{-2},
\end{equation}

The robot turns based on the difference between the sensor values multiplied by the two neural network weights, the velocity of the robot is equal to the average of the two sensor values. 
Thus, the following determines the location and orientation of the robot:
	\begin{equation}
	    \label{eq:neurips_ode}
		\begin{cases}
			\dot{x} = v(t) \cos \alpha	\\
			\dot{y} = v(t) \sin \alpha \\
			\dot{\alpha} = w_1s_1(t) - w_2s_2(t), \\
		\end{cases}
	\end{equation}
where $v$ is the velocity of the robot given by $2v(t) = w_1s_1(t)+w_2s_2(t).$

\subsection{Metrics}
\label{sec:neurips_metrics}
We here define two metrics $M_L$ and $M_{CI}$ that are measured over the $K=4$ environments. These metrics measure how a robot design impacts the weight space of the control policy and consequently provide a prediction of sample efficiency, if policy training were performed using that design. $M_L$ measures controller learnability: how easy it would be to learn a generalist controller. $M_{CI}$ measures resistance to catastrophic interference: the probability that starting training with an environment-specific controller will result in discovery of a generalist controller.

\begin{figure}
    \centering
    \includegraphics[width=0.3\linewidth]{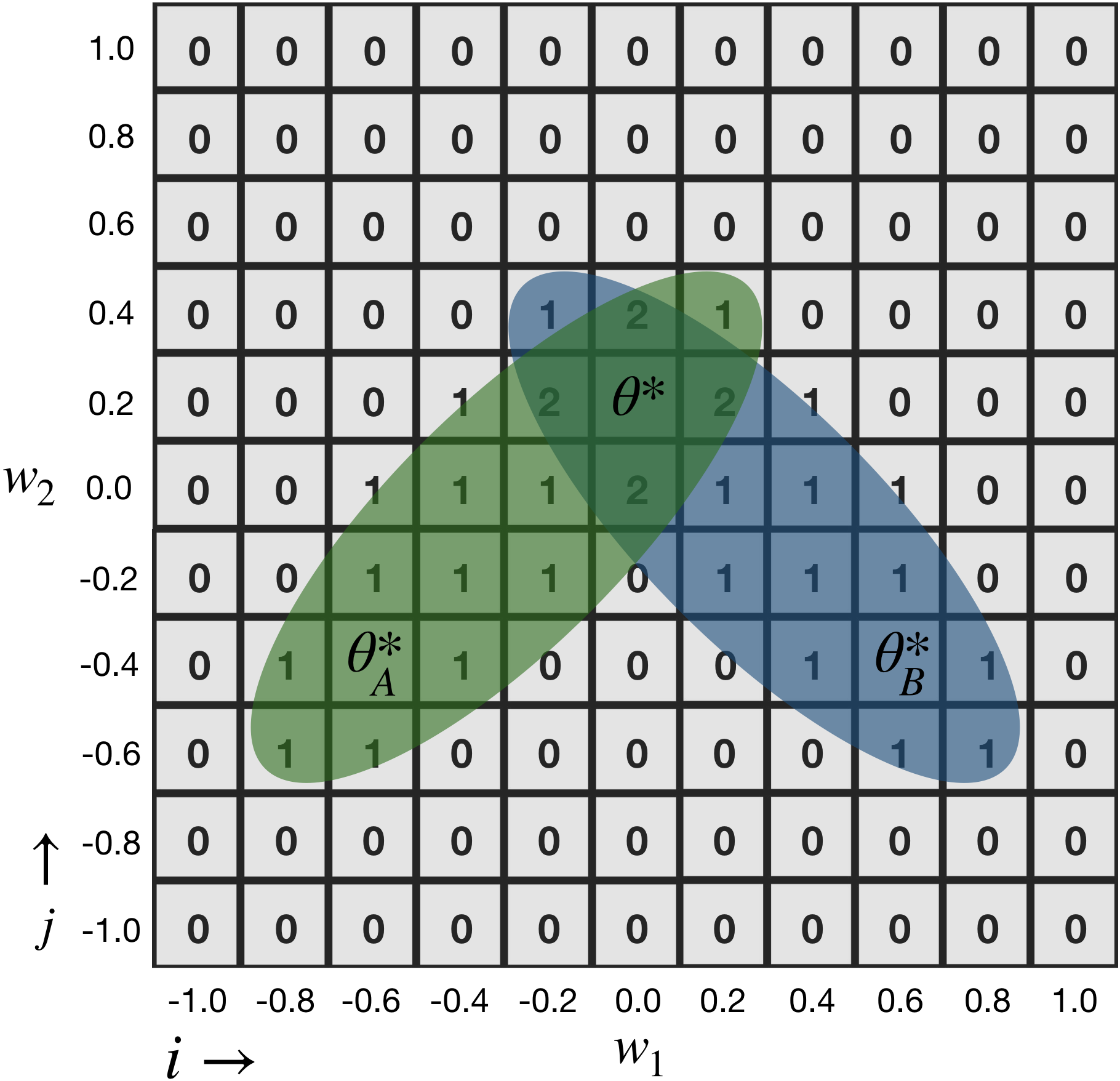}
    \caption{An example of overlapped binary success matrices for two environments A and B. Each element represents a different set of policy parameters. Generalist controllers lie within the intersection $\theta^*$ of successful environment-specific controllers $\theta_k^*$.}
    \label{fig:neurips_metrics}
\end{figure}

Given a design $(\ell_1, \ell_2)$ and environment $k$, a binary success matrix $S^{(k)}(\ell_1, \ell_2)$ is constructed such that each element $S^{(k)}_{i, j}(\ell_1, \ell_2)$ is either 1 (the light source was touched) or 0 (it was not) and each element corresponds to particular values of weights $w_1$ and $w_2$. By overlapping the success matrices for a fixed design across the four environments, we can visualize the manifolds $\theta_k^*$ where $k\in\{1, 2, 3, 4\}$ for the robot (Fig.~\ref{fig:metrics}).

We define the overlap $\mathcal{O}$ as an element-wise sum of the success matrices over each environment $k$:
\begin{equation}
    \label{eq:neurips_overlap}
    \mathcal{O} = \sum_{k=1}^K S^{(k)}(\ell_1, \ell_2) .
\end{equation}

The learnability metric is simply the proportion of $|K|=4$ values, which denote success in all environments, in the overlapped success matrices, compared to the entire matrix space:
\begin{equation}
    \label{eq:neurips_metric1}
    M_L = \frac{g_{|K|}(\mathcal{O})}{n^2},
\end{equation}
where $g_k$ is a function that counts the total elements of a matrix with value equal to $k$ and $n$ is the square dimension of the matrix defined by the discrete policy parameter set sweep.

Resistance to catastrophic interference is measured by:
\begin{equation}
    \label{eq:neurips_metric2}
    M_{CI} = \begin{cases}
    0 \; \text{ if } \mathcal{O} \text{ is a null matrix}, \\
    g_{|K|}(\mathcal{O}) \left[\sum_{k=1}^K g_k(\mathcal{O})\right]^{-1}
    \; \text{otherwise}.
    \end{cases}
\end{equation}
which is the number of policy parameter sets that solved all $K$ environments divided by the number of sets that solved at least one.

\subsection{Training}
\label{sec:neurips_training}
We used four different training methods to tune the weights of the control policies: random search, generating set search, separable natural evolution strategies (NES), and differential evolution (DE). These methods were chosen because they encompass different paradigms in optimization with the first (random search) serving as a baseline. Generating set search is considered a powerful derivative free algorithm from the direct search paradigm \cite{kolda2003optimization}. Natural evolution strategies is another derivative free method loosely based on biological evolution \cite{wierstra2014natural}. Differential evolution is also based on biological evolution but is generally considered to be in a different class of evolutionary based optimizers than NES \cite{price2013differential}. Thus, together, these methods give motivation that the results from this research extend beyond the actual methods used\footnote{All of the search methods were used as is from \url{https://github.com/robertfeldt/BlackBoxOptim.jl} using the default hyper parameters from the code base.}.

We applied each training method to each of the $6561$ robot designs, five times using different initial random policies. For each method the loss function was computed as the sum of the minimum distance from the light source achieved in every environment. During training we also kept track of the total number of environments in which the robot was successful. We then compute the average number of evaluations it took to achieve success in all environments for a given training method and plot them against the designs' metrics $M_L$ and $M_{CI}$ to measure the Pearson correlation and corresponding p-value. A strong negative correlation is desirable here: designs with higher metrics require a lower number of evaluations to achieve optimality. This would indicate that good designs predicted by the metrics yielded significantly better sample efficiency.

\subsection{Co-Optimization}
\label{sec:neurips_cooptimization}
To simultaneously optimize both the controller and sensor placement of the robot we use the Borg Multiobjective Evolutionary Algorithm (Borg MOEA) \cite{hadka2013borg}. As before the goal of the algorithm is to produce a policy that works in all four environments in as few evaluations as possible. However rather than optimize directly on the sum of score in each environment we treat each of these scores as it's own objective, and pick the best solution at the end based on the aggregate sum of the objectives. The algorithm works on 6 trainable parameters: the 2 values for the first light sensor vector, the 2 values for the other light sensor vector, and the 2 weight values of the controller.

We compare this to the same algorithm, but fix the design of the robot to the baseline design in Fig.~\ref{fig:neurips_robot}. In this baseline the algorithm only optimizes 2 trainable parameters which are the 2 weights of the controller.

We run each algorithm over 30 different random seeds and take the average number of evaluations to required to achieve success in all four environments. We then run a Mann Whitney U Test to compare if there were any significant differences between the two methods and report a corresponding p-value.

Lastly, we analyze the sensor values experienced by designs found over the course of co-optimization with dynamic time warping (DTW) \cite{giorgino2009computing}. The dynamic time warping algorithm is a state-of-the-art way to measure the similarity between two signals by determining the cost to align the signals via stretching and/or shifting. For each environment we record a vector of the light sensor signals experienced (four in total). For one of the light sensor we compute the average DTW distance between all combinations of experienced signals (six comparisons) to get a signal score for the light sensor, we then do the same for the other light sensor. We average both of those scores to get the aggregate DTW distance for the design, the lower this number is the more similar the light sensor signals were between environments.

%%%%%%%%%%%%%%%%%%%%%%%%%%%%%%%%%%%%%%%%%%%%%%%%%%%%%%%%%%%%%%%%%%%%%%%%%%%%%%%%

\begin{figure*}[ht]
    \centering
    \includegraphics[scale=0.5]{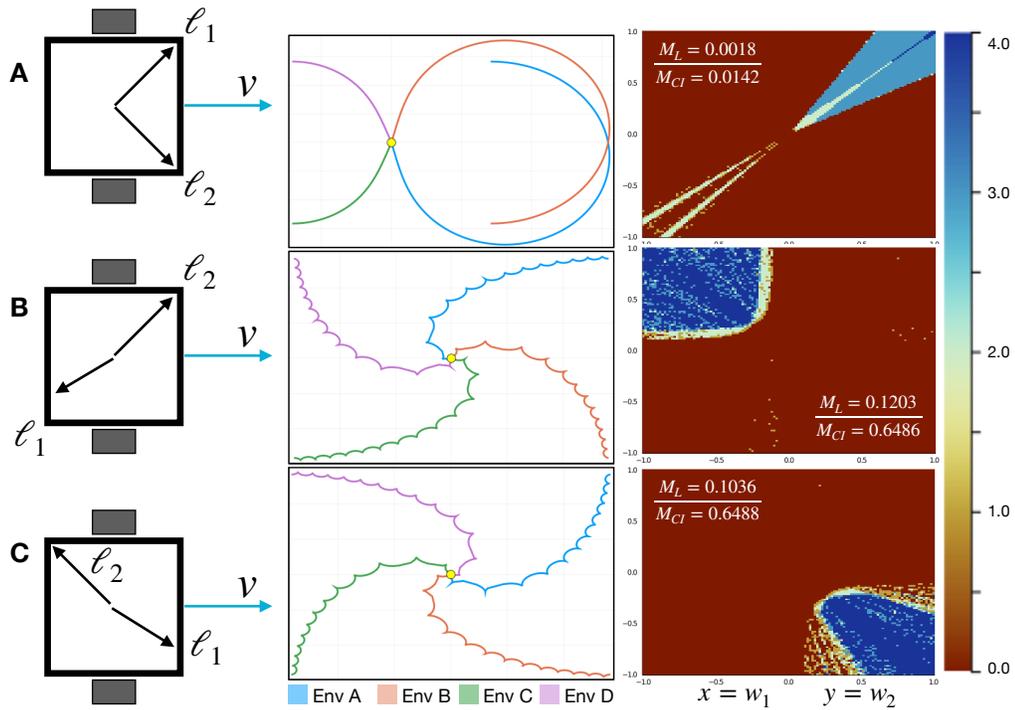}
    \vspace{-30pt}
    \caption{Examples of different robot designs with corresponding policy trace, success matrices, and metrics. The vectors $\ell_1$ and $\ell_2$ show the placement of the light sensors for a specific design. The colors in the second column show the trace of the $(x, y)$ coordinates of the robot in the four different starting positions as it moves toward the light source represented by the yellow point. The last column shows the weight assignments that solve different environments; red assignments solve no environments, while any other color solves at least one, and blue assignments solve all four environments. \textbf{(A)} The baseline morphology as shown in Fig.~\ref{fig:neurips_robot}. \textbf{(B)} The design with the best learnability metric ($M_L$). \textbf{(C)} The design with the best resistance to catastrophic forgetting ($M_{CI}$)}
    \label{fig:neurips_metrics_results}
\end{figure*}

\begin{figure*}[ht]
    \centering
    \includegraphics[width=0.5\linewidth]{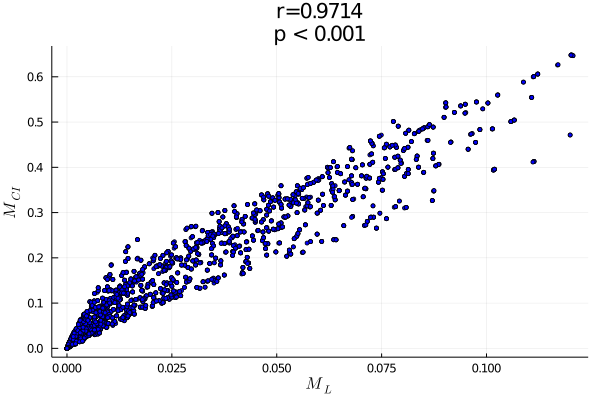}
    \caption{The correlation between design learnability and resistance to catastrophic interference. Each point is a design with its learnability on the x-axis and resistance on the y-axis. The two variables are strongly correlated in this case. The density of points is more concentrated near the origin showing that high scoring designs are relatively rare.}
    \label{fig:neurips_ml_mci_correlation}
\end{figure*}

\begin{figure*}[ht]
    \centering
    \textbf{Random Search}\\
    \includegraphics[width=0.40\linewidth]{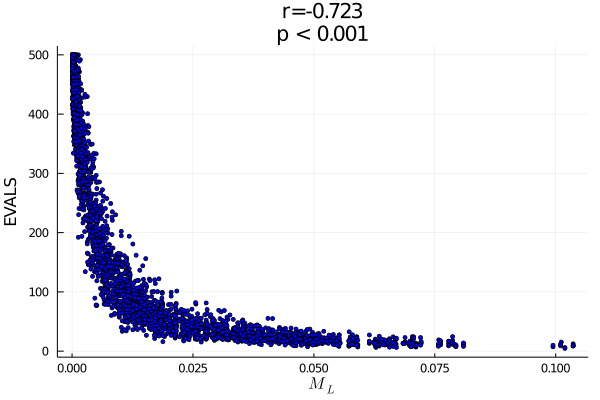}
    \includegraphics[width=0.40\linewidth]{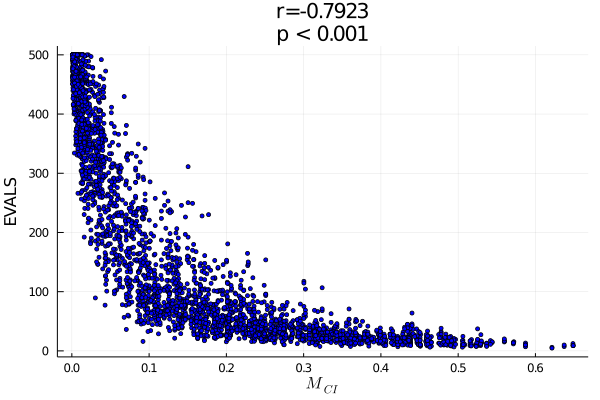}
    
    \textbf{Generating Set Search}\\
    \includegraphics[width=0.40\linewidth]{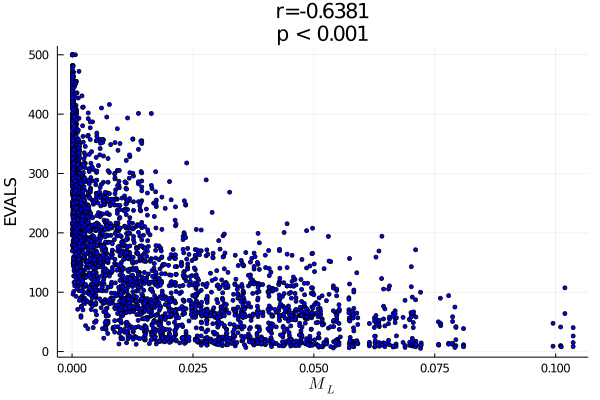}
    \includegraphics[width=0.40\linewidth]{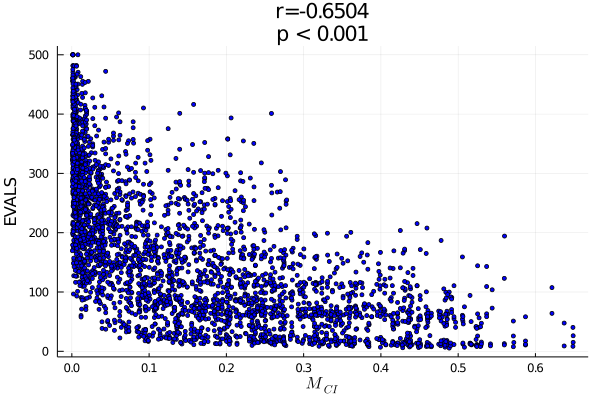}
    
    \textbf{Natural Evolution Strategy}\\
    \includegraphics[width=0.40\linewidth]{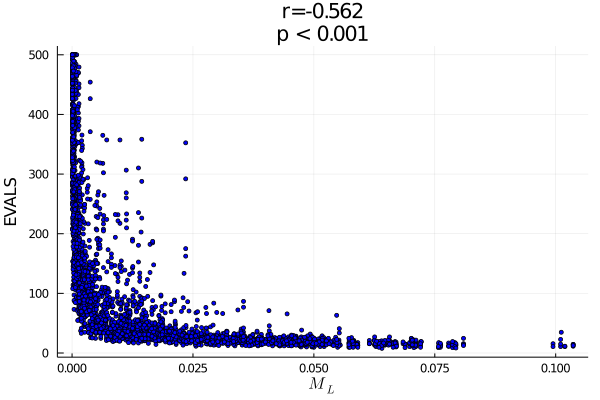}
    \includegraphics[width=0.40\linewidth]{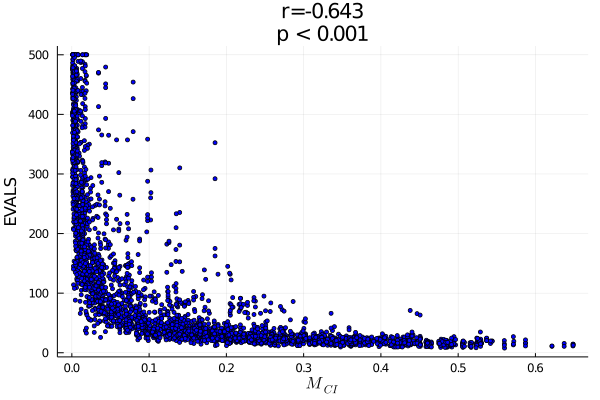}
    
    \textbf{Differential Evolution}\\
    \includegraphics[width=0.40\linewidth]{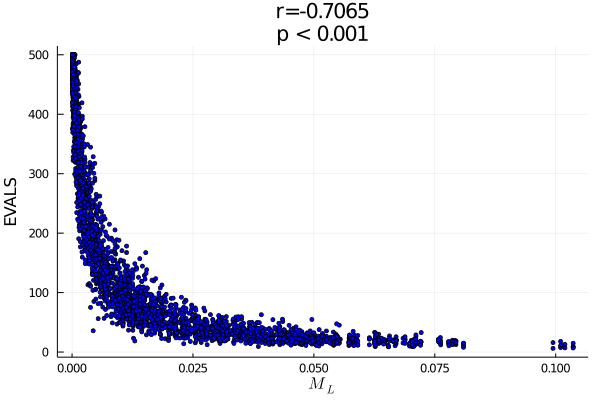}
    \includegraphics[width=0.40\linewidth]{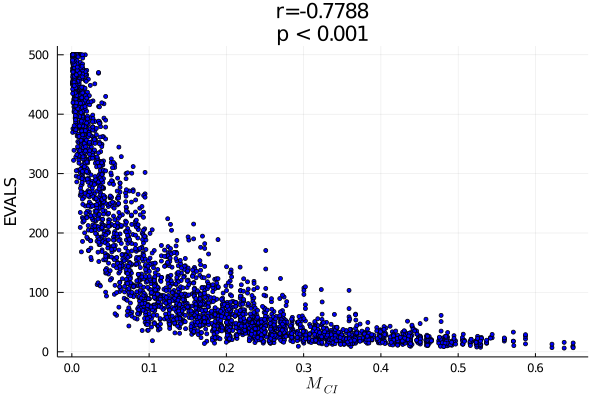}
    \vspace{-10pt}
    \caption{For each design we plot sample efficiency against the design's learnability ($M_L$) and resistance to interference ($M_{CI}$) scores across the search methods. In each case there is a significant negative correlation demonstrating that a design's score is predictive of the required effort to train a policy for that design.}
    \label{fig:neurips_evolution_correlation}
\end{figure*}

\begin{figure*}[ht]
    \centering
    \includegraphics[width=0.45\linewidth]{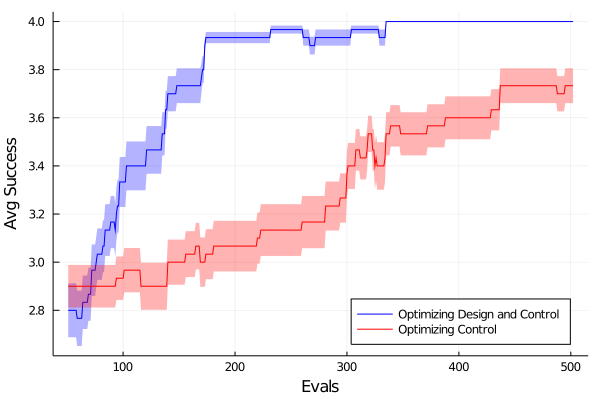}
    \includegraphics[width=0.45\linewidth]{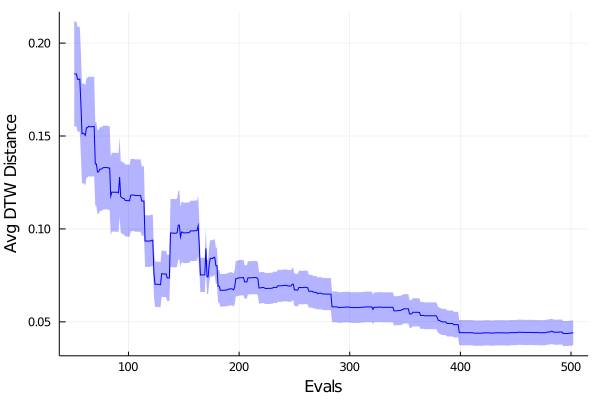}
    \caption{\textbf{(Left)} Average number of environments a policy is successful in (Average Success) as a function of evaluation time, for policy optimization using the baseline design (red) and co-optimization of design and control (blue). Co-optimization was significantly better that controller optimization alone. \textbf{(Right)} The average Dynamic Time Warping (DTW) distance between the robots' sensors in all of the environments for the robot designs found during co-optimization. As optimization progresses, designs progress from those that experience environments differently (high DTW) to those that experience differing environments more similarly (low DTW). Shading in both panels reports the $95\%$ confidence interval.}
    \label{fig:neurips_cooptimization}
\end{figure*}

\section{Results}
We employed the DifferentialEquations.jl package \cite{christopher_rackauckas_2021_4785349} for numerical integration of the robot's location and orientation using standard Euler Integration with a time step of $0.1$ for $10^5$ time steps.

For each evaluated mechanical design and policy (sensor locations and synapse weights), the robot's trajectory is computed in each of the four environments defined in \S\ref{sec:neurips_simulation}. Again, if robot's trajectory comes within 0.075 meters of the light source, the robot is determined to have succeeded in that environment. Otherwise, it is determined to have failed.

The mechanical design sketched in Fig.~\ref{fig:neurips_metrics_results}B had the highest learnability score, with $M_L = 0.1206$. A very similar design (Fig.~\ref{fig:neurips_metrics_results}C) scored the best in resistant to catastrophic interference: the proportion of resistant to nonresistant policies for that design was $M_{CI}=0.6479$. In general the best policies for both metrics featured a specific similar design feature of asymmetric placement of the light sensors. The baseline design (Fig.~\ref{fig:neurips_metrics_results}A) had a much lower learnability ($M_L=0.0018$) and resistance to catastrophic interference ($M_{CF}=0.0142$) then the best found asymmetrical designs.

The worst designs for both metrics have scores of zero and feature designs where the lights sensors are both placed in the same location or extremely close to one another. This makes particular sense because the closer the light sensors are the less differential there is between signals thus making it difficult or impossible (in the case of overlapping placement) for the robot to distinguish between different environments with out some sore of behavioral memory mechanism such, as a recurrent network.

In Fig.~\ref{fig:neurips_metrics_results} the successes matrices of weight manifolds for all the mentioned designs can be seen in detail, where dark blue represents weight assignments that succeed in all for environments for a given design. These weight manifolds show clearly that in this case the optimal weight assignments for the asymmetrical designs should be much easier to find during training, while the baseline design is akin to looking for a needle in a haystack. This also shows that for this specific robot it appears a high score in one metric correlates with the others. We specifically tested this by plotting the metric scores for all of the designs along with the correlation between the two metrics. As shown in Fig.~\ref{fig:neurips_ml_mci_correlation} there is a significant positive correlation ($p < 0.001$) between the metrics, thus giving an explanation for why the best designs under both metrics are so similar.

We can also see that based on the frequency of metrics $M_L$ and $M_{CF}$ that there are many designs (including intuitive symmetric ones) that score poorly on $M_L$ and $M_{CF}$ while there are relatively few designs that perform well.

When we conducted the training of the policies for each of these designs as described in \S\ref{sec:neurips_training}, we found that a design's score over the metrics was significantly correlated with the number of evaluations required to solve the phototaxis task for all four environments. This can be seen in Fig.~\ref{fig:neurips_evolution_correlation}. On the x-axis the metric is plotted and the y-axis is the average number of evaluations required to achieve success in all four environments, thus we see a strong negative correlation across the methods with a p-value less that $0.001$, meaning that a design's metric score was predictive of the required effort to train that design on the task.

Lastly, we compared co-optimization of the policy and robot design as described in \S\ref{sec:neurips_cooptimization} versus policy optimization of the baseline design. As can be seen in Fig.~\ref{fig:neurips_cooptimization}, co-optimization significantly outperformed ($p < 0.001)$ policy optimization alone, despite increased complexity in the dimension of optimizable parameters. In fact, in the number of evaluations attempted, optimizing the policy for the baseline design never achieved competence in all four environments, and thus, that design ultimately succumbed catastrophic interference. We further analyzed the designs found by co-optimization (Fig.~\ref{fig:neurips_cooptimization}) and discovered that they progressively move from designs with high DTW distance scores to ones with low DTW scores. Thus without any guidance, the optimization naturally found a homeostatic design (one that sensorally experiences differing environments similarly) or a design which was best suited to optimization in multiple environments.

%%%%%%%%%%%%%%%%%%%%%%%%%%%%%%%%%%%%%%%%%%%%%%%%%%%%%%%%%%%%%%%%%%%%%%%%%%%%%%%%

\section{Discussion}
Here, a simple robot and task were employed to investigate the entire loss landscape of the weight manifold at high resolution. In our sweep across policy parameter sets for many robot designs, the results suggested that changes in one element of a robot's design (sensor distribution) greatly altered the loss surface. We hypothesized that this is likely to influence sample efficiency during policy optimization for a given design, and that design's resistance to catastrophic interference. This hypothesis was strengthened by the observation that changing sensor location induced changes in the number and placement across the loss surface of specialized parameter sets optimal for one environment, as well as the size of specialist overlap regions containing optimal generalized parameter sets that confer resistance to catastrophic interference.

We confirmed this hypothesis for four orthogonal training methods: we found that a robot design's learnability and resistance scores correlated with its sample efficiency, for this phototaxis task.

Lastly, we demonstrated that these high scoring designs could be found via simultaneous co-optimization of robot design and control, and that doing so was significantly more sample efficient than optimizing just the policy for a manually designed baseline robot. We concluded with evidence showing sensor homeostasis increased as robot designs improved, even though the loss function does not reward homeostasis. This suggests a causal link between agent design and sample efficient policy search: some designs better enable a robot to move such that observations converge across seemingly different environments, and allow it to perform the same, appropriate actions when aligned appropriately in each environment.

Even in the seemingly trivial case of phototaxis with contralateral connections on a minimal robot as first described by \cite{braitenberg1986vehicles}, we found that when challenged to perform in multiple environments, adjustments in design, specifically to sensor location, could either suppress or exacerbate catastrophic interference by expanding or shrinking the overlap of optimal policy parameters for that design across different task environments.

The physical location of sensors is thus a relevant property of robots nevertheless abstracted away in the mostly disembodied systems that address catastrophic interference reported in the literature to date. While sensor location could in principle be dynamically controlled via a lattice of sensors \cite{kramer2011wearable} or adjustable antenna \cite{fend2003active}, change in (and rational control over) other design attributes such as geometry \cite{kriegman2019automated}, material properties \cite{narang2018transforming}, or the number and placement of actuators \cite{lipson2000automatic} is much more difficult in practice, and such design elements are almost always presupposed and fixed prior to training \cite{cheney2018scalable}.

Overall we found that machines with the right design may use a single policy to succeed at a set of tasks that appear different to a robot with an ill suited body plan. The degree to which design influences learnability in more complex robots, task environments and behaviors has yet to be investigated, but will be the focus of future work.

\chapter{A soft robot that adapts to environments through shape change}
\begin{abstract}
Many organisms, including various species of spiders and caterpillars, change their shape to switch gaits and adapt to different environments. Recent technological advances, ranging from stretchable circuits to highly deformable soft robots, have begun to make shape-changing robots a possibility. However, it is currently unclear how and when shape change should occur, and what capabilities could be gained, leading to a wide range of unsolved design and control problems. To begin addressing these questions, here we simulate, design, and build a soft robot that utilizes shape change to achieve locomotion over both a flat and inclined surface. Modeling this robot in simulation, we explore its capabilities in two environments and demonstrate the existence of environment-specific shapes and gaits that successfully transfer to the physical hardware. We found that the shape-changing robot traverses these environments better than an equivalent but non-morphing robot, in simulation and reality.
\end{abstract}

\section{Introduction}
Nature provides several examples of organisms that utilize shape change as a means of operating in challenging, dynamic environments.
For example, the spider Araneus Rechenbergi~\cite{jager_cebrennus_2014,bhanoo_desert_2014} and the caterpillar of the Mother-of-Pearl Moth (Pleurotya ruralis)~\cite{armour_rolling_2006} transition from walking gaits to rolling in an attempt to escape predation. Across larger time scales, caterpillar-to-butterfly metamorphosis enables land to air transitions, while mobile to sessile metamorphosis, as observed in sea squirts, is accompanied by radical morphological change. Inspired by such change, engineers have created caterpillar-like rolling~\cite{lin_goqbot:_2011}, modular~\cite{christensen2006evolution, yim2007modular, parrott2018hymod}, tensegrity~\cite{paul2006design, sabelhaus2015system}, plant-like growing~\cite{sadeghi_toward_2017}, and origami~\cite{miyashita2015untethered, rus2018design} robots that are capable of some degree of shape change. However, progress toward robots which dynamically adapt their resting shape to attain different modes of locomotion is still limited. Further, design of such robots and their controllers is still a manually intensive process.

% Machines in multiple environments
Despite the growing recognition of the importance of morphology and embodiment on enabling intelligent behavior in robots~\cite{pfeifer_self-organization_2007}, most previous studies have approached the challenge of operating in multiple environments primarily through the design of appropriate control strategies. For example, engineers have created robots which can adapt their gaits to locomote over different types of terrain~\cite{saranli2001rhex, raibert2008bigdog, kuindersma2016optimization}, transition from water to land ~\cite{ijspeert2007swimming, li2015design}, and transition from air to ground~\cite{myeong2015development, bachmann2009biologically, roderick2017touchdown}. Other research has considered how control policies should change in response to changing loading conditions~\cite{korayem2010dynamic, li2015discrete}, or where the robot's body was damaged~\cite{bongard2006resilient, cully2015robots, chatzilygeroudis2018reset}. Algorithms have also been proposed to exploit gait changes that result from changing the relative location of modules and actuators~\cite{rosendo_trade-off_2017}, or tuning mechanical parameters, such as stiffness~\cite{garrad_shaping_2018}. In such approaches, the resting dimensions of the robot's components remained constant. These robots could not, for instance, actively switch their body shape between a quadrupedal form and a rolling-optimized shape.

% Soft robotics
The emerging field of soft robotics holds promise for building shape-changing machines~\cite{hauser_resilient_2019}. For example, one robot switched between spherical and cylindrical shapes using an external magnetic field, which could potentially be useful for navigating internal organs such as the esophagus and stomach~\cite{yim2012shape}. Robotic skins wrapped around sculptable materials were shown to morph between radially-symmetric shapes such as cylinders and dumbbells to use shape-change as a way to avoid obstacles~\cite{shah2019morphing}. Lee et al. proposed a hybrid soft-hard robot that could enlarge its wheels and climb onto step-like platforms~\cite{lee_origami_2017}. A simulated soft robot was evolved to automatically regain locomotion capability after unanticipated damage, by deforming the shape of its remnant structure~\cite{kriegman2019automated}. With the exception of the study by Kriegman et al.~\cite{kriegman2019automated}, control strategies and metamorphosis were manually programmed into the robots, thereby limiting such robots to shapes and controllers that human intuition is capable of designing. However, there may exist non-intuitive shape-behavior pairings that yield improved task performance in a given environment. Furthermore, manufacturing physical robots is time-consuming and expensive relative to robot simulators such as VoxCad~\cite{hiller2014dynamic}, yet discovering viable shape-behavior pairs and transferring simulated robots to functioning physical hardware remains a challenge. Although many simulation to reality (``sim2real'') methods have been reported~\cite{jakobi1995noise, lipson2000automatic, bongard2006resilient, koos2013transferability, cully2015robots, bartlett20153d, rusu_sim--real_2017, chebotar_closing_2019, peng2018sim,  hwangbo2018learning, hiller2012automatic},
none have documented the transfer from simulation to reality of shape-changing robots.

\begin{figure}
    \centering
    \includegraphics[width=5 in]{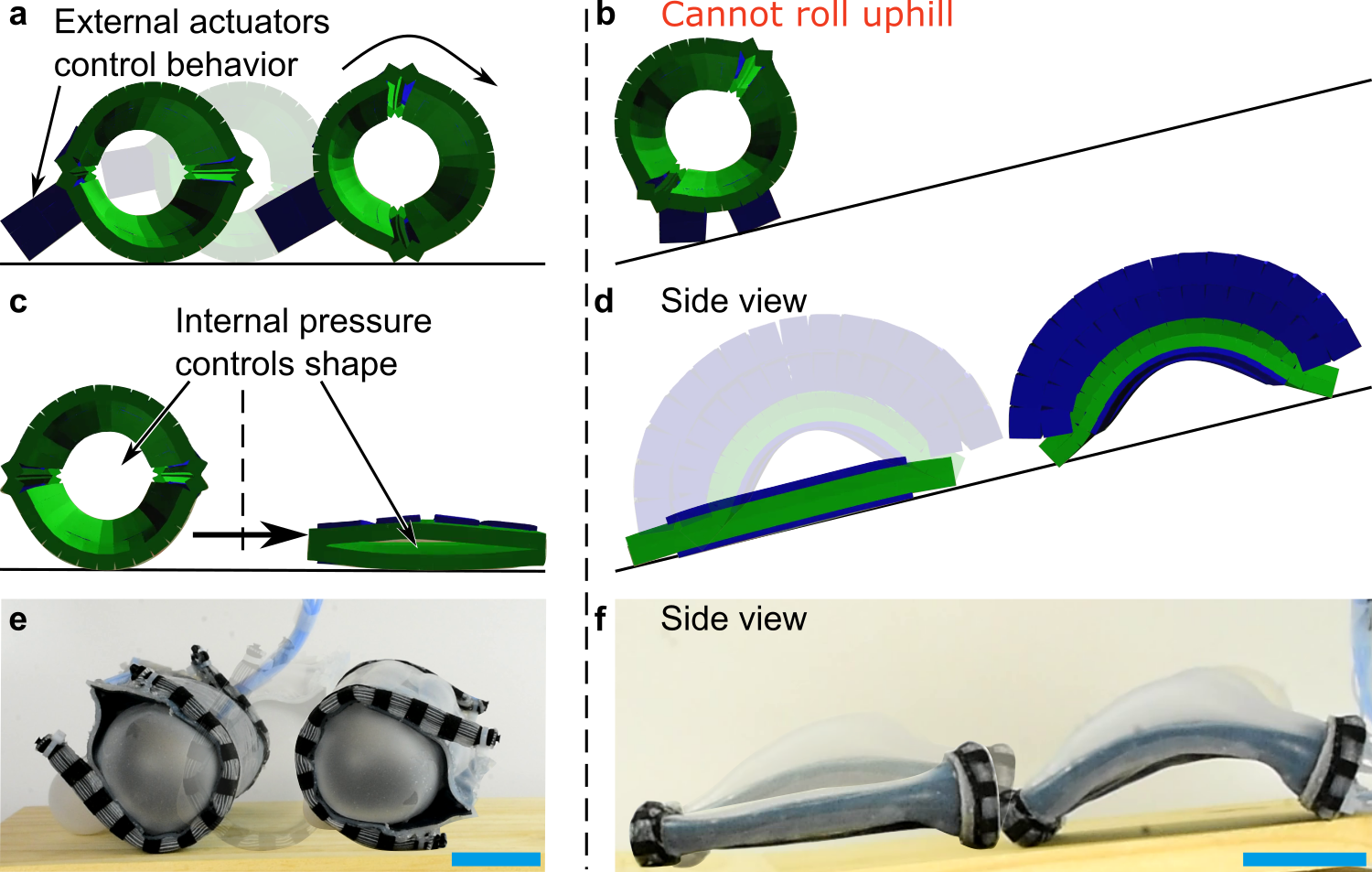}
    \caption{\textbf{Shape change can result in faster locomotion speeds than control adaptation, when a robot must operate in multiple environments.} \textbf{a,} Using inflatable external bladders, rolling was the most effective gait on flat ground. \textbf{b,} Rolling was ineffective on the inclined surface.
    We discovered
    a flat shape (achieved by deflating the inner bladder; \textbf{c}) and crawling gait (\textbf{d}) that allowed the robot to succeed in this environment. \textbf{e,f,} After
    we discovered
    these strategies in simulation, we transferred
    these
    strategies for rolling (\textbf{e}) and inchworm motion (\textbf{f}) to real hardware. Scale bars, 5 cm.}
    \label{fig: summary}
\end{figure}

\begin{figure}
    \centering
    \includegraphics[width=5 in]{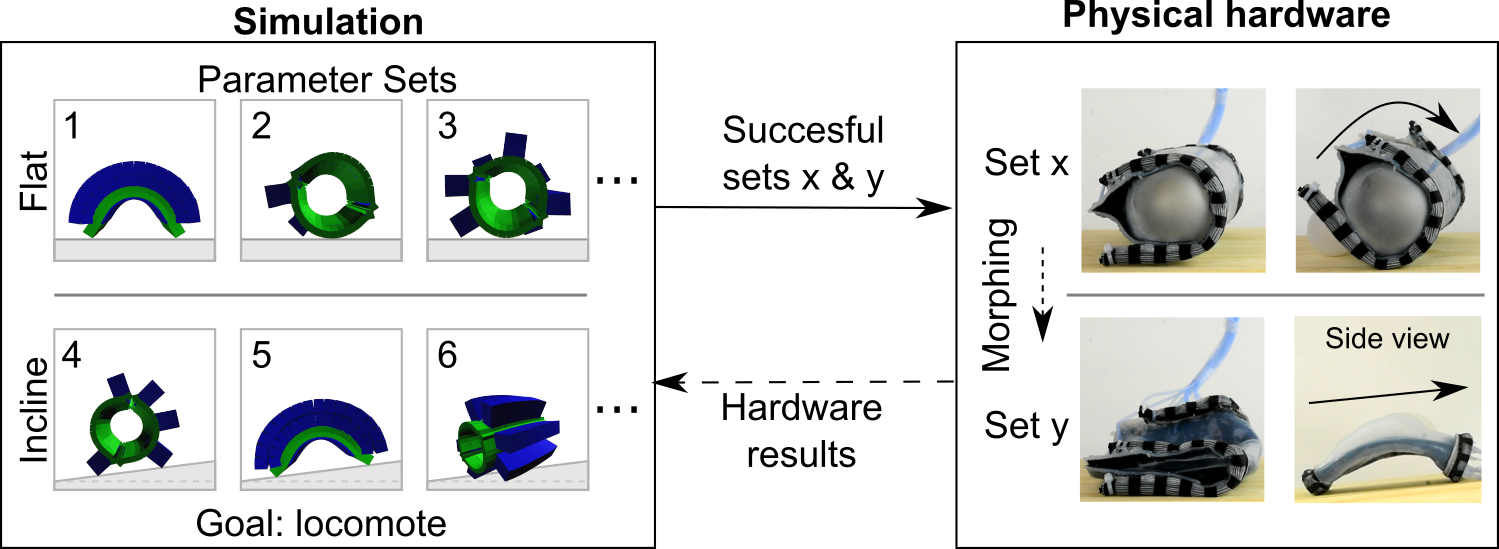}
    \caption{\textbf{Simulation revealed successful shapes and controllers, which we attempted to realize in hardware.} Sets consisting of a shape, an orientation, and a controller were generated for the robot in simulation. Each numbered sub-panel depicts a single automatically generated parameter set. After running simulations to determine the speed of each set, some were deemed too slow, while successful (relatively quick) sets were used to design a single physical robot that could reproduce the shapes and gaits found in simulation for both environments. During prototyping, actuator limits were measured and incorporated into the simulator to improve the accuracy of the simulation.}
    \label{fig: sim2real}
\end{figure}

% describe problem set-up more and what we are trying solve
To test whether situations exist where shape change improves a robot's overall average locomotion speed within a set of environments more effectively than control adaptations, here we present a robot which actively controls its shape to locomote in two different environments: flat and inclined surfaces (Fig.~\ref{fig: summary}). The robot had an internal bladder, which it could inflate/deflate to change shape, and a single set of external inflatable bladders which could be used for locomotion. Depending on the core's shape, the actuators created different motions, which could allow the robot to develop new gaits and gain access to additional environments. Within a soft multi-material simulator, an iterative ``hill-climbing'' algorithm~\cite{mitchell_when_1994} generated multiple shapes and controllers for the robot, then automatically modified the robots' shapes and controllers to discover new locomotion strategies. No shape-controller pairs were found that could locomote efficiently in both environments. However, even relatively small changes in shape could be paired with control policy adaptations to achieve locomotion within the two environments. In flat and even slightly inclined environments, the robot's fastest strategy was to inflate and roll. At slopes above a critical transition angle,
we could increase the robot's speed by flattening it and programming it with an inchworm gait
to exhibit an inchworm gait. A physical robot was then designed and manufactured to achieve similar shape-changing ability and gaits (Fig.~\ref{fig: sim2real}). When placed in real-world analogs of the two simulated environments, the physical robot was able to change shape to locomote with two distinct environmentally-effective gaits, demonstrating that shape change is a physically realistic adaptation strategy for robots.

% Sim & real summary

% Sim2real loop

This work points toward the creation of a pipeline that takes as input a desired objective within specified environments, automatically searches in simulation for appropriate shape and control policy pairs for each environment, and then searches for transformations between the most successful shapes. If transformations between successful shapes can be be found, those shape/behavior pairs are output as instructions for designing the metamorphosing physical machine. Here, we demonstrate that at least some shape/behavior pairs, as well as changes between shapes, can be transferred to reality. Thus, this work represents an important step toward an end-to-end pipeline for shape-changing soft robots that meet the demands of dynamic, real-world environments.

%%%%%%%%%% The simulated robot. %%%%%%%%%%
\section{Results}
\subsection{The simulated robot.}
\begin{figure} % Simulation results in graph format
    \centering
    \includegraphics[width=\textwidth]{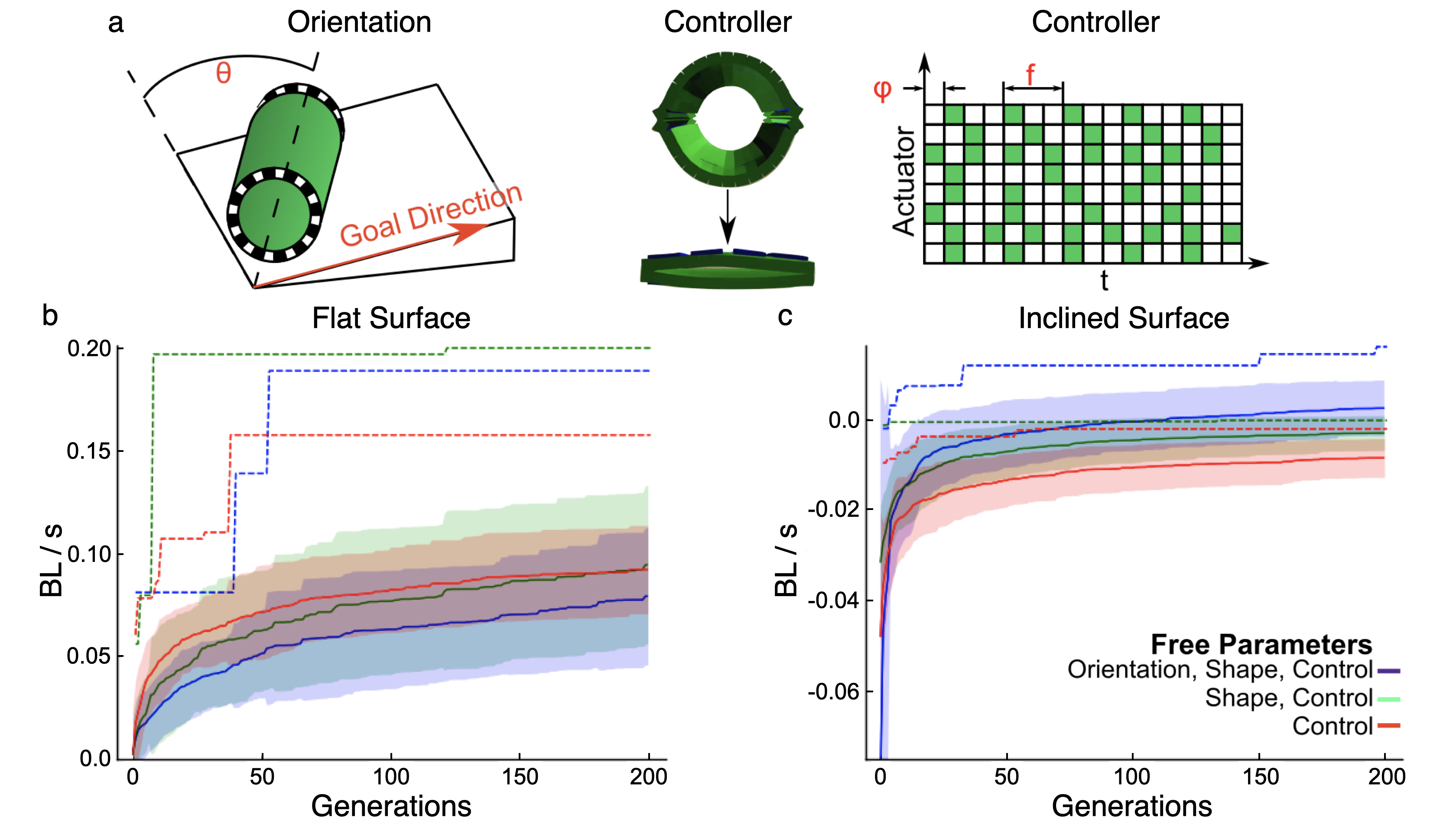}
    \caption{
    \textbf{Automated search attempting to improve gaits in both environments.}
    \textbf{a,} For each simulation, the algorithm could adjust the orientation, shape, and/or controller of the robot. Orientation ($\theta$) was measured by the angle between the robot's leading edge and a constant-elevation line on the surface. Shape was parameterized as the inner bladder's pressure, resulting in a family of shapes between the cylinder and flat shape shown. Control of each actuator was parameterized as the number of timesteps until its first actuation ($\phi$) and the number of timesteps between actuations ($f$). Here, we show an example controller for the eight main bladders, with green shaded squares illustrating inflation and white squares showing deflation.
    \textbf{b,} Results on a flat surface and \textbf{(c)} on an inclined surface. Shaded regions represent one standard deviation about the mean (solid line), while dashed lines represent maximum fitness. The legend indicates which parameters were to open to optimization, the others being held constant.
    }
    \label{fig:sim_results}
\end{figure}

% Intro to simulation
We sought
to automate search for efficient robot shapes and control policies in simulation, to test our hypothesis that shape and controller adaptation can improve locomotion speeds across changing environments more effectively when given a fixed amount of computational resources, as compared to controller adaption only. To verify that multiple locomotion gaits were possible with the proposed robot design, we first used our intuition to create two hand-designed shape and control policies: one for rolling while inflated in a cylindrical shape (Fig.~\ref{fig: summary}a), and the other for inchworm motion while flattened (Fig.~\ref{fig: summary}d). Briefly, the rolling gait consisted of inflating the trailing-edge bladder to tip the robot forward, then inflating one actuator at a time in sequence. The hand-designed ``inchworm'' gait consisted of inflating the four upward-facing bladders simultaneously to bend the robot in an arc. We then performed three pairs of experiments in simulation. Within each pair, the first experiment automatically sought robot parameters for flat ground; the second experiment sought parameters for the inclined plane. Each successive pair of experiments allowed the optimization routine to control an additional set of the robot's parameters, allowing us to measure the marginal benefit of adapting each parameter set when given identical computational resources (summarized in Table~\ref{tbl: simulation results} and Fig.~\ref{fig:sim_results}). The three free parameter sets of our shape-changing robot are shape, orientation relative to the contour (equal-elevation) lines of the environment, and control policy (Fig.~\ref{fig:sim_results}a). This sequence of experiments sought to determine whether optimization could find successful parameter sets in a high-dimensional search space, while also attempting to determine to what degree shape change was necessary and beneficial.

% Fitness
In all experiments, fitness was computed as the distance the simulated robot traveled in a desired direction for a fixed time period. To facilitate comparison with speeds achieved by the physical robot, the average speed of the simulated robot, when using hand-designed or evolved shapes and policies, was computed and is reported herein in body lengths/second (BL/s) attained over flat ground or uphill, depending on the current environment of interest, during a fixed period of time. 

Parameters for the simulation were initialized based on observations of previous robots~\cite{shah2019morphing,booth_omniskins:_2018}, and adjusted to reduce the simulation-to-reality gap after preliminary tests with physical hardware (see Methods for additional details). The results reported here are for the final simulations that led to the functional physically-realized robot and gaits.

\begin{table}[ht] % Multicolumn table https://tex.stackexchange.com/questions/188690/how-to-make-a-table-with-subcolumns-like-this
\small
  \caption{\textbf{Simulation results, reported as the mean and maximum velocity attained for each test condition. The simulator is deterministic, so no mean is reported for the hand-designed gaits (since they will always yield identical locomotion speed). Shape-change allowed the robot to switch between dissimilar locomotion gaits, outperforming the benchmark policies. Combined maximum was determined by averaging the maximum speed attainable in both environments. All values have units of body-lengths per second (BL/s).}}
  \label{tbl: simulation results}
  \begin{centering}
  \begin{tabular}{|p{5cm}|c|c|c|c|c|}
    \hline
    \multirow{2}{5cm}{Free parameters} & \multicolumn{2}{c|}{\textbf{Flat ground}} & \multicolumn{2}{c|}{\textbf{Hill}} & ~\\
    \cline{2-5}
    & Mean & Max & Mean & Max & Combined max\\
    \hline
    %Orientation, Shape, Control & 0.0790730346450634 & 0.188832647888239 & 0.00226589568399288 & 0.0158172698852103 & 0.1023249588867245\\
    Orientation, Shape, Control & 0.0791 & 0.1888 & 0.0023 & 0.0158 & 0.1023\\
    \hline
    %Shape, Control & 0.0945451890675152 & 0.199979314420894 & -0.00320097179706825 & -0.000426222040531251 & 0.0997765461901815\\
    Shape, Control & 0.0945 & 0.2000 & -0.0032 & -0.0004 & 0.0998\\
    \hline
    %Control & 0.0921274988912492 & 0.157649233065188 & -0.00870513768964238 & -0.00227232601960259 & 0.0776884535227925\\
    Control & 0.0921 & 0.1576 & -0.0087 & -0.0023 & 0.0777\\
    \hline
    %Hand-designed rolling & N/A & 0.175313607559892 & N/A & -0.574308147077608 & -0.199497269758858\\
    Hand-designed rolling & N/A & 0.1753 & N/A & -0.5743 & -0.1995\\
    \hline
    %Hand-designed inching & N/A & 0.0602816654437786 & N/A & 0.024556217516104 & 0.0424189414799413\\
    Hand-designed inching & N/A & 0.0603 & N/A & 0.0246 & 0.0424\\
    \hline
  \end{tabular}
  \end{centering}
\end{table}

% Summary of experiments
In the first pair of experiments, we sought to discover whether optimization could find any viable controllers within a constrained optimization space, which was known to contain the viable hand-designed controllers. Solving this initial challenge served to test the pipeline prior to attempting to search in the full search space, which has the potential to have more local minima. The shape and orientation were fixed (flat and oriented length-wise, $\theta=90{^\circ}$, for the inclined surface, cylindrical and oriented width-wise, $\theta=0{^\circ}$ for the flat surface). In the second pair of experiments, the algorithm was allowed to simultaneously search for an optimal shape and controller pair. Finally, in the third pair of experiments, all three parameter sets were open to optimization in both environments, allowing optimization the maximum freedom to produce novel shapes, orientations, and controllers for locomoting in the two different environments.
For each experiment, we ran 60 independent ``hill-climbers'' (instantiations  of the ``hill-climbing'' search algorithm~\cite{mitchell_when_1994}, not to be confused with a robot that climbs a hill) for 200 generations, thus resulting in identical resource allocation for each experiment (Fig.~\ref{fig:sim_results}b-c).
In addition, we ran a control experiment in which we fixed the shape of the robot to be fully inflated and oriented width-wise ($\theta=0{^\circ}$) for the inclined surface, to determine whether shape-change was necessary. %The best the robot could do was prevent itself from rolling backward, and it attained a fitness value of -0.001 BL/s. Comparing this last experiment to the others, over the inclined surface we find that given the same computational effort, the added parameters of shape and orientation increased the performance of the robot, despite the higher dimensionality of the search space.

When shape and orientation were set as fixed parameters, optimization found a rolling gait for the inflated shape on flat ground (max fitness 0.1576 BL/s; Table 1), but it was unable to discover a successful gait for the hill when in the deflated shape (max fitness -0.0023 BL/s). We were, however, able to manually program an inchworm gait (0.0246 BL/s) that enabled the deflated robot to climb the incline (Table~\ref{tbl: simulation results}). For reference, other robots that exclusively utilize inchworm gaits have widely varied speeds, ranging from 0.013 BL/s (for a 226 mm-long robot) ~\cite{felton_robot_2013}to 0.16 BL/s (for a centimeter-scale robot) ~\cite{lee_design_2011}.

For the second pair of experiments (orientation fixed), the best robots produced inflated shapes that rolled over flat ground (max fitness 0.2000 BL/s),
and similarly failed to produce inchworm motion on the inclined ground (max fitness -0.0004 BL/s).

In the last pair of experiments (all parameters open), the algorithm again discovered that cylindrical rolling robots were the most effective over a flat surface. However, over the inclined surface, the optimization algorithm found
designs with a
inflated
shape capable of shuffling up the hill when oriented at an angle (max fitness 0.0158 BL/s). Using this strategy, the robot achieved combined locomotion of 0.1023 BL/s, outperforming the hand-designed strategy of using crawling on inclines and rolling on flat ground (combined max of 0.0998 BL/s).
The robot was able to inflate two of its external bladders to balance itself like a table, increasing stability between the robot and the ground,
while the non-standard orientation reduced the amount of gravitational force opposing the direction of motion, thereby requiring less propulsive force and reducing the likelihood of the robot rolling back down the hill. However, this behavior is physically unrealistic. Thus, the best strategy for moving up the incline
remains the hand-designed flattened shape which traverses
the hill using an inchworm-like gait.
For the successful evolutionary trials,
the policies found were less finely tuned than those that were hand-designed.
Thus, even in the highly optimized rolling behaviors, there were
occasional counterproductive or superfluous actuations (see Supplementary Movie S1). Such unhelpful motions could likely be reduced via further optimization and by adding a fitness penalty for the number of actuators used per time step.

A similar trend is shown in Fig.~\ref{fig:sim_results}b-c, where the best combined robot for each environment was discovered by the pair of experiments in which the hill climbing algorithm had the most control over the optimization of the robot, despite the larger number of trainable parameters, and thus an increased likelihood of getting stuck in a local minimum. Additionally, the population of simulated robots continued to exhibit similar (and often superior) mean performance compared to the control-only experiments (Fig.~\ref{fig:sim_results}). These observations suggest that the robots avoided local minima, and that more parameters should be mutable during automated design of shape-changing robots.
We hypothesize that maximizing the algorithm's design freedom would be even more important when designing robots with increased degrees of freedom, using more sophisticated optimization algorithms that can operate in an exponentially growing search space.

Overall, this sequence of experiments showed that automated search could discover physically realistic shapes and controllers for our shape-changing robot in
one of the two environments we studied (flat ground).
Although the hand-designed controllers each performed comparably to the best discovered controllers in a single environment, by changing shape, the robot had a better combined average speed in both environments. Concretely, the best shape-controller pair found by hill climbing locomoted at a speeds of 0.1888 BL/s on flat ground and 0.0158 BL/s on incline, resulting in an average speed across the two environments of 0.1023 BL/s, compared to the average speed of -0.1995 BL/s for the round shape with a rolling gait and 0.0424 BL/s for the flat shape with an inchworm gait (Table~\ref{tbl: simulation results}).

%%%%%%%%%%%%%%% Transferring to a physical robot. %%%%%%%%%%%%%%%
\subsection{Transferring to a physical robot.}
% Introduce sim2real
Transferring simulated robots to reality introduces many challenges. For perfect transferal, the simulation and hardware need to have matching characteristics, including: material properties, friction modeling, actuation mechanisms, shape, geometric constraints, and range of motion. In practice, hardware and software limitations preclude perfect transferal, so domain knowledge must be used to achieve a compromise between competing discrepancies. Here, we sought to maximize the transferal of useful behavior, rather than strictly transferring all parameters. %, to accelerate the development of shape-changing robots.

% High-level summary of our instance
In simulation, we found that the same actuators could be used to create different locomotion gaits. When restricted to the cylindrical shape, we found that sequential inflation of the bladders could be used to induce rolling.
A hand designed policy enabled the flatter robots to move with inchworm motion.
To transfer such shape change and gaits to a physical robot, we created a robot which had an inflatable core, eight pneumatic surface-based actuators for generating motion, and variable-friction feet on each edge to selectively grip the environment (Fig.~\ref{fig: summary}). This suite of features allowed the robot to mirror the simulated robot's gaits, including rolling and inching. The ``hand-designed controllers'' from simulation were transferred to reality by sending the same command sequence from a PC to digital pressure regulators~\cite{booth_addressable_2018} that inflated the bladders, resulting in forward motion. However, it was found that different bladders expanded at different rates and had slightly different maximum inflation before failure, so in the experiments shown in this manuscript, the robots were manually teleoperated to approximate the hand-designed controllers with nonuniform timesteps between each actuation state. Further details on the robot hardware are presented in the Methods section.

\begin{figure*}
    \centering
    \includegraphics[width=0.95\textwidth]{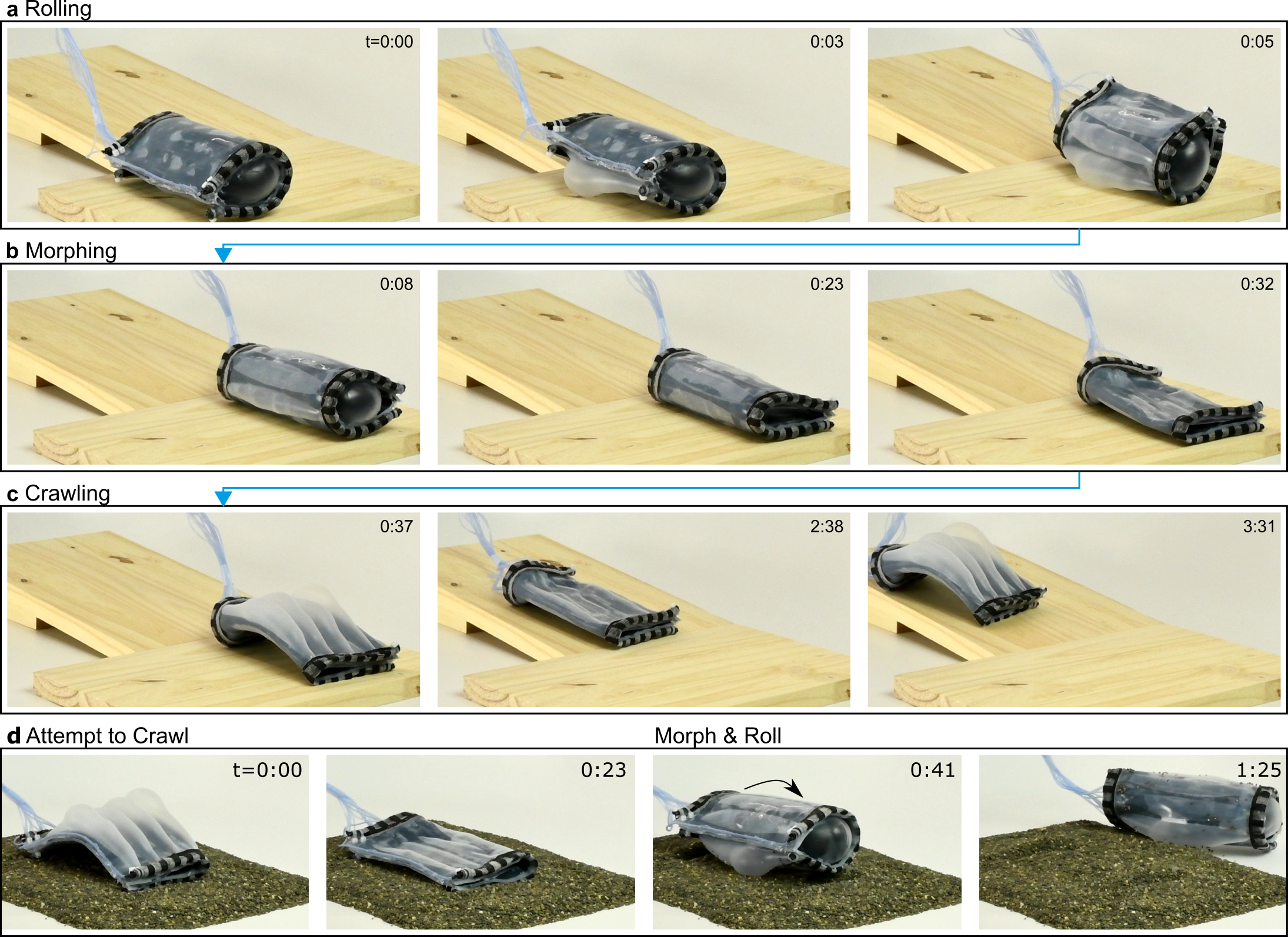}
    \caption{ \textbf{Shape change allowed the physical robot to operate in previously inaccessible environments.} \textbf{a,} While round, the robot's actuators created a rolling gait which was effective on flat ground. \textbf{b,} By deflating its inner bladder, the robot could flatten. \textbf{c,} While flat, the outer bladders induced an inchworm-like gait, allowing the robot to ascend inclines up to ${\sim}14$~degrees. \textbf{d,} The inchworm gait gripped the ground to crawl forward, making it ineffective on granular surfaces. When faced with such a situation, the robot could expand its inner bladder to begin rolling. For length-scale reference, the robot is 10 cm by 15 cm while flattened, and 7~cm diameter by 15~cm while round. Panels \textbf{a-c} correspond to times from a single trial, while panel \textbf{d} is from a different trial and has a separate start time.}
    \label{fig: real_motion}
\end{figure*}

% Rolling
Mirroring simulation, rolling was achieved by inflating the trailing edge bladder to push the robot forward, exposing new bladders that were then inflated one at a time, sequentially (Fig.~\ref{fig: summary}a and Fig.~\ref{fig: real_motion}a).  Each inflation shifted the robot's center of mass forward so the robot tipped in the desired direction, allowing the robot to roll repeatedly. This motion was effective for locomoting over flat ground (average speed 0.05 BL/s). % where a body length was the robot's longest edge length). % 0.7~cm/s
When we attempted to command the robot to roll up inclines, the slope of the incline and the robot's seam made it difficult for the robot to roll. These observations suggest the existence of a transition regime on the physical robot, where the ideal shape-locomotion pair switches from a rolling cylinder to a flat shape with inchworm gait.
However, the boundary is not cleanly defined on the physical hardware: at increasing inflation levels approaching the strain limit of the silicone, the robot could roll up increasingly steep inclines up to ${\sim}9{^\circ}$. After just a few such cycles, the bladders would irreversibly rupture, causing the robot to roll backward to the start of the incline.

% Flattening
By accessing multiple shapes and corresponding locomotion modes, shape-changing robots can potentially operate within multiple sets of environments. For example, when our robot encountered inclines, it could switch shapes (Fig.~\ref{fig: real_motion}a-c and Supplementary Movie S1). To transition to a flattened state capable of inchworm motion, the robot would deflate its inner bladder, going from a diameter of 7~cm (width-to-thickness ratio $\gamma=1$) to an outer height of ${\sim}1.2$~cm ($\gamma{\sim}8.3$) (Fig.~\ref{fig: real_motion}b). The central portions of the robot flatten to ${\sim}7$~mm, which is approximately the thickness of the robot's materials, resulting in $\gamma{\sim}14$. During controlled tests, an average flat-to-cylinder morphing operation at 50~kPa took 11.5 seconds, while flattening with a vacuum ($-80$ kpa) took 4.7 seconds (see Methods for additional details).

% Crawling
Flattening reduced the second moment of area of the robot's cross-section, allowing the bladders' inflation to bend the robot in an arc (Fig.~\ref{fig: real_motion}c). At a first approximation, body curvature is given as $\kappa = \frac{M}{EI}$, where $M$ is the externally-induced moment, $E$ is the effective modulus, and $I$ is the axial cross-section's second moment of area. Thus, flatter robots should bend to higher curvatures for a given pressure. However, even for the flattest shape, bending was insufficient to produce locomotion: on prototypes with unbiased frictional properties, bending made the robot curl and flatten in-place.

\begin{figure*}
    \centering
    \includegraphics[width=6in]{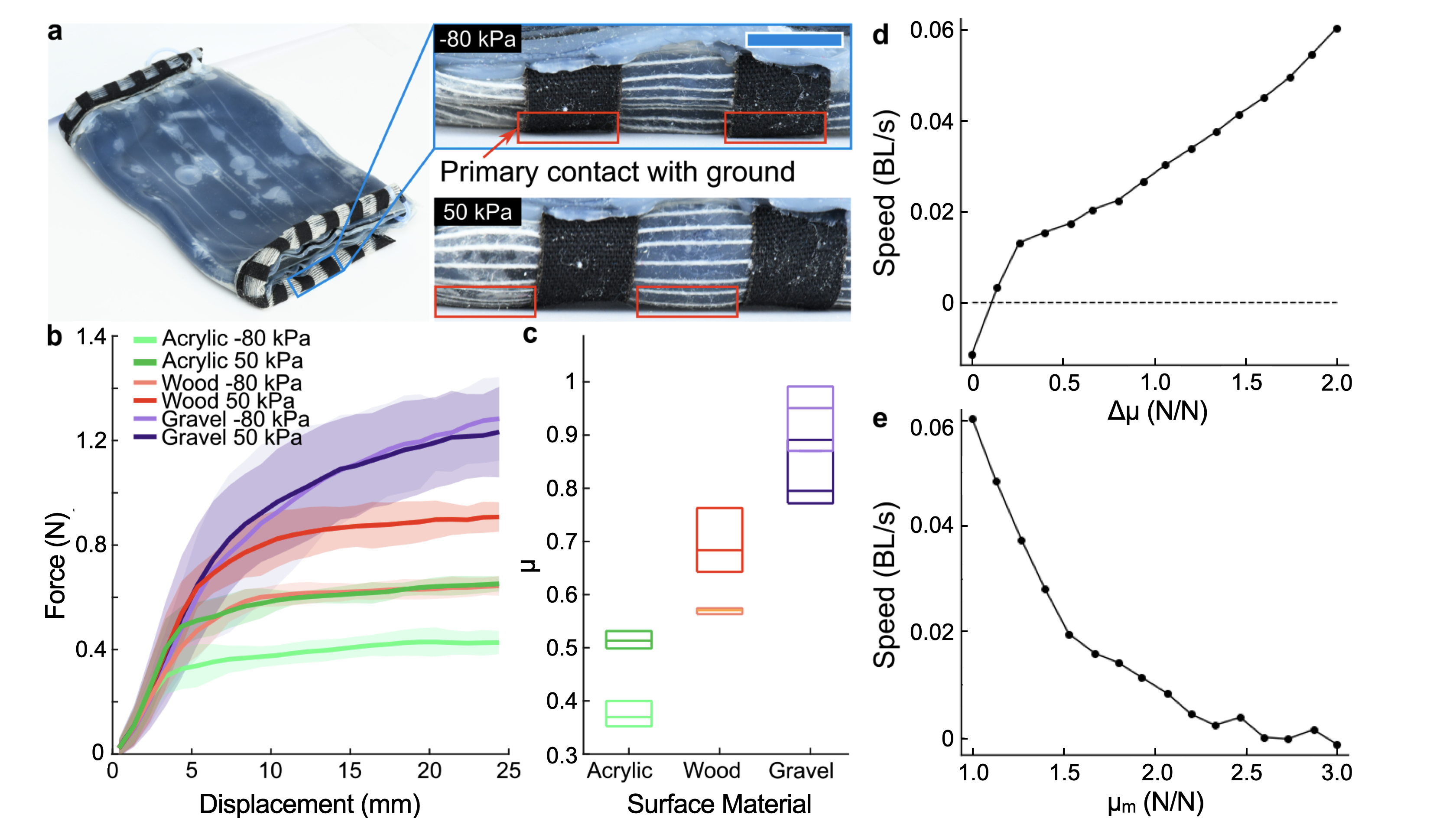}
    \caption{\textbf{The variable-friction feet change their frictional properties when inflated.} \textbf{a,} When the robot's feet are inflated, silicone bladders protrude from their fabric sheath to contact the ground. Blue scale bar on inset represents 1~cm.
    \textbf{b,} Force vs. displacement when the robot was slid over wood, acrylic, and gravel. Each shaded region represents $\pm$one standard deviation about the mean (solid line).
    \textbf{c,} Coefficient of static friction. The boxes denote 25th and 75th percentiles, while the bars represent the median.
    \textbf{d,} Speed (in simulation) as a function of the difference between friction values, $\Delta \mu = \mu_i - \mu_u$ (where $\mu_i$ is friction while the foot is inflated to 50 kPa, and $\mu_u$ is friction while uninflated at -80 kPa). \textbf{e,} Speed (in simulation) as a function of average friction value, $\mu_m = (\mu_i + \mu_u)/2$. In \textbf{d-e}, the hand-tuned inchworm gait was used.}
    \label{fig: friction}
\end{figure*}

% VF feet
Variable-friction ``feet'' were integrated onto both ends of the robot and actuated one at a time to alternate between gripping in front of the robot and at its back, allowing the robot to inch forward (average speed of 0.01~BL/s on flat wood). % 0.17~cm/s, or 
The feet consisted of a latex balloon inside unidirectionally stretchable silicone lamina~\cite{kim_reconfigurable_2019}, wrapped with cotton broadcloth. When the inner latex balloon was uninflated (-80 kPa), the silicone lamina was pulled into its fabric sheath, thus the fabric was the primary contact with the ground. When the balloon was inflated (50 kPa), it pushed the silicone lamina outward and created a higher-friction contact with the ground (Fig.~\ref{fig: friction}a). To derive coefficients of static friction ($\mu$) for both the uninflated ($\mu_u$) and the inflated ($\mu_i$) cases, we slid the robot over various surfaces including acrylic, wood, and gravel. As the robot slid over a surface, it would typically exhibit an initial linear regime corresponding to pre-slip deformation of the feet, followed by slip and a second linear kinetic friction regime (Fig.~\ref{fig: friction}b). From the pre-slip regime, we infer that on a wood surface $\mu_u$ = 0.56 and $\mu_i$ = 0.70 --- an increase of ${\sim}25\%$ (Fig.~\ref{fig: friction}c). On acrylic, $\mu_u$ = 0.38 and $\mu_i$ = 0.51, which is an increase of 35$\%$, yielding an inching speed of 0.007~BL/s. % 0.11~cm/s or 

When the difference in friction ($\Delta \mu = \mu_i - \mu_u$) for the variable-friction feet was too low (such as on gravel), inchworm motion was ineffective, as predicted by simulation (Fig.~\ref{fig: friction}d). Similarly, when the average friction ($\mu_m = (\mu_i + \mu_u)/2$) was too high, it would overpower the actuators and lead to negligible motion (Fig.~\ref{fig: friction}e). On wood, the inchworm gait was effective on inclines up to ${\sim}14{^\circ}$, at a speed of 0.008 BL/s % 0.12 cm/s or 
(Fig.~\ref{fig: real_motion} and the Supplementary Video). Thus, the robot could quickly roll over flat terrain (0.05 BL/s) then flatten to ascend moderate inclines, attaining its goal of maximizing total traveled distance. 

% Real robot can change shape to gain access to different environments

% Variable-friction feet characteristics

%%%%%%%%%%%%%%% Discussion %%%%%%%%%%%%%%%
\section{Discussion}
In this study, we tested the hypothesis that adapting the shape of a robot, as well as its control policy, can yield faster locomotion across environmental transitions than adapting only the control policy of a single-shape robot. In simulation, we found that a shape-changing robot traversed two test environments faster than an equivalent but non-morphing robot. Then, we designed a physical robot to utilize the design insights discovered through the simulation, and found that shape change was a viable and physically-realizable strategy for increasing the robot's locomotion speed. % In contrast, while inflated, the simulated robot could locomote rapidly on flat ground, and roll in the negative direction on the incline. Similarly, the inflated physical robot could roll on flat ground, but its bladders ruptured during attempts to roll up an incline. The simulated and physical robots both needed to change into a flattened shape in order to locomote up the incline.

% Sim2real
We have also shown progress toward an automated sim2real framework for realizing metamorphosing soft robots capable of operating in different environments. In such a pipeline, simulated shape-changing robots would be designed to achieve a desired function in multiple environments, then transferred to physical robots that could attain similar shapes and behaviors. We demonstrated each component of the pipeline on a representative task and set of environments: locomotion over flat ground and an incline. Starting with an initial robot design, the search method sought valid shapes and control policies which could succeed in each
environment, however, it is notable that the search found it difficult to find a successful inchworm gait. This is likely due to a lack of gradient between control policies that do not produce inching and those that do. Thus, a complete pipeline would need to include more sophisticated search methods that are better able to search this space.
The effective shapes and gaits were then transferred to physical hardware. However, the simulation was able to generate some non-transferable behavior by exploiting inaccuracies of some simulation parameters. For example,
when the friction coefficient was too low, the robot would make unrealistic motions such as sliding over the ground.
Other parameters, such as modulus, timescale, maximum inner bladder pressure, and resolution of the voxel simulation (i.e., the number of simulated voxels per bladder), and material density, could be adjusted without causing drastic changes in behavior. Developing a unified framework for predicting the sim2real transferability of multiple shapes and behaviors to a single robot remains an unsolved problem.

% real2sim
Insights from early physical prototypes were used to improve the simulator's hyperparameters (such as physical constants), resulting in more effective sim2real transferal.
Pairing hardware advances with multiple cycles through the sim2real pipeline, we plan to systematically close the loop such that data generated by the physical robot can be used to train a more accurate simulator, after which a new round of simulation to reality transfers can be attempted. This iterative process will be used to reduce the gap between simulation and reality in future experiments.

% Advances in hardware + software --> complicated shape-controller pairs
With advances such as increased control of the physical robots' shape and more efficient, parallelized soft-robot simulators, the pipeline should be able to solve increasingly challenging robot design problems and discover more complicated shape-controller pairs. While the sim2real transfer reported in this manuscript primarily tested intermediate shapes between two extremal shapes | a fully-inflated cylinder and a flattened sheet | future robots may be able to morph between shapes embedded within a richer, but perhaps less intuitive morphospace.
For example, robots could be automatically designed with a set $C$ of $N_c$ inflatable cores and corresponding constraining fabric outer layers. To transition between shapes, a different subset $C$ could be inflated, yielding $2^{N_c}$ distinct robot morphologies. Designing more sophisticated arrangements of actuators and inflatable cores could be achieved using a multilayer evolutionary algorithm, where the material properties of robots are designed along with their physical structure and control policies.~\cite{howard_evolving_2019}.
Additionally, it is unclear how to properly embed sensors into the physical robot to measure its shape, actuator state, and environment. Although some progress has been made toward intrinsically sensing the shape of soft robots~\cite{soter_bodily_2018}, and environmental sensing~\cite{umedachi_gait_2016}, it remains an open challenge for a robot to detect that it as encountered an unforeseen environment and edit its body morphology and behavioral control policy accordingly.

% Those advances --> additional environments? Generate hypotheses and SI
Future advances in hardware and search algorithms
could be used to design shape-changing robots that can operate across more challenging environmental changes.
For example, swimming or amphibious robots could be automatically designed using underwater soft-robot simulation frameworks~\cite{corucci_evolving_2018}, and changing shape within each gait cycle might allow robots to avoid obstacles~\cite{shah2019morphing} or adapt to environmental transitions. We have begun extending our framework to include underwater locomotion, where locomoting between terrestrial and aquatic environments represents a more extreme environmental transition than flat-to-inclined surface environments. 
Our preliminary results suggest that multiple swimming shape-gait pairs can be evolved using the same pipeline and robot presented herein (see Supplementary Information). 
While recent work has shown the potential advantages of adapting robot limb shape and gait for amphibious locomotion~\cite{baines_variable_2020}, closing the sim2real gap on shape-changing amphibious robots remains largely unstudied.

Collectively, this work represents a step toward the closed-loop automated design of robots that dynamically adjust their shape to expand their competencies. By leveraging soft materials, such robots potentially could metamorphose to attain multiple grasping modalities, adapt their dynamics to intelligently interact with their environment, and change gaits to continue operation in widely different environments.

%%%%%%%%%%%%%%% Methods %%%%%%%%%%%%%%%
\section{Methods}
\label{sec: Methods}

\subsection{Simulation environment.}
The robots were simulated with the multi-material soft robot simulator Voxelyze~\cite{hiller2014dynamic}, which represents robots as a collection of cubic elements called voxels. A robot can be made to move via external forces or through expansion of a voxel along one or more of its 3 dimensions.

% \begin{figure}[!ht] % Simulation details
%     \centering
%     \includegraphics[width=\textwidth]{2020nmi_figures/simulation_details.png}
%     \caption{\textbf{a,} Visual rendering of the robot in both its inflated and flat shapes, with the corresponding underlying Euler-Bernoulli beams shown on the right, in red. \textbf{b,} A voxel is represented as a point mass connected to its neighbors by beams, adapted from~\cite{hiller2014dynamic}. \textbf{c,} The pressure ($P$) vectors (red) acting on each interior voxel (green) when fully inflated.}
%     \label{fig: sim_details}
% \end{figure}

\begin{figure}[!ht] % Simulation details
    \centering
    \includegraphics[width=\textwidth]{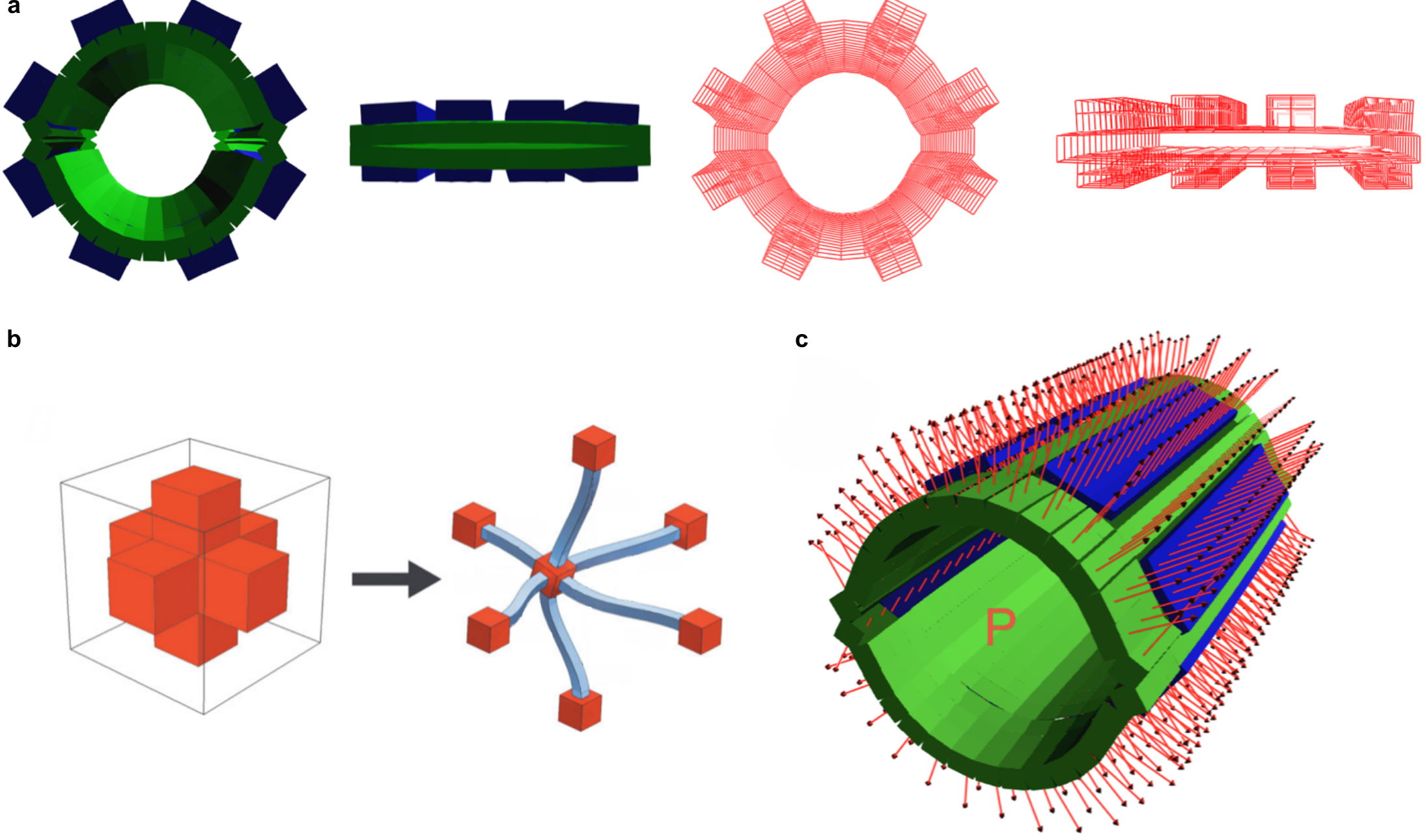}
    \caption{\textbf{The simulated robot could switch between round and flat shapes, as modeled by a shape-changing lattice.} \textbf{a,} Visual rendering of the robot in both its inflated and flat shapes, with the corresponding underlying Euler-Bernoulli beams shown on the right, in red. \textbf{b,} A voxel is represented as a point mass connected to its neighbors by beams, adapted from~\cite{hiller2014dynamic}. \textbf{c,} The pressure ($P$) vectors (red) acting on each interior voxel (green) when fully inflated.}
    \label{fig: sim_details}
\end{figure}

Voxels were instantiated as a lattice of Euler-Bernoulli beams (Fig.S~\ref{fig: sim_details}a). Thus, adjacent voxels were represented as points connected by beams (Fig.S~\ref{fig: sim_details}b). Each beam had length $l=0.01~m$, elastic modulus $E=400~kPa$, density $\rho=3000~\frac{Kg}{m^3}$, coefficient of friction
$\mu=2.0$
, and damping coefficient $\zeta=1.0$ (critically damped). For comparison, silicone typically has a modulus of ${\sim}100-600~kPa$ and density of ${\sim}1000~\frac{Kg}{m^3}$. These parameters were initially set to $E=100~kPa$ and $\rho=1000~\frac{Kg}{m^3}$, but were iteratively changed to increase the speed and stability of the simulation while maintaining physically realistic behavior. % Dragon Skin 10 E = 151 kPa, DS30 E = 592 kPa. DS 10 & 30 density 1.06 g/cc  = 1070 kg/m^3.
% Source: https://www.smooth-on.com/tb/files/DRAGON_SKIN_SERIES_TB.pdf
We simulated gravity as an external acceleration ($g=9.80665~\frac{m}{s^2}$) acting on each voxel. For the flat environment, gravity was in the simulation's negative z direction. Since changing the direction of gravity is physically equivalent to and computationally simpler than rotating the floor plane, we simulated the slope by changing the direction of gravity.
The robot could change shape by varying the force pushing outward, along to the interior voxels' surface normals, representing a discrete approximation of pressure (Fig.S~\ref{fig: sim_details}c). % The core's inflation force ranged from 0 to $\approx1.4~N$ per voxel, representing pressure of $14 ~kPa$. 
The maximum pressure was set at
$12~kPa$
after comparison to prior results (for example, the robotic skins introduced by Shah et al. inflated their pneumatic bladders to under 20 kPa~\cite{shah2019morphing}) and after initial experiments with hardware that suggested only 10${\sim}$35 kPa was necessary. % By comparison, the bladders of the rolling soft robot in our previous work used inflation pressures of 3 psi~\cite{shah2019morphing}. % PancakeBot used 7.5 psi = 50 kPa

The robots' external bladders were simulated via voxel expansion such that a voxel expanded along the z-dimension of its local coordinate space at $3e^{-4}~m$ per simulation step and $1.5e^{-5}~m$ along the x-dimension. Expansion in the y-dimension created a bending force on the underlying skin voxels. This value was changed on a sliding scale from $1.76e^{-4}~m$ to $3e^{-5}~m$ based on the pressure of the robots' core, such that bladder expansion created minimal bending force when the robot was inflated, simulating the expansion of physically-realizable soft robots. Concretely, the y-dimension expansion was computing using a normalizing equation $(b - a) * ((P - P_{MIN})/(P_{MAX} - P_{MIN})) + a$ where $a = 1.7$, $b=10$, $P_{MAX}$ is the maximum outward force per voxel in the robot's core ($1.4~N$), $P_{MIN}$ is the minimum outward force per voxel $0~N$ and $P$ is the current outward force per voxel. These values were adjusted iteratively, until simulated and physical robots with the same controllers exhibited similar behavior in both the inclined and flat environments. Lastly, to prevent the robot from slipping down the hill, and to enable other non-rolling gaits, the robot was allowed to change the static and kinetic friction of its outer voxels between a low value ($\mu=1e{-4}$) when inactive and high value ($\mu=2.0$) when active.

\subsection{Optimization}
The optimization algorithm searched over 3 adjustable aspects of the robots: shape (parameterized as inner bladder pressure), orientation of the robot relative to the incline, and actuation sequence. The algorithm searched over a single number
$p\in[0, 12]$ ($kPa$)
for shape and $\theta\in[0^{\circ}, 90^{\circ}]$ for orientation (see Fig.S~\ref{fig:sim_results}a for illustrations of each parameter). % $0^{\circ}$ faces the robot up the incline, while $90^{\circ}$ places it sideways on the slope

The robot's actuation sequence $S$ over $T$ actuation steps was represented by a binary $10 \times T$ matrix where a $1$ corresponds to bladder expansion and $0$ corresponds to bladder deflation. Each of the first eight rows corresponded to one of the inflatable bladders, and the last two rows controlled the variable friction feet. Each column represented the actuation to occur during a discrete amount of simulation time steps $t$, resulting in a total simulation length of $t*T$. $t$ was set such that an actuation achieved full inflation, followed by a pause for the elastic material to settle. Actuating in this manner minimizes many effects of the complex dynamics of soft materials, reducing the likelihood of the robots exploiting idiosyncraies of the simulation environment.
In this study, we used $t=11670$ timesteps of $\approx$0.000106 ($dt$) seconds each and $T = 8$ for all simulations. We the actuation matrix $T$ is repeated twice for each simulation thus doubling the length of simulation for a total simulation time of $\approx$19.79 seconds ($(t*dt)*2T)$.
To populate $S$, the algorithm searched over a set of parameters (frequency $f$ and offset $\phi$) for each of the ten actuators. Both of these parameters were kept in the range $0-T$ where in our case we set $T=16$. $f$ determined the number of columns between successive actuator activations, where $f=0$ created a row in the actuation matrix of all $1$'s, $f=1$ created a row with every other column filled by a $1$, $f=2$ every two columns filled by a $1$, and so on. $\phi$ specified the number of columns before that actuator's first activation.

We optimized the parameters of shape, orientation and actuation using a hill climber method. This method was chosen for computational efficiency, since a single robot simulation took considerable wall-clock time (approximately 2.5 minutes on a 2.9 GHz Intel Core i7 processor). The hill-climber algorithm needs only one robot evaluation per optimization step, in contrast to more advanced optimization algorithms that often require multiple evaluations per optimization step. The current set of parameters $C$ was initialized to randomly-generated values and evaluated in the simulation, where fitness was defined as the distance traveled over flat ground, or distance traveled up the incline. A variant $V$ was made by mutating each of the parameters by sampling from a normal distribution centered around the current parameters of $C$.
$V$ was then tested in the simulation, and if it traveled farther, the algorithm replaced $C$ with $V$ and generated a new $V$. The process of generating variations, evaluating fitness, and replacing the parameters was done for 200 generations.
To determine the repeatability of such an algorithm, we ran 60 independent hill climbers for each of the 6 experiments, as described in the Results section.

%%%% Physical Robot
\subsection{Manufacturing the physical robot.}
The physical robot was designed to enable transfer of function, shapes, and control policies from simulation, while maximizing locomotion speed and ease of manufacture. In summary, the inner bladder was silicone (Dragon Skin 10, abbreviated here as DS10, Smooth-On Inc.), the cylindrical body was cotton dropcloth, and the outer bladders were made with a stiffer silicone (Dragon Skin 30, abbreviated here as DS30, Smooth-On Inc.) for higher force output. The variable-friction feet were made out of latex balloons, unidirectionally stretchable lamina (STAUD prepreg, described in~\cite{kim_reconfigurable_2019}), and cotton dropcloth. Complete manufacturing details follow.

\begin{figure*}
    \centering
    \includegraphics[width=0.95\textwidth]{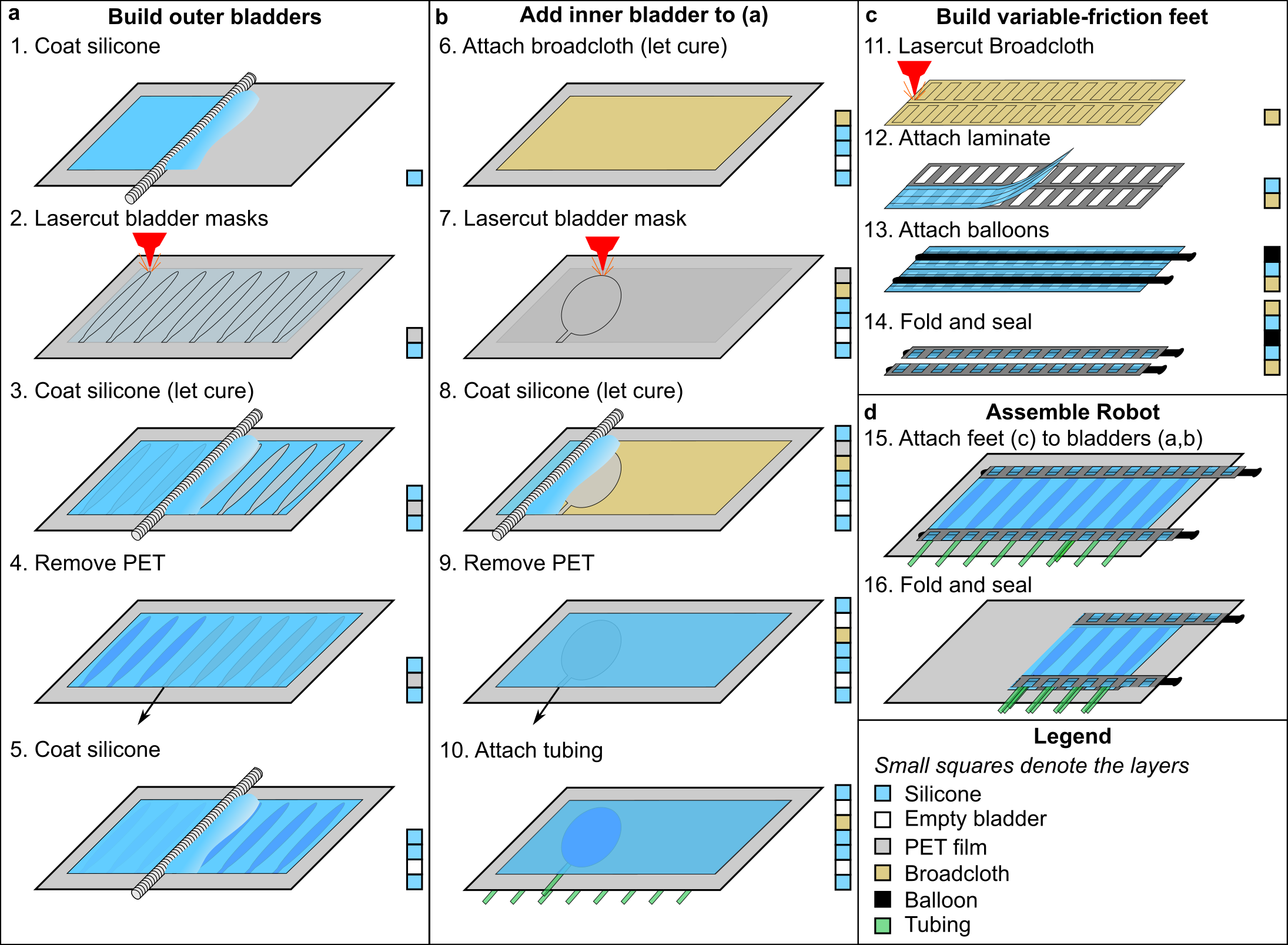}
    \caption{ \textbf{Manufacturing the physical robot.} 
    \textbf{a,} First, the outer bladders were made out of silicone. \textbf{b,} The outer bladders were bonded to silicone-soaked cotton broadcloth, and the inner bladder was fabricated. \textbf{c,} To make variable-friction feet, rectangular slits were lasercut into broadcloth, and unidirectionally stretchable lamina~\cite{kim_reconfigurable_2019} and latex balloons were attached with silicone. \textbf{d,} The robot was assembled by attaching the feet to the main robot body, and the robot was folded to bond the inner bladder to the bladder-less half.
    Small squares to the right of each schematic depict a simplified cross-section of the robot.
    }
    \label{fig:manufacturing}
\end{figure*}

First, the outer bladders were made (Fig.S~\ref{fig:manufacturing}a). Two layers of DS10 were rod coated onto a piece of polyethylene terephthalate (PET). After curing, the substrate was placed in a laser cutter (ULS 2.0), PET-side up, and an outline of the eight bladders were cut into the PET layer. The substrate was removed from the laser cutter and the PET not corresponding to the bladders (i.e., the outer ``negative'' region) was removed. Two layers of DS30 were rod coated onto the substrate. DS30 is stiffer than DS10, and was used to increase the outer actuators' bending force, while DS10 was used in all other layers to keep the robot flexible. Using ethanol as a loosening agent, the encased PET was then removed from all eight bladders. Finally, a layer of DS10 was cast over the bladders' DS10 side for attaching broadcloth to begin manufacturing of the inner bladder.

% % Manufacturing the real robot
% \begin{figure*}
%     \centering
%     \includegraphics[width=0.95\textwidth]{2020nmi_figures/manufacturing.png}
%     \caption{ \textbf{Manufacturing the physical robot.} 
%     \textbf{a,} First, the outer bladders were made out of silicone. \textbf{b,} The outer bladders were bonded to silicone-soaked cotton broadcloth, and the inner bladder was fabricated. \textbf{c,} To make variable-friction feet, rectangular slits were lasercut into broadcloth, and unidirectionally stretchable lamina~\cite{kim_reconfigurable_2019} and latex baloons were attached with silicone. \textbf{d,} The robot was assembled by attaching the feet to the main robot body, and the robot was folded to bond the inner bladder to the bladder-less half.
%     Small squares to the right of the schematic for each step depicts a simplified cross-sectional view of the robot at the end of that step}.
%     }
%     \label{fig:manufacturing}
% \end{figure*}

The inner bladder was made by first soaking cotton broadcloth (15 cm by 20 cm) with DS10, and placing it on the uncured layer on top of the outer bladders (Fig.S~\ref{fig:manufacturing}b). PET was then laid on the robot, and the inner bladder outline was lasercut into the PET. Again, the outer PET was removed, and DS10 was rodcoated to complete the inner bladder. The PET was removed using ethanol and tweezers, and silicone tubing (McMaster-Carr) was inserted into each bladder and adhered with DS10.

To make the variable-friction feet, rectangular slits were lasercut into broadcloth, and unidirectionally stretchable laminate~\cite{kim_reconfigurable_2019} was attached using Sil-Poxy (Smooth-On Inc.) (Fig.S~\ref{fig:manufacturing}c). Latex balloons were attached using Sil-Poxy, and the feet were sealed in half with Sil-Poxy to make an enclosed envelope for each foot. When at vacuum or atmospheric pressure, the fabric would contact the environment, leading to a low-friction interaction. When the feet were inflated, the silicone would contact the environment, allowing the feet to increase their friction.

Finally, the robot was assembled by attaching the feet to the main robot body using Sil-Poxy, and the robot was folded to bond the inner bladder to the bladder-less half, using DS10 (Fig.S~\ref{fig:manufacturing}d).

\subsection{Experiments with the physical robot.}
To test the robot's locomotion capabilities, we ran the physical robots through several tests on flat and inclined ground. The pressure in the robots' bladders was controlled using pneumatic pressure regulators~\cite{booth_addressable_2018}. The robots were primarily operated on wood (flat and tipped to angles up to ${\sim}15^\circ$), with additional experiments carried out on a flat acrylic surface and a flat gravel surface (see Fig.~\ref{fig: real_motion} and the Supplementary Movie S1).

The variable-friction feet were assessed by pulling the robot across three materials (acrylic, wood, gravel) using a materials testing machine (Instron 3343). The robot was placed on a candidate material and dragged across the surface at $100~mm/min$ for $130~mm$ at atmospheric conditions ($23^\circ~ C$, $1~atm$). This process was repeated 10 times for each material, at two feet inflation pressures: vacuum ($-80~kPa$) and inflated ($50~kPa$). The static coefficient of friction, $\mu_s$, was calculated by dividing the force at the upper end of the linear regime by the weight of the robot.

The robot's shape-changing speed was assessed by manually inflating and deflating the robot's inner core for 20 cycles. For each cycle, the robot body was inflated to a cylindrical shape with a line pressure of $50~kPa$, and the time required to attain a diameter of ${\sim}7~cm$ was recorded. The body was then deflated with a line pressure of $-80~kPa$, and the time required to flatten to a height of ${\sim}1.2~cm$ was recorded. % 82.5 kPa

\section*{Supplementary Material}
\label{sec: Supplementary Material}

\begin{itemize} %[noitemsep,topsep=0pt,parsep=0pt,partopsep=0pt]
    \item Movie S1. A soft robot that adapts to environments through shape change
    \item Figure S1. The simulated robot
    \item Figure S2. Manufacturing the physical robot
\end{itemize}

% Manufacturing the real robot

\chapter{Arguments}
\section{Overview}
In this dissertation, I have presented an embodied approach to improve the multitask learning performance of neural networks in robotics. This work introduces novel metrics that quantify how certain designs will perform across multiple tasks and ultimately demonstrates how aspects of a robot's design can improve sample efficiency and resistance to catastrophic interference, allowing for better multitask behavior. Furthermore, the discoveries in this dissertation can easily be combined with other neural centric approaches to these problems and are thus synergistic rather than competitive with other current work in this domain.

In this chapter I will provide a discussion of the significance of the results found in the preceding chapters as well as its potential impact on future work. Thus, it is assumed that the reader is familiar with the main results in the preceding chapters, each of which is a published paper (chapter 4 is in review) that details how a robot's design is a vital component of embodied neural control, and specifically influences whether or not a specific neural network and training algorithm will succumb to the effects of catastrophic interference.

In Chapter 2 I showed, that across three equally capable robots, design variations were the principle factor in determining the amount of catastrophic interference they experienced during training. In Chapter 3 I developed novel metrics that provide a theoretical foundation for measuring the learning ability of a robot design as well as its likely hood to resist catastrophic interference. In Chapter 4 I tested the metrics developed in Chapter 3 across various robot designs and training algorithms, demonstrating that they were predictive of a robot's performance. I also showed how such designs could be found via the co-optimization of morphology and control and that evolution naturally found designs that exhibited sensor homeostasis. Lastly, in Chapter 5 I provided examples from a physical soft robot that was optimized based on these principles to achieve locomotion in different environments.

\section{Significance and Future Work}
This work has implications for a variety of disciplines, thus here I describe how my research compliments the fields of neural network based control, robotics, embodied cognition, automated design, and morphological computation as well as hypothesize how future work may include and improve on discoveries from this dissertation. I also discuss homeostasis and the reality gap, two topics that typically fall under evolutionary robotics research.

\subsection{Neural-Network-Based Control}
In this dissertation, I have shown that a robot's design (e.g. sensor location) can fundamentally alter the optimal weight manifolds of a neural network that controls that robot, thus influencing the controller's learnability and resistance to catastrophic interference. More specifically, by changing sensor location, we observed changes in the number and placement along the loss surface of control parameters suitable for individual environments, and in how these optimal yet environment-specific parameters overlap across different environments to produce generalist controllers which resist catastrophic interference.
%As mentioned, previous efforts to avoid catastrophic interference have relied almost exclusively on increased controller complexity, via different neural architectures and network weight tuning~\cite{kirkpatrick2017overcoming,masse2018alleviating,french1991using,robins1995catastrophic,he2018overcoming,beaulieu2018combating,schwarz2018progress,titsias2019functional}, or algorithmic developments in training protocols~\cite{finn2017model,gidaris2018dynamic}.
Traditional methods for overcoming catastrophic interference focus on finding better ways to navigate the loss surface of task, while the methods here instead alter the loss surface itself. Using this strategy, I have shown that regardless of the algorithm used, it is possible to alleviate catastrophic interference by changing aspects of the robot's design, without increasing controller complexity. Specifically, design changes can suppress or exacerbate the potential for catastrophic interference by expanding or shrinking the overlap of performant controller parameters for that body plan across different environments.

As mentioned, one of the reasons that neurocentric approaches dominate the catastrophic interference literature is likely due to the \emph{Universal Approximation Theorem}~\cite{cybenko1989approximation}, which states that for any continuous function there exists a set of parameters for a sufficiently large neural network that can approximate that function. In these terms, standard approaches are continually seeking new ways to find these parameters no matter the difficulty of the task. The work in this dissertation seeks to change the function itself from a more difficult function to an easier one; one that has a larger set of parameters that can approximate the function. This work shows that the physical design of a robot can influence the learning capabilities of a neural controller, but it is a property that is often abstracted away as an environmental component in other research that addresses this problem. Thus, this research has the capability to transform the current thought paradigm in this and other problems in machine learning research since it shows that including design in the scope of the problem allows for the power to render the loss landscape more benign for learning. This may even extend to currently nonembodied fields such as vision tasks, and natural language processing (NLP). For example, some recent work in NLP trained robots to act similarly to semantically-similar word2vec~\cite{church2017word2vec} encoded commands. This work showed that inducing an alignment between motoric and linguistic similarities (thus improving performance) was facilitated or hindered by the mechanical structure of the robot~\cite{matthews2019word2vec}.

Similarly, state of the art methods in image classification rely solely on neural network training and in particular the convolutional neural architecture. However, certain types of classification tasks are still difficult and unsolved. For instance, currently a neural network can outperform a human on digit classification in the MNIST data set, yet the same model cannot compete with humans in the real world; MNIST is a toy problem when it comes to real world variation in digit appearance which vary in writing medium, writing utensils, color, surroundings context, etc. This same reasoning extends to networks that outperform humans on the ImageNet data set; it is still not as challenging as everyday vision problems the human visual system solves. It is unclear what would allow current systems to move from the world of data sets to human level competence and generality. While, it is likely that there are many advances that will need to be made, this work provides new motivation for embodied approaches to vision tasks. For example, a human classifies an object by much more that what is looks like in two dimensions (2D). In reality objects are three dimensional (3D), they have unique smells, tastes, and physical interactions with ourselves and the rest of the environment. Many objects are classified by what they do more than what they look like. Even when humans make classification in static 2D images, these types of inferences and experiences are brought to bear on the task. This is also true in any discussion of natural language processing; words are  more than the recognition of their characters. The word ``fall'' conveys not only an action but conjures the sensory knowledge of how the stomach feels in free fall. Work that includes embodiment as an element in these tasks was initiated in the past~\cite{sims1994evolving, fitzpatrick2003learning, metta2000babybot}, but has largely been overshadowed in more recent research by rapid deep learning advances. Thus, the work in this dissertation provides the motivation to embody neural networks in other domains and to mate them with robot designs that have been optimized for the network inside to facilitate neural training. However, finding such robot designs may be difficult and unintuitive.

\subsection{Robotics}
Conventional methods for building robots require intensive engineering steps that still heavily rely on human intuition and individual expertise. This design process requires a massive investment of time for both simulated and real robots. Given these current time constraints, a robot progresses through relatively few design and evaluation stages. Even then, it is not clear that the final robot design is highly optimized for the task or its controller~\cite{mehta2018robot, zhang2016performance}. As stated previously, this is typically not seen as a problem in light of Universal Approximation Theorem~\cite{cybenko1989approximation}. Despite this, current robot systems tend to be limited to performing very specific and repetitive tasks that can be programmed without the use of machine learning. This is particularly the case for industrial robots which perform a single function on a production line. Furthermore, using neural controllers in current robotic systems would be inefficient due to the long training times of current neural networks. Thus, neural control in robotics faces two problems: a fundamental inability to compete  with traditional hand designed control algorithms and sample inefficiency.

In this dissertation, I have shown that, from a learning perspective, the best robot designs are often configured into unintuitive designs that would likely be overlooked by human designers. In contrast to human-derived designs, these unintuitive designs have been shown to be easily attuned to their environment and can be generalized to perform well across tasks. Thus, this work provides a foundation for the creation of more general purpose robots, whose designs can be found can be found automatically via optimization. I have also shown that the process of co-optimization (optimization of both design and controller) is more sample efficient than traditional controller optimization alone. In light of the current computational constraints preventing the adoption of neural controllers in robotics, these advances provide a path towards automated design of trainable robots, ultimately costing less human time, energy, and money. While most of this work has been conducted within the domain of traditional rigid robots, the results of this dissertation may be more immediately applicable to the emerging field of soft robotics.

One reason neural control is not used in rigid robots is because of their predictable nature, i.e., we have good mathematical models to describe how they behave given any internal forces or external forces in the environment. Thus, they can usually be controlled with a simpler, and standard proportional-integral-derivative (PID) controller~\cite{bennett1993development}. Where it is often easy to predict the cause and reaction cycle in rigid robots, soft robots  tend to exhibit complex physical properties and often chaotic motion patterns that are difficult to predict, and thus, (PID) control has been infeasible for many soft robotic platforms~\cite{gillespie2018learning, della2021soft}. Soft robots' unintuitive nature results from the vast possibilities of shapes, material properties, motors, and actuators compared to traditional rigid robots. Thus, there is particular interest to use automation in both the design and control of soft robots~\cite{hiller2012automatic,tapia2020makesense,maloisel2021automated,lipson2014challenges}, an example of which I have demonstrated in this work via the optimization of a soft shape-changing robot. 

Another reason soft robots are an ideal platform for design and control optimization is that amorphous material properties allow them to change their design (including shape, rigidity, sensors) over the course of performing some task. Some robots are achieve this by using a subset of a lattice of sensors \cite{kramer2011wearable} or using adjustable antenna\cite{fend2003active}. Some soft robotics projects have allowed for changes in other morphological attributes, such as geometry \cite{kriegman2019automated}, material properties \cite{narang2018transforming}, or the number and placement of actuators \cite{lipson2000automatic}. However, to date the motivation for creating such shape and design changing robots has been scant. The work in this dissertation suggests that a robot that can dynamically change its shape could continually provide a favorable learning environment, from the point of view of its embodied controller, in a variety of environments. Here I have shown a specific example of this by utilizing both automated control and design to create a multi-environment soft robot that can traverse different terrains. Yet, this work is in its infancy, and the robot used in this work has a relatively small space for design variation and actuation. Thus, future work in this area should focus on using more complex robot designs as well as including the metrics developed in this dissertation to measure the effects on the neural controller.

%Collectively, this dissertation provides motivation for the optimization of design elements and even their continuous control over a task. This work opens a path for simplifying and hastening the processes for both the design and control of robots (via automated design) and provides evidence for the increased use of automated design, especially for platforms that rely or plan to rely on neural control algorithms which must be optimized.

\subsection{Embodied Cognition}
As has been mentioned, current neural centric approaches to multitask robotics tend to ignore the consequences of a specific robot design, often touting the generality of the neural control algorithm to apply equally well to any platform. However, in this work I have shown that the robot design has a distinct impact on the ability to train a neural control policy across various training methods. If we think of the robot design as a part of the task environment, this means that changing robot design before or during training is akin to having the ability to alter the task to better suit the neural controller. The success of this strategy in this work reinforces and is reinforced by current trends in the examination of human intelligence in the field of embodied cognition. Cognition is embodied when aspects beyond the brain take a significant role in cognitive processing. In both traditional machine learning and cognitive science, the body is considered as a peripheral component of intelligent behavior~\cite{pfeifer2006body}. Proponents of embodied cognition have been able to show that many of the intelligent processes in nature are a result of deep connections between both the body and brain of organisms~\cite{adams2010embodied}. For example it has been shown that in humans, many of  our peripheral systems work passively and only send signals to the brain when they themselves compute something worth further cognitive computation~\cite{muller2017morphological, baluvska2016having}. Other work mentioned in the introduction has also shown that certain structures (such as the human foot) allow the brain and limbs to perform a consistent function with minimal changes in explicit computations~\cite{kent2019changes, muller2017morphological}. This work, thus, adds a significant contribution to this field of research by showing that benefits of an integrated view of intelligence work in a robotic system as well. This work is also supported by evolutionary biology, which, at varying time scales, co-optimizes both the neural and physical structures of an organism simultaneously. However, this and other work has yet to reach the sophistication of nature's design algorithms.

\subsection{Automated Design}
Although I was able to co-optimize control and design with success, in this work, we relied on simple evolutionary methods (mostly due to the simplicity of the design). For this work to take a larger role in other robotics platforms, much work still needs to be done on the automated design of complex systems.

This dissertation relies on the most simple form of design optimization, called direct encoding, where an optimization algorithm directly manipulates the physical properties of the robot. In biological terms this means that a single gene alone determines a specific phenotypic trait. For instance the optimization algorithm in Chapter 5 mutated one quantity that was used directly as the inflation pressure of the robot core. In Chapters 3 and 4, the evolved floating point numbers were the actual coordinates of the sensors on the robot's body. This method of optimization is not seen in biology and has been shown to be less competent in the literature to date~\cite{gillespie2017comparing, hotz2004comparing}.

Recently, methods have been created that utilize a more biologically plausible indirect-encoding scheme where multiple genes may be used to determine a single design trait and such genes are interpreted (an indirect step) to determine the effect of a design aspect \cite{kassahun2007common}. This later method is prevalent in nature and, is often considered more powerful as it allows for regularity in a design. Regularity is typically used to describe the compressibility of the information required to produce a structure that involves symmetries and repetition of design motifs~\cite{lipson2007principles}. Thus, a good design pattern the is produced by a set of genes can be reused and varied more easily, for example like the finger of a hand. The utility of indirect encoding has been proven by recent work in automated design \cite{clune2011performance, stanley2002evolving, cheney2018scalable}, which has been shown to outperform a typical direct encoding scheme. However, much of this work is still performed on relatively simple robot designs compared to those currently used in reality, and work is still ongoing on working to improve encoding schemes for real world complexity~\cite{clune2011performance}.
 
This research gives insight into how robotic design is a key element in the goal of achieving general artificial intelligence and provides a basis to continue to improve the state of automatic design algorithms. This work also provides motivation to develop these methods in a manner consistent with simultaneous co-optimization of both brain and body.

\subsection{Homeostasis}
The work presented here provides more than a simple motivation to renew work on automated design or co-optimization; it proposes a mechanism by which such work may be judged. Here we have shown that the design of a robot is an important factor in determining its learning performance, but this alone does not allow one to determine how to measure the quality of a design during the optimization process. Without such a mechanism, it is impossible to create a purely design-oriented optimization problem seeing as it would lack any goal. However, in chapter 4 I proposed a possible proxy for the measurement of body designs resistant to catastrophic interference, namely, homeostasis. In that chapter it was shown that during co-optimization of body and control, with the sole goal being performance, the designs found followed a natural gradient from those that experience ``different'' environments very differently to those that experience them as nearly identical. Specifically, if we keep track of the light signal over time, and measure the similarity between that signal across different environments, we find that designs later in the optimization have much more similar signals as measured by dynamic-time-warping. Furthermore, these are the same designs that were found to have high scores for learnability and resistance to catastrophic  forgetting. Thus, signal similarity could be used as a proxy metric for the more comprehensive measurements of $M_{L}$ and $M_{CI}$. While this was not explicitly tried in this work, future work could focus on using homeostasis between a robot's sensors to discover designs that are more favorable to learning. It is also likely that there are other mechanisms that describe why certain designs perform better than others, thus future work should also be focused on expanding the scope of possible mechanisms beyond homeostasis.

\subsection{Morphological Computation}
One branch of evolutionary robotics that stands to benefit from the discoveries in this work is morphological computation. Research in this field focuses on showing that the body of an organism or robot actually can and does perform computations. However, this field has often struggled with two distinct problems. The first is that research often simply finds new ways to invent mechanical computers that are inspired by lifelike forms. Yet, these structures become completely devoid of the functions found from their counterparts in nature. For instance one paper showed how the waves produced by an octopus-like arm could be used for computing~\cite{nakajima2015information}, however, it does not state that octopi can or would use their arms in this manner. My work does not contribute to this type of morphological computation research. However, the second type of work done in this field suffers from the  problem of quantifying what or how much computation a morphological component is performing. In this line of reasoning, one of the first morphological computation experiments is that of passive dynamic walkers~\cite{tedrake2004actuating, collins2005efficient, vaughan2004evolution}. These are bipedal structures without motors that can walk with a natural human like gait down a slight incline with perfect balance; with added motors these robots have a morphology that allows for a simple controller to perform walking, thus, the morphology, by its design, is said to be handling much of the computation necessary to walk since the controller no longer performs many of the difficult calculations for balance and joint control. There are many other such experiments of this manner~\cite{hauser2014morphological, muller2017morphological},  which all amount to describing ways that a design can make the controller task easier or require less computation. However, as we have discussed, they often fall short of quantifying just how much computation their designs are performing. Furthermore, most research does not specifically focus on improving designs for neural control.

The work I have shown here introduces two novel metrics $M_{L}$ and $M_{CI}$, that could be used in morphological computation research to explicitly define the type of computation being performed by a specific design. Specifically, $M_{L}$  could be  used to describe the computational work done by a design as an increase in the optimal parameter space, thus quantifying what it means for a design to make a control problem easier. This is especially true since the metric was shown to be causally correlated with sample efficiency.

\begin{figure*}[t]
    \centering
    \includegraphics[width=0.9\linewidth]{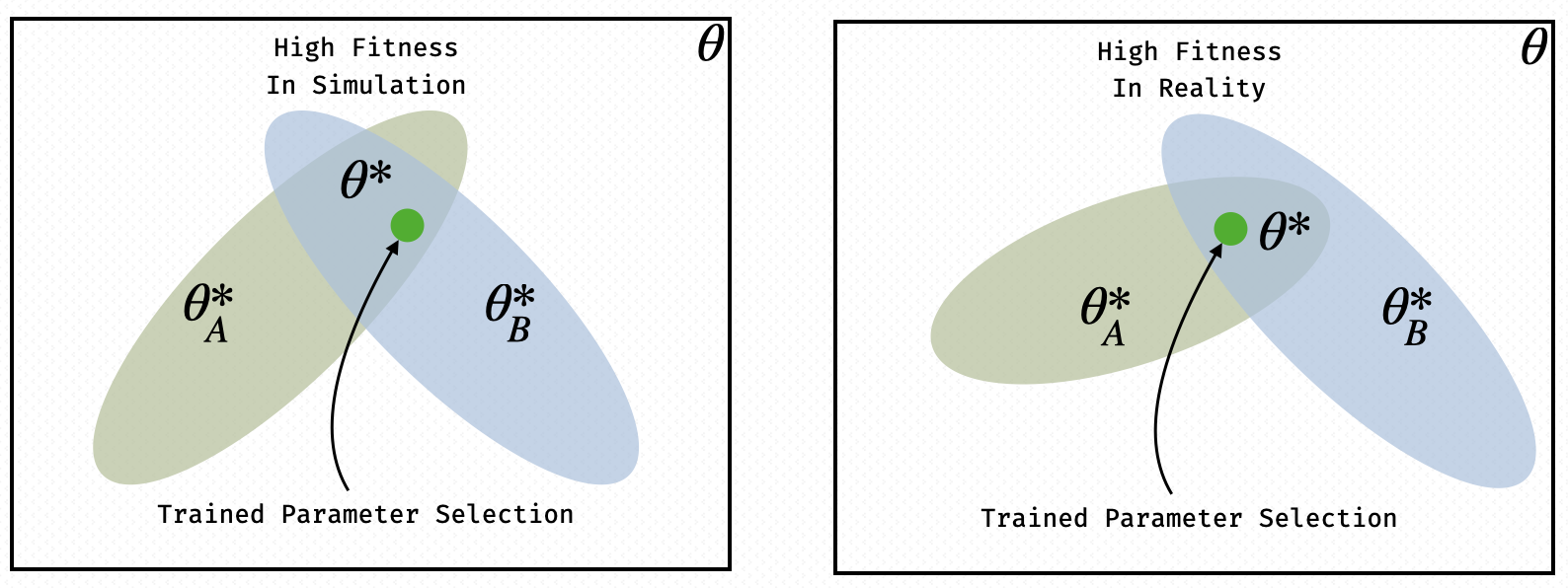}
    \vspace{15pt}
    \caption{\textbf{(Left)} A hypothetical weight manifold for a robot trained in simulation. \textbf{(Right)} The weight manifold for the same task but showing the true optimal parameter space dictated by reality. We can imagine a shift in physics as a shift in the location of the optimal weight manifold $\theta^*$ in the total space $\theta$. Thus a robot design that creates a larger $\theta*$ would be more likely to successfully transfer its learned behavior to reality transfer to reality.}
    \label{fig:realitygap}
\end{figure*}

\subsection{Reality Gap}
The reality gap is the term given to the failure of systems trained and tested in simulation to perform adequately when the system is transferred into a real physical counterpart (when we move from simulation to reality). Here the gap refers to the difference in performance which is often catastrophic, meaning the system is not usable once transferred to reality. The  reality gap problem was first introduced by Jakobi in 1997~\cite{jakobi1995noise}. However, this problem has resurfaced with the use of neural networks in robotics~\cite{koos2013transferability,tremblay2018training,cruz2020closing,prakash2019structured}. Neural networks require many samples of data, which for robotics is most easily obtained by running physics simulations that recreate the characteristics of a real task. Again, when the trained neural network is transferred to a real robot, the performance often varies to the degree that the transfer is considered a failure. The most common way to combat this is to either increase the generalization ability of the network~\cite{tremblay2018training}, this can involve introducing noise or varying physics in the simulation, or using standard neurocentric methods: normalization, training schedules, different architectures, rapid adjustment learning, etc~\cite{lomnitz2020general,james2019sim}. However, Future work based on this dissertation may have the potential to shift the current paradigm by focusing on finding robot designs that are robust to environmental changes.

In this work, I found robot designs that were robust to differing environments (locations of a light source); this principle could similarly be applied to the reality gap by finding robot designs robust to other types of changes in the environment such as physics. It is also possible the designs found in this dissertation, whose robustness allows for multitask behavior, are not mutually exclusive from the same designs that may allow for successful transfer to reality. As described in Fig.~\ref{fig:realitygap}, we could similarly compare the weight manifold for one of the best phototaxis designs found in this dissertation against the baseline human designs. We can imagine a shift in physics as a shift in the location of the optimal weight manifold $\theta^*$ in the total space $\theta$. When we look at the problem in this manner, we can see that the hand designed robot is very unlikely to experience a shift that would allow it to maintain the same parameters and experience the same behavior from simulation to reality. This is because there are very few parameters that allow for optimal behavior. On the other hand, the optimized design may experience many types of shifts, and the parameters which were optimal in the simulation would have a much higher probability of remaining in the shifted optimal region in reality. The possible problems with this theory are that the physics may also manipulate not just the location but the total size of the optimal weight manifold as well. Testing this theory might also be difficult by requiring a large number of robot to be fabricated. However, the work done thus far provides the motivation to test this hypothesis in future and to further apply the principles of design optimization to this and other areas in robotics research.

\section{Conclusion}
In this dissertation, I have presented an embodied approach to improve the multitask learning performance of neural networks in robotics. This work introduces novel metrics that quantify how certain designs will perform across multiple tasks and ultimately demonstrates how aspects of a robot's design can improve sample efficiency and resistance to catastrophic interference, allowing for better multitask behavior. However, this work may also be applicable to many other areas of machine learning, robotics, and engineering. While the research in this dissertation has been mainly focused on examining neural control in simple simulated robots in an arguably simple task, this work strongly suggests that the conclusions drawn in this dissertation would apply generally to more complex robots and tasks. Thus, future work should focus on proving this explicitly by scaling up to more complex robots and environments. This would be most helpful to the larger community by choosing multitask environments that have been used in other robotics experiments to compare the performance of design-optimized robots against a common benchmark. This would also provide the ability to test synergies between the methods described in this dissertation and the more common neurocentric approaches described throughout this dissertation. Other avenues of future work should also explore the relationship between design and other common robotics problems. I have mentioned some of these problem here such as problems in soft robotics, the reality gap, or morphological computation. Lastly, future work should focus on deploying these techniques in real robotic platforms as some designs that may be most beneficial in simulation could prove to be less feasible from a manufacturing standpoint. The best designs in this dissertation were often unintuitive, thus, learning to modify the space of possible designs by realistic manufacturing capabilities could be necessary to realize the benefits of design optimization in real robotic platforms.

\newpage
\singlespacing
\nocite{powers2018effects}
\nocite{powers2020morphology}
\nocite{powers2021good}
\nocite{shah2021soft}
\printbibliography

\end{document}